\documentclass[]{gtech}

\usepackage{amssymb}
\usepackage{multirow}
\usepackage{bigdelim}
\usepackage{todonotes}
\usepackage{longtable}
\usepackage{tabularray}
\usepackage{wrapfig}
\usepackage[most]{tcolorbox} 
\usepackage{xcolor}
\usepackage{url}
\usepackage{algorithm}  
\usepackage{algpseudocode} 
\usepackage[toc,page]{appendix}
\usepackage{multirow}
\usepackage[normalem]{ulem}
\usepackage{adjustbox}
\usepackage{booktabs}  
\usepackage{natbib}
\usepackage{colortbl}  
\usepackage{subcaption}
\usepackage{pifont}
\usepackage{amsmath}
\usepackage{amsfonts}       
\usepackage{multirow}
\usepackage{mathrsfs}
\usepackage{tabularx}
\usepackage[utf8]{inputenc} 
\usepackage[T1]{fontenc}    
\usepackage{hyperref}       
\usepackage{url}            
\usepackage{booktabs}       
\usepackage{nicefrac}       
\usepackage{microtype}      
\usepackage{xcolor}         
\definecolor{ourreward}{HTML}{32bbee}
\definecolor{sparsereward}{HTML}{ed9a81}
\usepackage{graphicx}
\usepackage{wrapfig}
\usepackage{cleveref}

\usepackage{bbding}

\usepackage{CJK}

\usepackage{bbm}
\useunder{\uline}{\ul}{}

\title{%
  \texorpdfstring{%
    \raisebox{-0.23\height}{\includegraphics[scale=0.03]{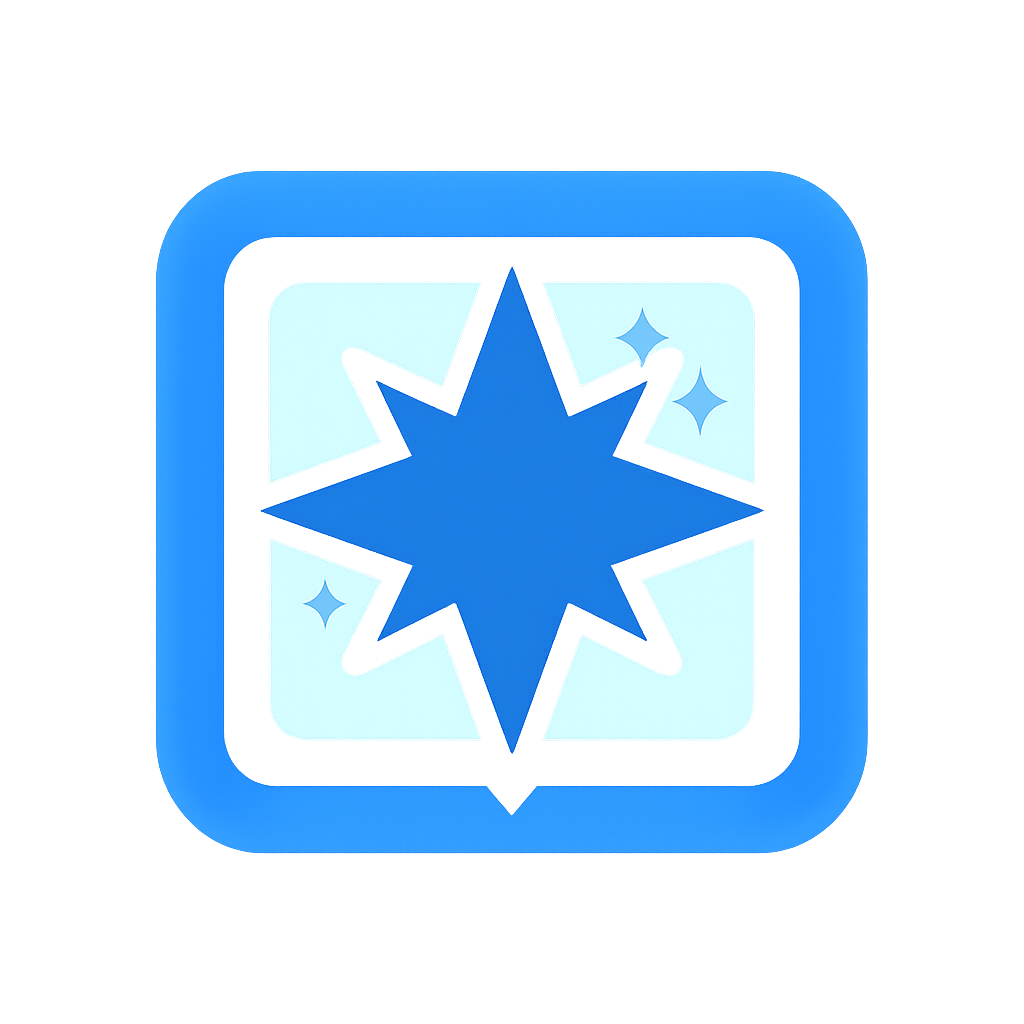}}%
    ~UI-Venus-1.5 Technical Report%
  }{UI-Venus-1.5 Technical Report}%
}

\author{Venus Team, Ant Group}

\abstract{
GUI agents have emerged as a powerful paradigm for automating interactions in digital environments, yet achieving both broad generality and consistently strong task performance remains challenging.
In this report, we present \textbf{UI-Venus-1.5}, a unified, end-to-end GUI Agent designed for robust real-world applications. 
The proposed model family comprises two dense variants (2B and 8B) and one mixture-of-experts variant (30B-A3B) to meet various downstream application scenarios.
Compared to our previous version, UI-Venus-1.5 introduces three key technical advances: (1) a comprehensive Mid-Training stage leveraging 10 billion tokens across 30+ datasets to establish foundational GUI semantics; (2) Online Reinforcement Learning with full-trajectory rollouts, aligning training objectives with long-horizon, dynamic navigation in large-scale environments; and (3) a single unified GUI Agent constructed via Model Merging, which synthesizes domain-specific models (grounding, web, and mobile) into one cohesive checkpoint. Extensive evaluations demonstrate that UI-Venus-1.5 establishes new state-of-the-art performance on benchmarks such as ScreenSpot-Pro (\textbf{69.6\%}), VenusBench-GD (\textbf{75.0\%}), and AndroidWorld (\textbf{77.6\%}), significantly outperforming previous strong baselines. In addition, UI-Venus-1.5 demonstrates robust navigation capabilities across a variety of Chinese mobile apps, effectively executing user instructions in real-world scenarios. 
}

\gtechdata[Code]{\url{https://github.com/inclusionAI/UI-Venus}}
\gtechdata[Model]
{\url{https://huggingface.co/collections/inclusionAI/ui-venus}}

\begin{document}
\maketitle

\renewcommand{\thefootnote}{}



\begin{figure}[htbp]
	\centering
	\includegraphics[width=0.9\textwidth]{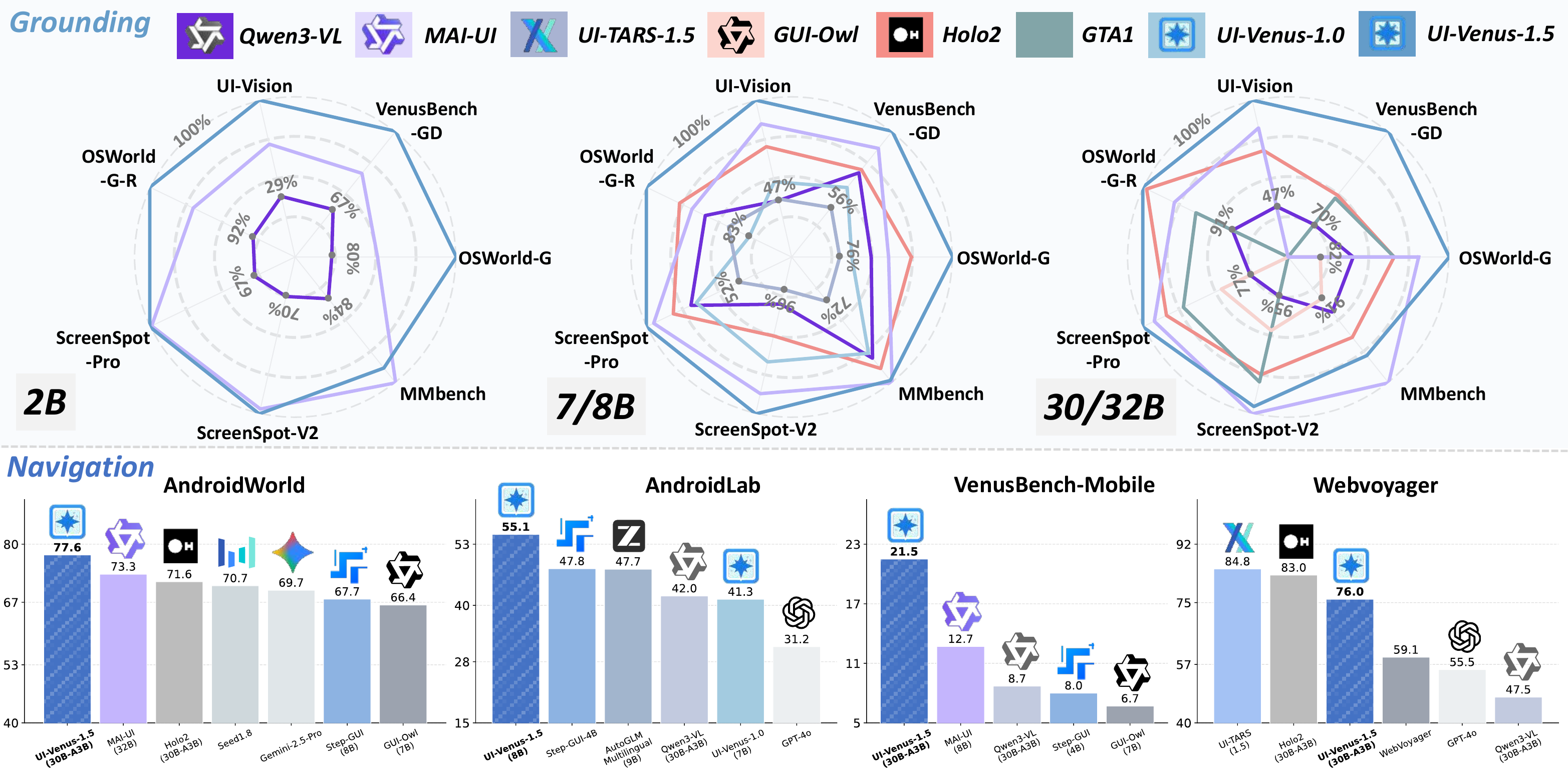}
	\caption{\textbf{UI-Venus-1.5} achieves SOTA performance across multiple GUI grounding and navigation benchmarks. Note that in the three radar charts of grounding, we have normalized the results of the top-performing model to 100\% to more clearly differentiate comparisons among various baselines.}
	\label{fig:performance_uivenus}
\end{figure}



\section{Introduction}

The pursuit of creating intelligent systems capable of autonomously operating digital devices has long been a central goal in artificial intelligence. With the rapid evolution of Multimodal Large Language Models (MLLMs)~\cite{anthropic2024cuda,bai2025qwen25vltechnicalreport,zhu2025internvl3exploringadvancedtraining,glm-4.5v,seed1.8,Qwen3-VL,team2025every1,ai2025ming,team2025every2}, GUI Agents~\cite{gu2025ui,guig2,ye2025mobile,yan2025step,zhou2025mai,wang2025opencuaopenfoundationscomputeruse,liu2024autoglm,hai2025holo2modelfamily,zhang2026omegausebuildinggeneralpurposegui} have emerged as a promising solution to bridge the gap between human instructions and digital execution. Unlike traditional automation tools that rely on rigid APIs, these agents leverage visual perception to interact directly with graphical interfaces, effectively mimicking human behavior to navigate web and mobile environments.

Currently, this field is experiencing a period of intense development. The research community is actively exploring various dimensions of agent construction, ranging from the curation of large-scale GUI datasets~\cite{zhou2025mai,zhou2025venusbench,li2024screenspot-pro,androidworld,androidcontrol,gu2023mobile,he2024webvoyager,guiodyssey} to the optimization of training paradigms~\cite{chen2025tgrpo}. While early works focused on basic feasibility like purely Supervised Finetuning(SFT), recent studies have shifted toward more sophisticated approaches, such as designing rule-based rewards for offline/online reinforcement learning and optimizing token usage for better efficiency. Despite this progress, building an agent that is both universally capable and easy to deploy remains a significant challenge. Building on this momentum, we firstly released UI-Venus-1.0~\cite{gu2025ui} not long ago. By relying solely on reinforcement learning, UI-Venus-GD and UI-Venus-Navi achieved previous state-of-the-art (SOTA) results in grounding and mobile navigation tasks, respectively.

While the GUI agent landscape has expanded rapidly, the rapid influx of new GUI agents has escalated performance standards, challenging the dominance of UI-Venus-1.0. Beyond pure performance metrics, we observe a critical discrepancy between step-level and trace-level accuracy during both SFT and offline reinforcement learning phases. This mismatch is largely due to the sparsity of rewards in individual steps and the inherent domain shift between training data and real-world benchmarks.
Moreover, we posit that an ideal agent suitable for daily usage should be an \textbf{end-to-end} system that adheres to a \emph{simple yet effective} design philosophy. To address these challenges, we present \textbf{UI-Venus-1.5} in this report, a substantially enhanced version of our previous system. Compared to UI-Venus-1.0, UI-Venus-1.5 introduces three key technical advances that jointly improve the final performances:

\begin{itemize}

	\item \textbf{Mid-Training Stage:} Unlike the previous approach, we have added a comprehensive mid-training stage before the reinforcement learning phase. This stage utilizes an extensive corpus of GUI data, comprising 30+ datasets and a total of 10B tokens. By incorporating this step, we equip the base model with robust inherent GUI knowledge, enabling it to effectively solve GUI-related VQA, grounding, and simple navigation tasks even before entering the reinforcement learning stage.
	\item \textbf{Scaled Online Reinforcement Learning:} Recognizing that Online Reinforcement Learning is a highly effective method for training GUI Agents, we have integrated it into UI-Venus-1.5 specifically for mobile and web scenarios. Inspired by T-GRPO~\cite{chen2025tgrpo}, we perform full trajectory rollouts and reward calculations across different environments, which also contributes to address the challenging step-trace accuracy mismatch problem during GUI Agent finetuning. By scaling up the interaction devices, we have further improved the model's performance in complex navigation tasks.
	\item \textbf{Unified Single-Agent via Model Merging:} A major distinction from UI-Venus-1.0 is that UI-Venus-1.5 is a purely end-to-end model, which greatly simplifies deployment for users. To achieve this, we first conduct finetuning for grounding, web, and mobile domains by using domain-specific reward functions, data, and prompts. Once we obtain these three specialized models, we apply a model merge strategy. This allows us to combine them into a single, unified model with minimal performance loss across individual domains.

\end{itemize}

We extensively evaluate UI-Venus-1.5 on diverse benchmarks to verify its versatility and robustness. In terms of GUI Grounding, our model establishes a new state-of-the-art on challenging datasets like ScreenSpot-Pro~\cite{li2024screenspot-pro} and VenusBench-GD~\cite{zhou2025venusbench}, achieving accuracies of \textbf{69.6\%} and \textbf{75.0\%} respectively, which substantially surpasses existing strong baselines like Seed1.8~\cite{seed1.8}, Holo2~\cite{hai2025holo2modelfamily} and MAI-UI~\cite{zhou2025mai}. Furthermore, in dynamic Navigation tasks, UI-Venus-1.5 proves the efficacy of our proposed training pipeline. On the highly competitive AndroidWorld~\cite{androidworld} benchmark, it attains a success rate of \textbf{77.6\%}, outperforming Mobile-Agent-v3~\cite{ye2025mobile}, MAI-UI~\cite{zhou2025mai} and StepGUI~\cite{yan2025step}. When extended to web-based interaction on WebVoyager~\cite{he2024webvoyager}, UI-Venus-1.5 achieves 76.0\% accuracy, closely matching the performance of leading prior baselines. Even when compared to significantly larger models, UI-Venus-1.5 delivers superior efficiency and decision-making accuracy across both grounding and navigation domains.

Beyond achieving state-of-the-art results on standard GUI benchmarks, we place a strong emphasis on the practical utility of UI-Venus-1.5. To this end, we have specifically optimized the model for 40+ \textbf{Chinese mobile ecosystem}, ensuring it can handle a wide array of complex, real-world tasks within third-party applications. These capabilities include, but are not limited to, ticket booking, purchase of goods, and automated conversation management. By mastering these diverse and high-demand scenarios, UI-Venus-1.5 moves closer to becoming a truly helpful digital assistant for everyday life.

\section{Methodology}



\begin{figure}[htbp]
\centering
\includegraphics[width=\textwidth]{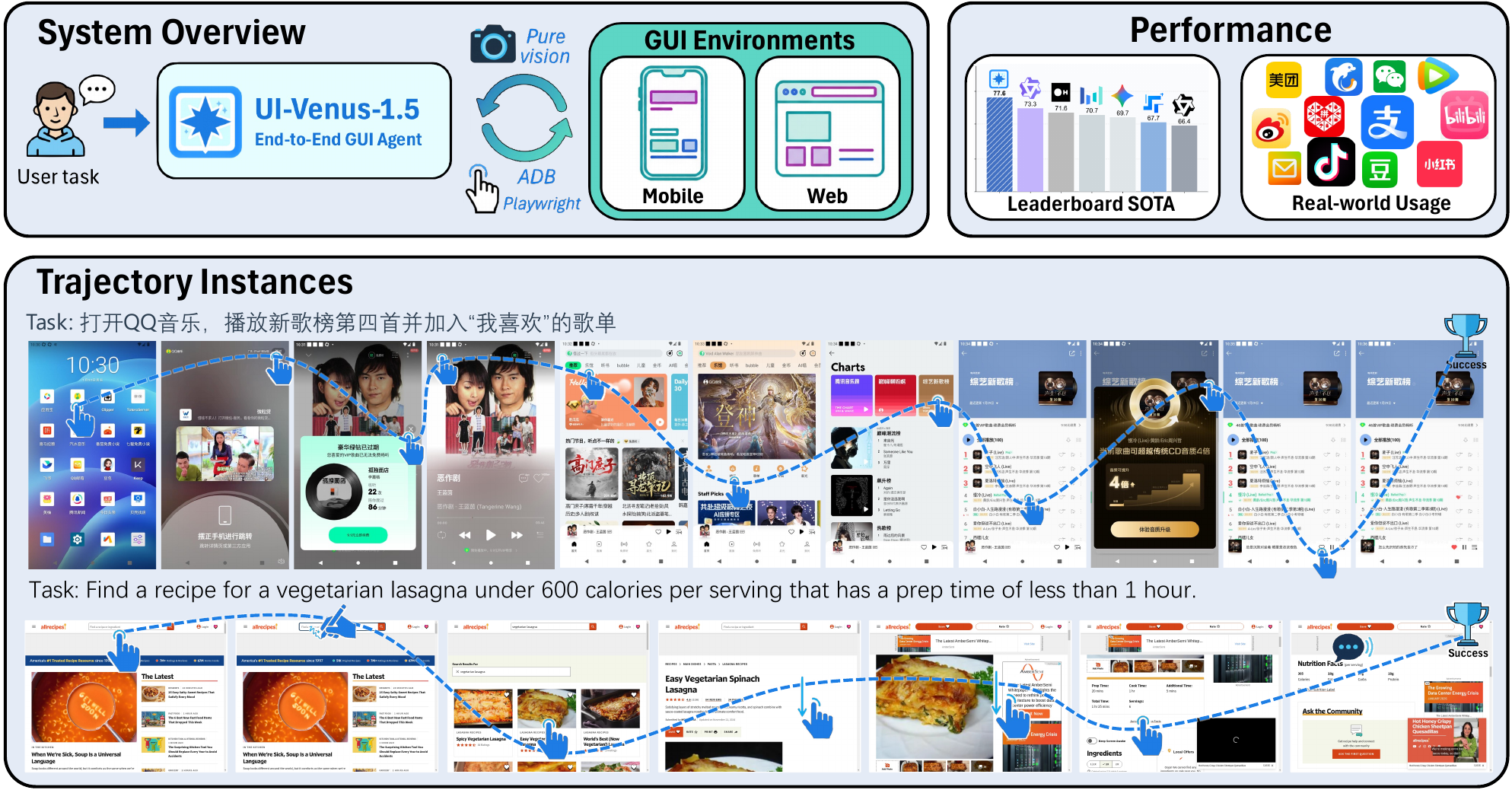}
\caption{\textbf{System Overview of UI-Venus-1.5}. It operates as an end-to-end GUI Agent that interprets user instructions, perceives interface states through screenshots, and executes interactive actions (e.g., clicking, typing, scrolling) to accomplish tasks across diverse executable environments. 
}
\label{fig:venus15_overview}
\end{figure}

\textbf{System overview:} UI-Venus-1.5 is an end-to-end multimodal agent designed to bridge high-level user intentions with concrete GUI interactions across mobile and web platforms. As shown in Figure~\ref{fig:venus15_overview}, the system operates via a closed-loop perception-action mechanism: given a natural language command, the model interprets visual screenshots, grounds semantic intentions into executable actions, and iteratively interacts with the environment until task completion. This unified architecture eliminates the need for handcrafted intermediate representations or API integrations, enabling seamless deployment on heterogeneous interfaces. Beyond state-of-the-art benchmark performance, UI-Venus-1.5 demonstrates robust practical applicability, having been validated across a diverse array of real-world applications ranging from media streaming to complex e-commerce platforms. Whether manipulating native mobile apps or navigating dynamic web content, the model exhibits consistent robustness in handling complex, context-dependent workflows, positioning it as a versatile solution for automating daily digital tasks.

\textbf{Action Spaces:} Building upon the foundational action space of UI-Venus-1.0, we expand the model's capabilities to encompass web-specific interactions. Specifically, we introduce three additional primitives: \textit{Hover}, \textit{DoubleClick}, and \textit{Hotkey}. This augmented action space (Table~\ref{tab:action_space}) unifies mobile and web interaction modalities, allowing the end-to-end model to execute precise operations across diverse environments.

\begin{figure}[htbp]
\centering
\includegraphics[width=0.9\textwidth]{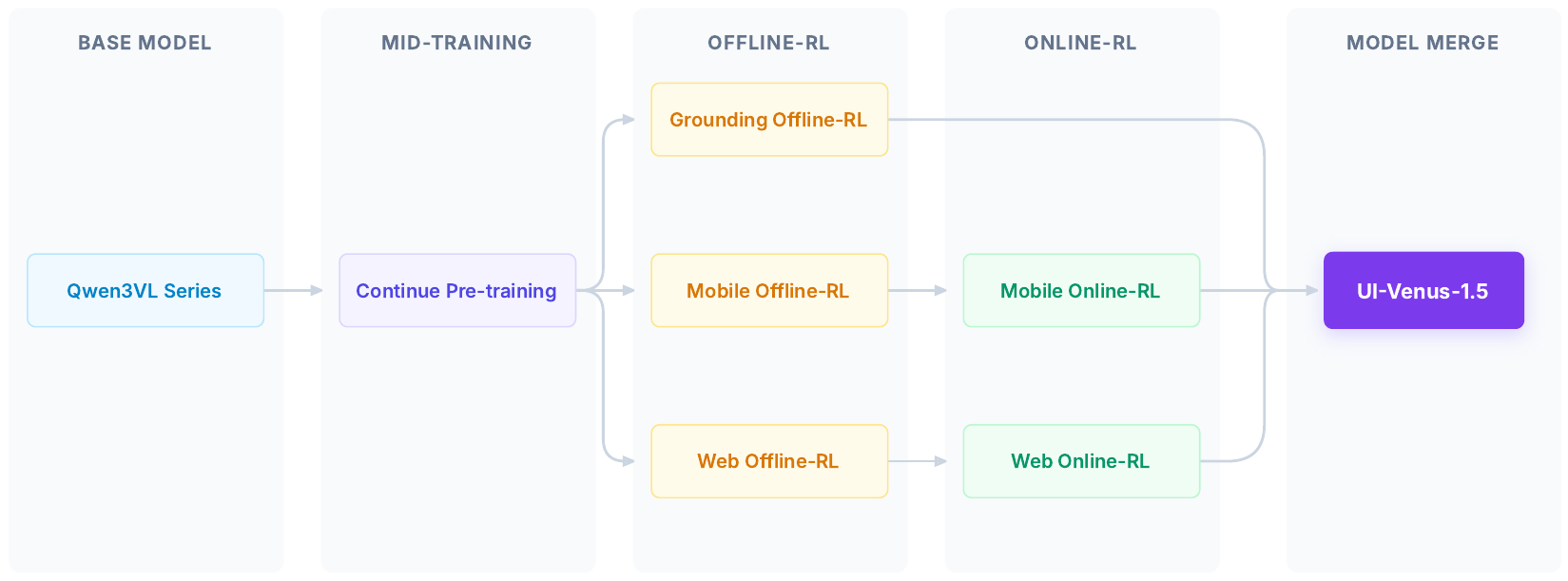}
\caption{\textbf{The Four-Stage Pipeline of UI-Venus-1.5}. Starting from Qwen3-VL Series, the model progresses through a multi-stage curriculum: (1) Mid-Training on large-scale GUI data for domain knowledge injection; (2) Offline-RL for task-specific optimization across grounding, mobile, and web objectives; (3) Online-RL to enhance navigation in complex, real-world settings; and (4) Model Merge, which unifies the specialized models into the final UI-Venus-1.5.}
\label{fig:venus15_pipeline}
\end{figure}

\textbf{Overall Training Pipeline:} 
Subsequently, we elaborate on the model’s training process, which is divided into four distinct stages as shown in Figure~\ref{fig:venus15_pipeline}: (1) a Mid-Training phase for knowledge injection using large-scale GUI data in Section~\ref{sec:mid-train}; (2) task-specific Offline-RL training for each of the three objectives in Section~\ref{sec:offline-rl}; (3) Online-RL to further enhance the agent's navigation capabilities in complex, real-world scenarios in Section~\ref{sec:online-rl}; and (4) a model merge strategy to unify the specialized models in Section~\ref{sec:model-merge}.

\subsection{Mid-Training}\label{sec:mid-train}

\subsubsection{Motivation and Data Collection}



The Mid-Training phase is designed to bridge the semantic gap between general visual perception and the fine-grained structural understanding required by GUI Agents. Specifically, general-purpose vision-language models often lack the granularity needed to capture the structural nuances of user interfaces. In the subsequent reinforcement learning phase, this deficiency severely hinders effective exploration, leading to sparse reward signals and preventing policy improvement from bootstrapping in complex interaction scenarios. This limitation is mainly due to the scarcity of GUI-specific structural modeling in standard pre-training corpora. Consequently, rather than relying solely on capability elicitation, we shift our objective toward \textbf{foundational representation building}. This phase (Mid-Training) enables the model to encode diverse GUI layouts and interaction logic, providing a robust initialization for subsequent policy optimization, which can also be experimentally verified in Section~\ref{sec:mid-train-exp}.

\begin{figure*}[ht]
    \centering
    \begin{subfigure}{0.35\textwidth}
        \centering
        \includegraphics[width=\linewidth]{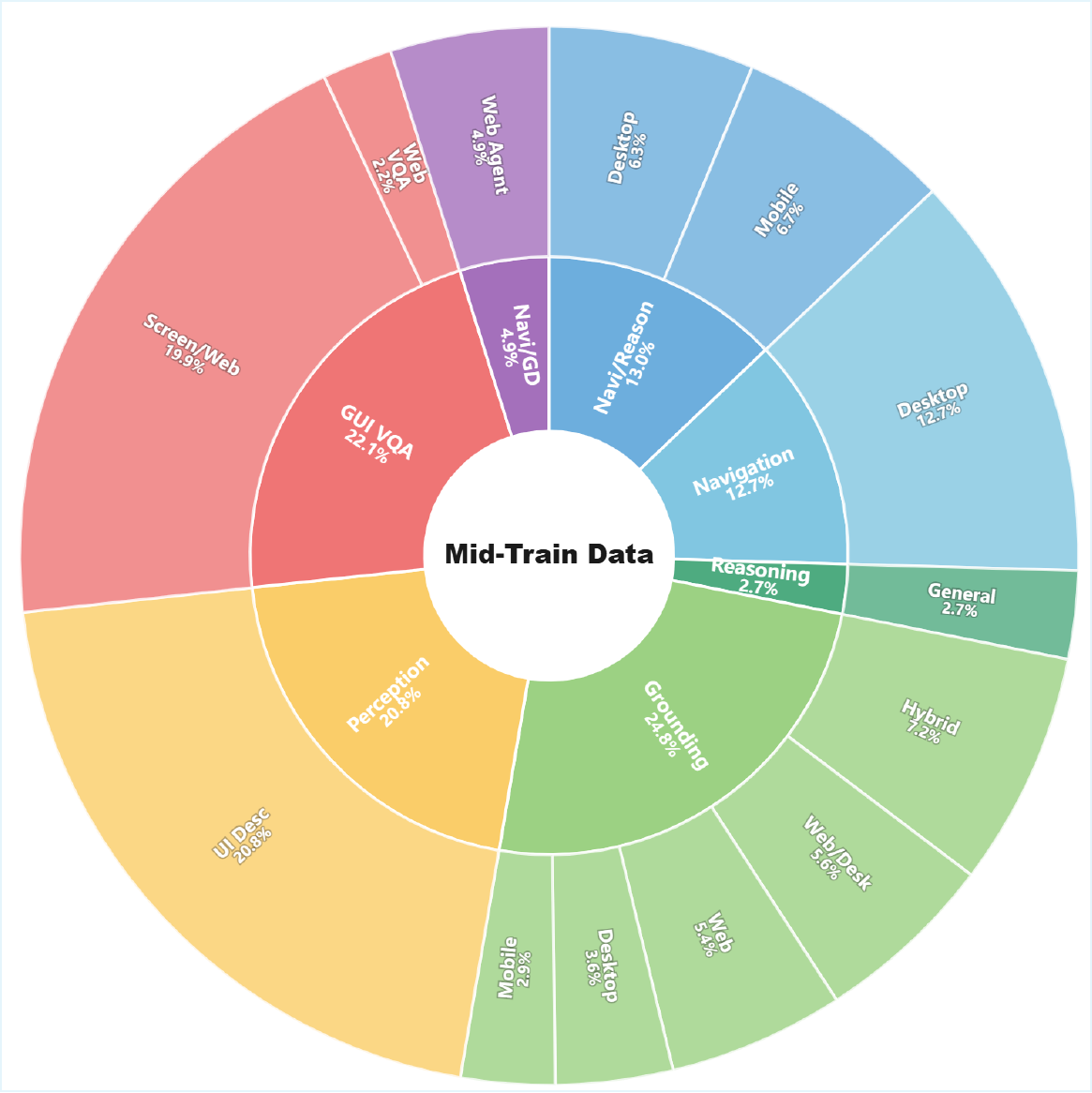} 
        \caption{\textbf{Mid-Training Corpus}}
        \label{fig:data_dist}
    \end{subfigure}
    \hfill
    \begin{subfigure}{0.6\textwidth}
        \centering
        \includegraphics[width=\linewidth]{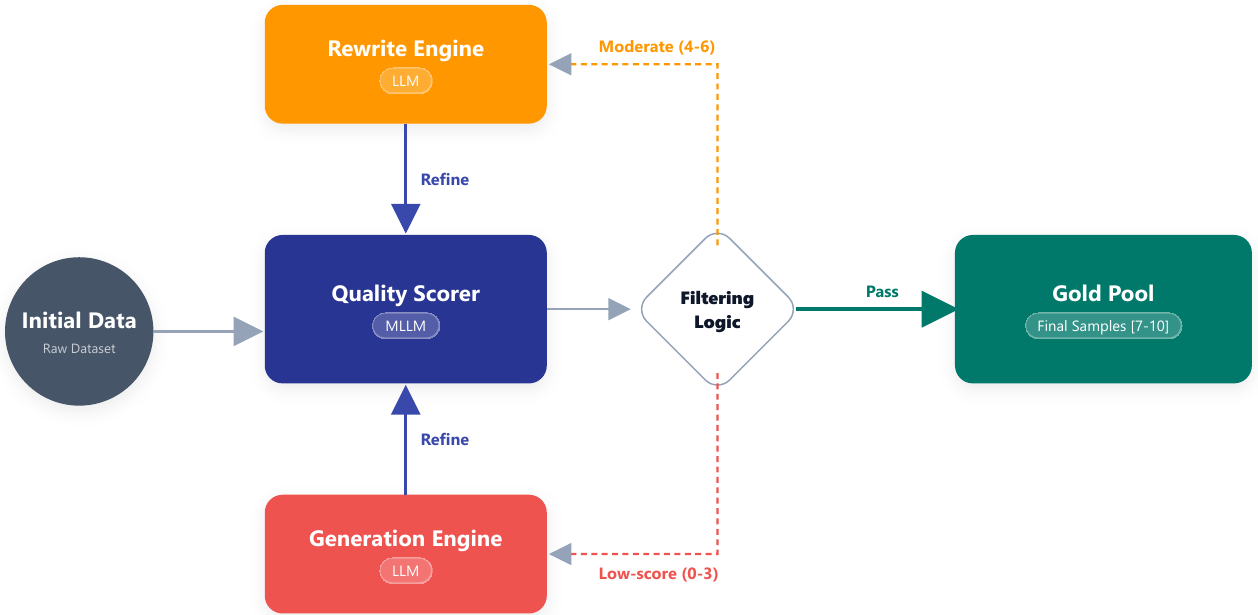} 
        \caption{\textbf{Iterative Data Refinement Pipeline}}
        \label{fig:idr_arch}
    \end{subfigure}
    \caption{(a) The inner part represents the functional task categories (\emph{e.g.}, GUI-VQA, Grounding, Perception), while the outer one details the distribution of specific data sources and target platforms (Web, Desktop, Mobile); (b) Iterative data refinement pipeline with teacher scoring, trace rewriting/reconstruction, and manual verification.}
\end{figure*}

\begin{figure}[htbp]
    \centering
    \includegraphics[width=0.9\textwidth]{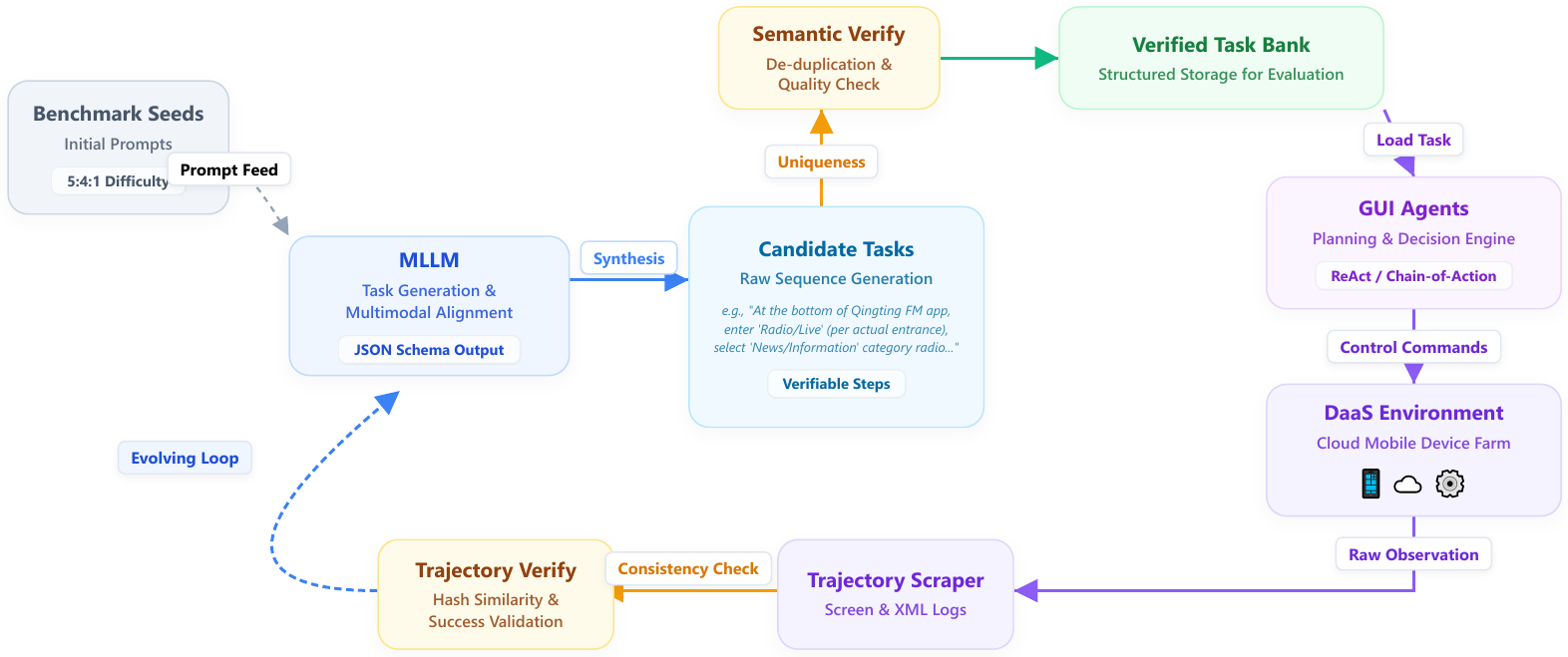}
    \caption{\textbf{Data generation loop via DaaS environment}. By iteratively performing this pipeline, the success rate of total trace generation raises from 17.9\% to over 70\%.}
    \label{fig:daas_feedback}
\end{figure}

To support the Mid-Training phase, we constructed a unified corpus by aggregating over 30 diverse sources, including Mind2Web~\cite{deng2023mind2web}, ShowUI~\cite{lin2024showuivisionlanguageactionmodelgui}, AITW~\cite{rawles2023androidinthewild} and so on. The hierarchical distribution of this corpus is detailed in Figure~\ref{fig:data_dist}. This 10B-token dataset is strategically stratified to ensure functional diversity: semantic perception (20.8\%) and GUI-VQA (22.1\%) provide the representational foundation, while grounding (24.8\%) and hybrid navigation-reasoning tasks ensure execution robustness.

Based on this unified corpus, the Mid-Training data supports four complementary supervision objectives covering perception, reasoning, and action alignment. Specifically, the dataset provides supervisions for:
\begin{itemize}
\item \textbf{Navigation \& Grounding:} Learning the precise alignment between natural language instructions and executable agent actions.
\item \textbf{Sequential Reasoning:} Generating Chain-of-Thought (CoT) traces that decompose high-level goals into intermediate steps.
\item \textbf{GUI-VQA:} Providing semantic interpretations of GUI components, functional descriptions, and layout logic.
\item \textbf{Fine-Grained Perception:} Capturing detailed attributes of visual elements, including icon recognition, widget state detection, and OCR-free dense captioning.
\end{itemize}


\subsubsection{Iterative Data Refinement}

With the large scale data we collected for Mid-Training, the next step is to clean and refine the low-quality navigation traces since some open-source datasets often contain noise that can limit performance gains.
To address this, we propose a teacher-based quality refinement module to rank, select, and rewrite the Mid-Training data. 
Specifically, we first utilize Qwen3-VL-235B-A22B~\cite{Qwen3-VL} to evaluate and rank the input data with numerical quality assessments as illustrated in Figure~\ref{fig:idr_arch} following the LLM-as-a-judge fashion. We chose this model because of its superior generalization and reasoning capabilities among open-source models. Mid-Training samples are scored from 0 to 10 based on action-visual alignment and task reachability according to our carefully designed prompts. 
Among the results, high-quality traces(score $\ge 7$) are retained in gold pool since the goal is accomplished; mid-quality traces(4-6) are routed to a rewriting model to refine the instruction according to the last state; and low-quality samples(0-3) are totally reconstructed or just discarded. 

By performing recursive refinements, the proportion of high-fidelity samples in our dataset increased from 69.7\% to 89.7\%. 
Finally, we conduct subsequent manual verification of sampled trajectories from the gold pool and ensure that the training signal remains both dense and accurate.

\subsubsection{Data Generation Loop}

To further improve robustness in real-world environments, we augment the Mid-Training corpus with interaction trajectories collected from real-device execution. Compared with static open-source datasets, real-device interaction data better captures execution failures, GUI dynamics, and environment-dependent behaviors that GUI agents must handle in practice.
We therefore build a data generation loop on top of our DaaS system (Section~\ref{sec:daas}). As illustrated in Figure~\ref{fig:daas_feedback}, an open-source MLLM first generates candidate task prompts from seed instructions. After semantic verification based on embedding similarity, valid and non-duplicate prompts are stored in a task bank and executed by GUI agents on cloud-hosted devices. The system then performs GUI trajectory scraping followed by multi-annotator verification to collect high-quality interaction trajectories.
A key feature of this pipeline is its iterative generation loop. Verified trajectories are fed back to the MLLM as in-context examples for subsequent task generation, enabling progressively more executable and realistic task prompts. As a result, the success rate of the trajectory generation pipeline improves from 17.9\% to over 70\% after several iterations.
In total, this real-device generation loop produces more than 30,000 verified interaction trajectories.

\subsection{Offline Reinforcement Learning}\label{sec:offline-rl}

\subsubsection{Grounding}


\textbf{Reward:} Following our previous work, the reward function is composed of two components formally: We first check whether the predicted answer string conforms to a predefined syntax as the format reward $R_{\text{format}}$. 
Next, given a screenshot and the instruction, the model must predict a center point that localizes the element. We use the commonly used point-in-box reward noted as $R_{\text{point-in-box}}$ to train the model.


Combining all components, the final action-wise reward is computed as:

\begin{equation}
R = R_{\text{format}} \cdot w_1 + R_{\text{poin-in-box}} \cdot w_2,
\end{equation}
where $w_1$ and $w_2$ control the relative importance of format correctness and location precision.

\textbf{Refusal Samples:} A major departure from UI-Venus-1.0 is the modification of our grounding prompts to incorporate what we define as refusal capability. Specifically, when faced with an instruction that refers to an element or icon, not present in the image, the model is trained to return a fixed output of $[-1, -1]$, refer to our prompt as shown in \ref{grounding_prompt}. Compared to models that strictly output coordinates regardless of the instruction's validity, this refusal-aware approach is significantly more intelligent and effectively mitigates hallucinations during user interactions.

Although the introduction of refusal prompts may lead to a marginal performance trade-off on benchmarks lacking refusal examples (such as ScreenSpot-Pro~\cite{li2024screenspot-pro}), UI-Venus-1.5 nevertheless maintains its state-of-the-art (SOTA) standing. Furthermore, on benchmarks that explicitly include refusal tasks—such as VenusBench-GD~\cite{zhou2025venusbench}, and OSWorld-G-Refine~\cite{xie2025scalingcomputerusegroundinguser}, our model achieves new SOTA results, particularly demonstrating superior accuracy on refusal-specific samples.



\subsubsection{Navigation}

In this section, we detail the Offline-RL formulation and empirical observations across Mobile and Web navigation tasks. 
In reinforcement learning, a well-designed and robust reward system is essential for stable policy optimization. 
Despite the differences between Mobile and Web platforms, 
GUI agents typically share a substantially overlapping action space
allowing for a unified reward design across both domains.

\textbf{Reward: }
To ensure the agent model generates structured, executable outputs, we build upon our prior UI-Venus-1.0 \cite{gu2025ui} framework and adopt a decoupled reward system comprising two primary components: (i) a format reward and (ii) an action reward.

\begin{itemize}
\item \textbf{Format reward }{$R_{\text{format}}$: }
This component ensures the agent model follows a predefined XML-based template. Specifically, the agent is rewarded for enclosing its response within \texttt{<think>}, \texttt{<action>}, and \texttt{<conclusion>} tags, which represent the reasoning process, the GUI action, and a concise action summary, respectively.
\item \textbf{Action reward }{$R_{\text{action}}$: }
The primary objective of the action reward is to encourage the model to predict valid and contextually appropriate actions for the current step. 
It is decomposed into two parts: an action-type reward $R_{\text{type}}$ and either a content-related reward $R_{\text{content}}$ or a coordinate-related reward $R_{\text{coord}}$. 
Specifically, $R_{\text{type}}$ is a binary reward determined by whether the predicted action type matches the ground-truth type. 
$R_{\text{content}}$ is applied to text-based actions and is computed as the token-level F1-score between the predicted and ground-truth content.
For $R_{\text{coord}}$, we adopt a hierarchical reward strategy that gradually relaxes the tolerance on coordinate errors, smoothing the reward scale and reducing the difficulty of policy optimization.
\end{itemize}

Compared to the UI-Venus-1.0 baseline, we enhance navigation performance in web environments by introducing several specialized GUI actions, such as \texttt{Hover} and \texttt{Hotkey}. 
Furthermore, we implement domain-specific constraints for the \texttt{Scroll} action to align with the different operational logics of each platform: 
in mobile tasks, the model must predict precise start and end coordinates, 
whereas in web tasks, it is only required to specify the scrolling direction. 

Overall, the total reward $R$ is formulated as:
\begin{equation}
R = w_1 \cdot R_{\text{format}} + w_2 \cdot R_{\text{action}},
\end{equation}
where $w_1$ and $w_2$ control the trade-off between the format reward and the action reward. 

\begin{figure}[htbp]
    \centering
    \begin{subfigure}{0.45\textwidth}
        \centering
        \includegraphics[width=\linewidth]{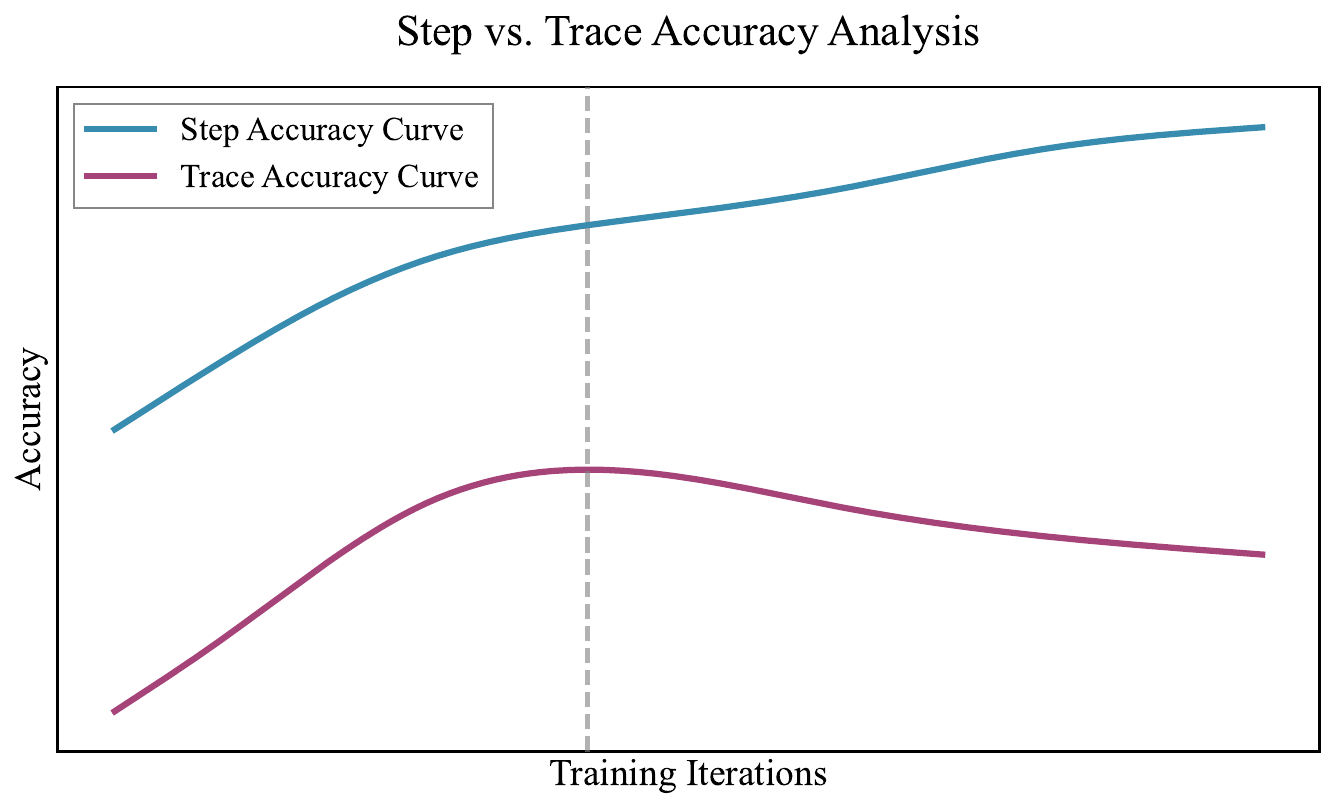} 
        \caption{\textbf{Mobile Scenario}}
        \label{fig:step_vs_trace_mobile}
    \end{subfigure}
    \hfill
    \begin{subfigure}{0.45\textwidth}
        \centering
        \includegraphics[width=\linewidth]{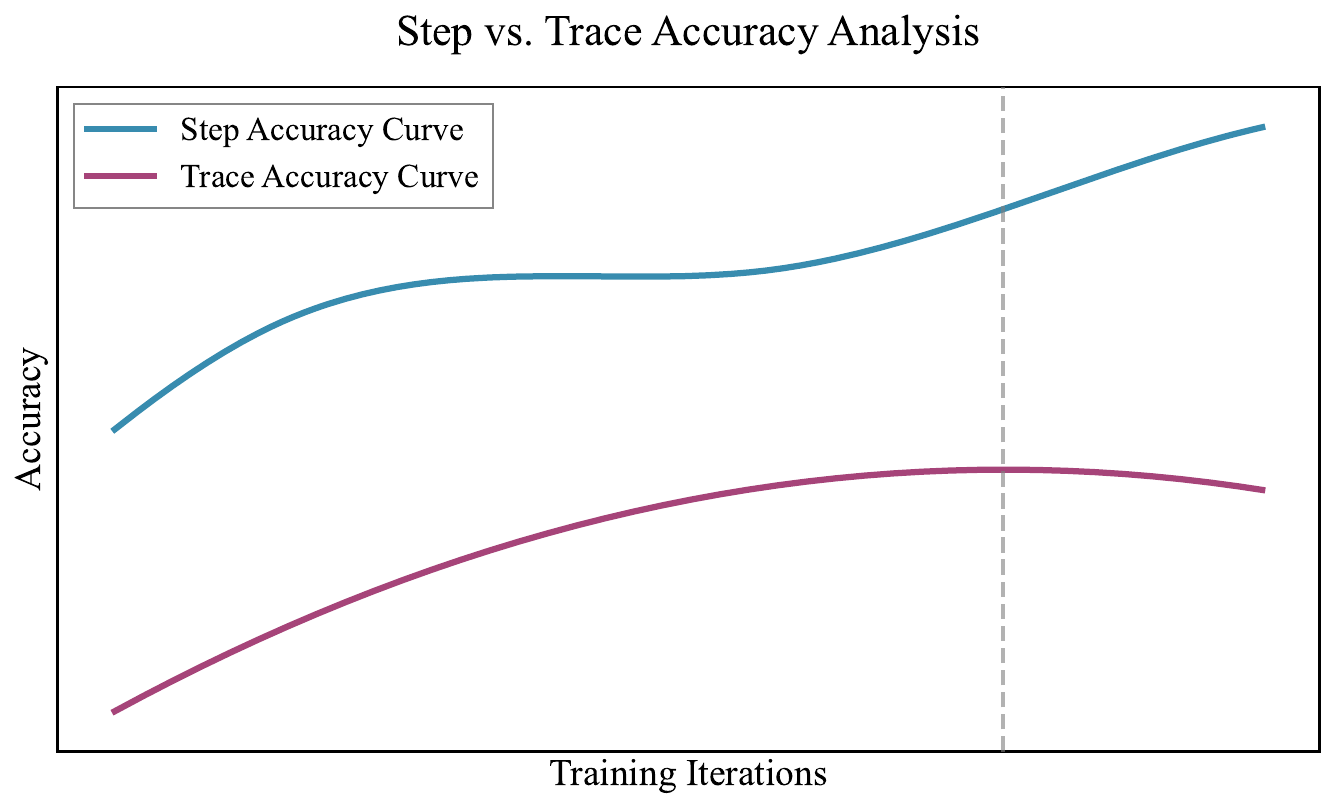} 
        \caption{\textbf{Web Scenario}}
        \label{fig:step_vs_trace_web}
    \end{subfigure}
    \caption{\textbf{Step vs. trace success rates} during Offline-RL training in (a) Mobile and (b) Web scenarios. We show the performance of training iterations on mobile and web benchmarks with two curves. The dashed line marks the peak trace-level success rate.}
    \label{fig:step_vs_trace}
\end{figure}

\textbf{Discrepancy Between Step and Trace Success Rates:}
We use the above reward system for stable Offline-RL. However, we observe a notable trend during Offline-RL training. 
As shown in Figure~\ref{fig:step_vs_trace}, while step-level success rates increase steadily, trace-level success rates eventually peak and then decline. 
We attribute this behavior to an inherent limitation of Offline-RL: 
step-level rewards only optimize individual actions and fail to guide the successful composition of a full multi-step trace.
To improve the deployable performance of the model, we therefore append an online reinforcement learning stage after the Offline-RL, explicitly optimizing for trace-level rewards and substantially enhancing 
the model's end-to-end task completion.

\subsection{Online Reinforcement Learning}\label{sec:online-rl}

\subsubsection{Motivation}
While SFT and Offline-RL provide a solid initialization for GUI agents, their effectiveness is inherently constrained by existing static datasets and predefined interaction distributions. 
In real-world GUI environments, agents are required to navigate dynamic GUI states, stochastic system behaviors, and extended decision-making horizons, where real-time feedback during execution is critical to performance.
Pure offline training is insufficient to address these challenges. 
Online Reinforcement Learning(Online-RL) responds to these limitations by enabling the agent to learn and adapt in real time through direct interaction with the environment. This allows for rich and immediate feedback and better handling of uncertainties in dynamic settings especially in long interaction sequences.
Consequently, policies trained solely offline often exhibit brittleness when deployed in novel GUI layouts or unexpected intermediate states. These limitations are further corroborated by the inconsistencies in step-trace accuracy observed in Section~\ref{sec:offline-rl}.


To address these limitations, we introduce online reinforcement learning as a complementary training paradigm. By enabling direct environmental interaction, Online-RL enables the agent to collect trajectories that reflect the actual deployment distribution and to iteratively refine and update its policy based on observed feedback. This leads to the design of a dedicated online learning framework, which comprises a robust execution infrastructure called DaaS (Section~\ref{sec:daas}), comprehensive task generation and reward formulations (Section~\ref{sec:olrl-reward}), and a stable RL training algorithm (Section~\ref{sec:olrl-training}).

\subsubsection{Device as a Service (DaaS)}\label{sec:daas}

\begin{figure}[htbp]
\centering
\includegraphics[width=0.9\textwidth]{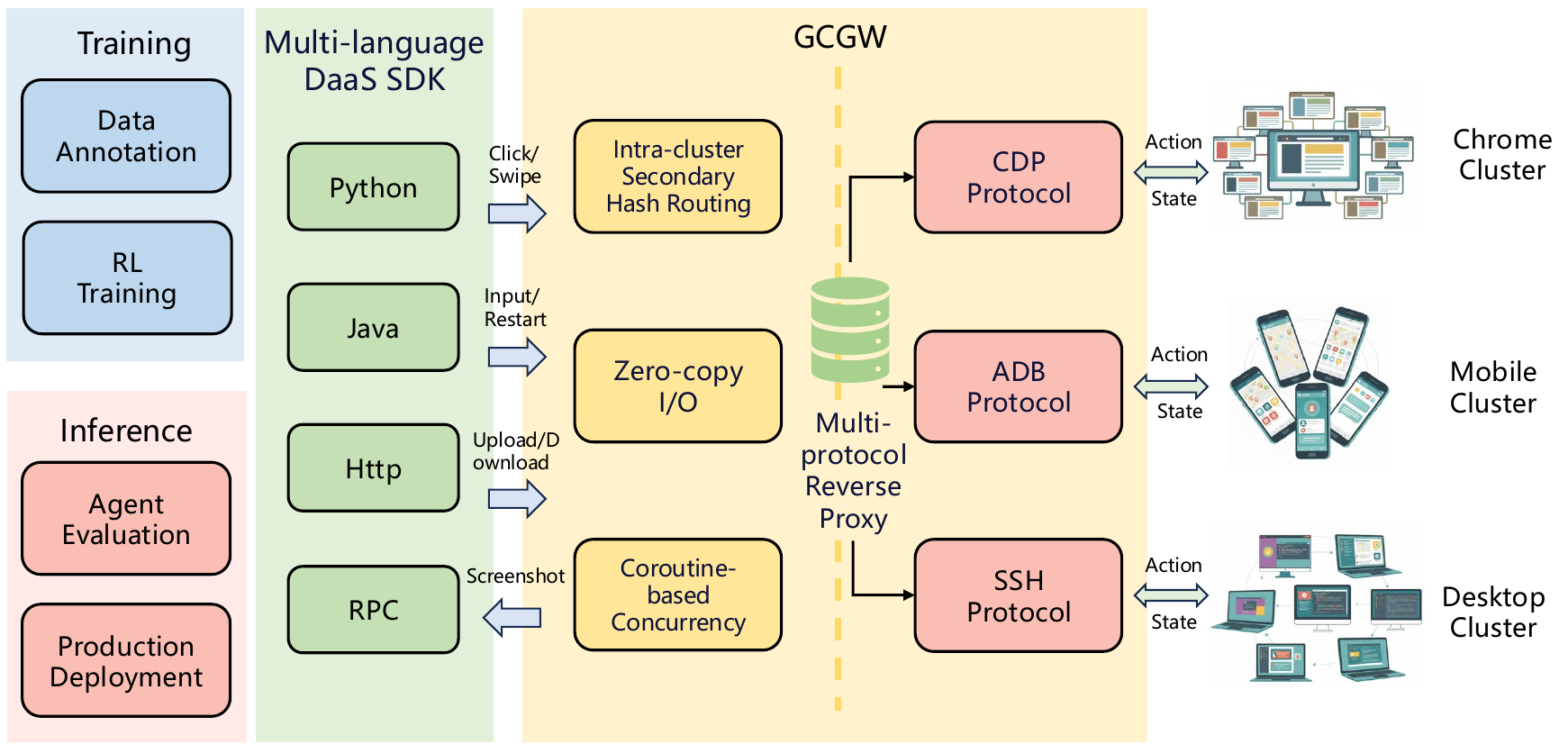}
\caption{\textbf{Architecture of the Unified Device-as-a-Service (DaaS) layer.} This framework bridges upstream tasks (Training and Inference) with downstream heterogeneous device clusters (Chrome, Mobile, and Desktop). The Multi-language DaaS SDK provides a unified abstraction for diverse interaction protocols, while the Group Control Gateway (GCGW) ensures high-performance and secure resource exposure through secondary hash routing, zero-copy I/O, and a multi-protocol reverse proxy.}
\label{fig:daas_framework}
\end{figure}

Training and deploying a GUI Agent capable of generalizing across heterogeneous devices imposes stringent demands on the underlying execution environment in terms of uniformity, extensibility, and performance. Unlike simulators designed for a single device category, such an environment must accommodate a wide variety of device types, offer a unified abstraction over diverse interaction protocols, and—under strict network-isolation policies, securely and efficiently expose large-scale device resources to upstream training and deployment frameworks. Therefore, we build a unified Device-as-a-Service (DaaS) layer (Figure~\ref{fig:daas_framework}) to meet these requirements. It consists of two core components: the Group Control Gateway (GCGW) and the Unified Client SDK, whose design and implementation are detailed next.

\textbf{Group Control Gateway (GCGW).} The GCGW serves as a high-performance centralized reverse proxy and the orchestration core of the DaaS layer. It abstracts heterogeneous control protocols across diverse platforms—including Android (ADB), Chrome (CDP), and Linux containers (SSH)—enabling extensible support through a protocol-centric abstraction.

To ensure system stability and performance under high-throughput workloads, the GCGW integrates several key architectural optimizations. First, to handle stateful protocols like ADB and CDP which rely on long-lived connections, the GCGW employs an internal secondary hash routing algorithm. This mechanism ensures that all requests for a specific device are consistently routed to the same gateway node, effectively preventing the ``$M\times N$ connection explosion'' problem($M$ gateway nodes and $N$ devices) and significantly reducing the connection overhead per node. To support this routing strategy without sacrificing performance, the gateway utilizes streaming transmission and zero-copy I/O for internal forwarding between nodes, ensuring near-zero additional latency. Furthermore, the entire GCGW architecture is built on a high-concurrency coroutine model. This design is specifically tailored for the ``high-concurrency, low-frequency'' access patterns of device operations, allowing the system to maintain hundreds of thousands of concurrent connections with minimal memory overhead.



\textbf{Unified Client SDK.} To further shield downstream users from the complexities of infrastructure management, a multi-language Unified Client SDK was developed that acts as a high-level API layer atop the GCGW. This SDK encapsulates several critical functions into a streamlined workflow. Specifically, it automates device lifecycle management, including device preemption, heartbeat maintenance, and resource release. 
Furthermore, it provides a unified semantic interaction interface that standardizes communication across various internal protocols. 
By abstracting these low-level operations, the SDK significantly lowers integration barriers, allowing downstream teams to concentrate on building and operating the end-to-end training pipeline for large-scale online reinforcement learning of GUI-specialized models, as well as the production deployment of those GUI-focused models—rather than dealing with protocol-specific technicalities.



By implementing these architectural optimizations, the DaaS layer achieves the following performance benchmarks:
\begin{itemize}
    \item \textbf{Scale and Throughput:} The system successfully integrates thousands of heterogeneous devices, establishing a resilient architecture that processes millions of operation requests daily.
    \item \textbf{Resource Allocation Efficiency:} Device resource allocation and scheduling achieve millisecond-level responsiveness, enabling rapid elastic provisioning.
    \item \textbf{High-Concurrency Support:} The system has successfully supported large-scale reinforcement learning training tasks involving hundreds to thousands of concurrent devices, which demonstrated superior stability and extensibility under high-load conditions.
\end{itemize}


\subsubsection{Task Formulation and Reward Design}\label{sec:olrl-reward}


Ahead of the training principles and pipeline of online-RL, we introduce task formulation and reward design which are fundamental to the success of online-RL. The quality, diversity, and calibrated difficulty of the input tasks define the potential ceiling of policy optimization, while the specialized reward function serves as the primary driver for training efficiency and stability. 

\textbf{Task Generation and Stratified Sampling:}
The performance ceiling of online-RL is fundamentally governed by the diversity and quality of its task pool $\mathcal{T}$. To this end, we employ a hybrid generation strategy combining static heuristics with dynamic evolution:
\begin{itemize}
    \item \textbf{Static Task Library via LLM:} For a predefined set of applications $\mathcal{A}$ and target websites $\mathcal{W}$, we leverage Large Language Models to extract functional maps and generate common tasks covering core user flows.
    
    \item \textbf{Dynamic Trajectory-based Generation via MLLM:} To capture long-tail interaction patterns, we randomly sample screenshots $s_t$ from offline trajectories $\tau$ and use MLLMs to infer plausible task query $q'$ from the observed state. To maintain task uniqueness, each newly generated goal is filtered by a deduplication function $\psi(\cdot)$:
    \begin{equation}
    \mathcal{T}_{new} = \{q' \mid \psi(q', \mathcal{T}_{pool}) < \epsilon\},
    \end{equation}
    where $\epsilon$ is a semantic similarity threshold that promotes uniform coverage of the task space.
    
    \item \textbf{Stratified Sampling:} We stratify tasks by difficulty, which correlates positively with the minimum steps to completion. Tasks are bucketed based on expected step count $N_{steps}$ into: \textbf{Easy} ($N_{steps} \le 10$), \textbf{Medium} ($10 < N_{steps} \le 20$), and \textbf{Hard} ($N_{steps} > 20$). During each training iteration, batches are sampled proportionally from these three buckets to support structured curriculum learning.
\end{itemize}

\textbf{Reward:}
To guide the agent effectively in complex GUI environments, we design a composite reward function $R$. For a given execution trajectory $\tau = (a_0, a_1, \dots, a_T)$ with $T$ steps, the total reward consists of a task completion reward $R_{comp}$, an action constraint penalty $R_{p}$, and a trace length decay coefficient $\eta \in (0, 1]$:
\begin{equation}
R(\tau) = \mathbbm{1}_{success} \cdot R_{comp} \cdot \eta^{\frac{T-T_{\text{min}}}{T_{\text{min}}}} + \sum_{t=0}^{T} R_{p}(a_t),
\end{equation}
where $T_{\text{min}}$ is the minimum number of steps to success among a group of trajectories collected during the online RL process. Specifically, to encourage the agent to learn the shortest operational path, we introduce a decay coefficient $\eta$. A larger step count $T$ results in a lower final reward, thereby suppressing redundant or circular actions during policy gradient optimization.
Another important factor of GUI Agent is to generate the correct actions in the predefined action pool, \emph{i.e.}, the agent's output must adhere to specific syntactic specifications. If the agent’s generated response cannot be recognized by the parser as a valid action, a negative penalty $\lambda$ is assigned at the current step. The penalty term is defined as:
\begin{equation}
    R_{p}(a_t) = 
    \begin{cases} 
    -\lambda, & \text{if } a_t \text{ is unparseable} \\
    0, & \text{otherwise}
    \end{cases}
\end{equation}
By incorporating $R_{p}$, we significantly reduce the number of invalid attempts during the online exploration phase, thereby improving sample efficiency.

After that, we implement a dual-track verification mechanism to determine task success ($\mathbbm{1}_{success}$):
\begin{itemize}
\item \textbf{Rule-based Verification:} For tasks with clear system-side outcomes (\emph{e.g.}, URL redirection, specific file generation, or system setting changes), success is verified deterministically by querying low-level system APIs.
\item \textbf{MLLM-as-a-Judge:} For semantically ambiguous tasks where visual feedback is prominent, the initial task $q$ and the final keyframe screenshot $s_i$ are fed into an MLLM to judge whether the logical intent has been satisfied.
\end{itemize}

\subsubsection{Training Algorithm}\label{sec:olrl-training}
We employ the Group Relative Policy Optimization (GRPO) algorithm. Unlike the conventional Actor-Critic framework, GRPO estimates relative advantages directly from a group of sampled trajectories, thereby bypassing the need for a separate value function network. This approach substantially reduces computational complexity and effectively addresses the convergence challenges posed by sparse reward signals in GUI-based tasks.

In each training step, for a task $q$ sampled from the task pool, the agent generates a group of $G$ complete interaction trajectories $\{\tau_i\}_{i=1}^G$ using the current policy $\pi_{\theta_{old}}$. The GRPO loss function $L_{GRPO}(\theta)$ is defined as follows:
\begin{equation}
L_{GRPO}(\theta) = - \frac{1}{G} \sum_{i=1}^{G} \frac{1}{|\tau_i|} \sum_{t=1}^{|\tau_i|} \min \left( r_{i,t}(\theta) \hat{A}_i, \text{clip}(r_{i,t}(\theta), 1-\epsilon, 1+\epsilon) \hat{A}_i \right),
\end{equation}

where $r_{i,t}(\theta) = \frac{\pi_{\theta}(a_{i,t} | s_{i,t}, q)}{\pi_{\theta_{old}}(a_{i,t} | s_{i,t}, q)}$ denotes the importance sampling ratio between the new and old policies at the action (or token) level.

\textbf{Trajectory-level Advantage Calculation and Assignment:}
Given the long execution horizons of GUI tasks and the difficulty in identifying critical actions, we forgo step-wise reward signals in favor of evaluating the overall quality of each complete trajectory $\tau_i$. The trajectory-level advantage $\hat{A}_i$ is derived by normalizing relative scores within the sampled group:

\begin{equation}
\hat{A}_i = \frac{R(\tau_i, q) - \text{mean}(\{R(\tau_j, q)\}_{j=1}^G)}{\text{std}(\{R(\tau_j, q)\}_{j=1}^G) + \epsilon},
\end{equation}

where $R(\tau_i, q)$ is the composite reward defined in Section ~\ref{sec:olrl-reward} (incorporating task success rewards and invalid action penalties). The calculated $\hat{A}_i$ is \textbf{uniformly assigned} to all action steps within the trajectory. 
By relying on competition within the sampled group, this approach filters out environmental stochasticity and supplies a stable credit assignment signal, thereby supporting stable policy updates in long-horizon decision-making.

\textbf{Training Stability and Exploration Enhancement:}
Rewards in GUI navigation are both sparse and costly to verify, which can lead to policy collapse during extended online training. To mitigate these issues, we implement two complementary regularization mechanisms:
\begin{itemize}
    \item \textbf{Adaptive KL Constraint:} To prevent the model from losing the fundamental GUI manipulation capabilities (\emph{e.g.}, basic clicking and swiping logic) acquired during SFT or Offline-RL, we incorporate a KL divergence penalty between the current policy $\pi_{\theta}$ and reference policy $\pi_{ref}$:
    \begin{equation}
    L_{KL}(\theta) = \beta \mathbb{D}_{KL}(\pi_{\theta} \| \pi_{ref}).
    \end{equation}
    To prevent the reference policy from becoming a stationary constraint that limits progress, we update it adaptively. When the current policy outperforms $\pi_{ref}$ on a held-out validation set by a margin $\delta$, we smoothly blend the two policies:
    \begin{equation}
    \pi_{ref} \leftarrow (1-\alpha)\pi_{ref} + \alpha\pi_{\theta}.
    \end{equation}
    This allows the KL penalty to dynamically track the policy’s improvement, preserving stability while allowing continued optimization.
    \item \textbf{Annealed Entropy Regularization:} After SFT or Offline-RL, policies often become overly deterministic, hampering exploration early in online training. We encourage action diversity via an entropy term:
    \begin{equation}
    L_{entropy}(\theta) = -\lambda_t \mathbb{H}(\pi_{\theta}(\cdot|s,q)).
    \end{equation}
    To avoid divergence from sustained high entropy, we anneal the coefficient $\lambda_t$exponentially with training steps $k$:
    $$\lambda_t = \lambda_0 \cdot \sigma^k, \quad \sigma \in (0, 1).$$
    By maintaining a high $\lambda_t$ to trigger exploration in the early stages and gradually reducing the weight to converge toward an optimal deterministic policy, this method effectively balances exploration and exploitation.
\end{itemize}

The final total optimization objective for the Online-RL stage is:
\begin{equation}
J(\theta) = L_{GRPO}(\theta) - L_{KL}(\theta) + L_{entropy}(\theta).
\end{equation}

\subsection{Model Merge}\label{sec:model-merge}

After offline-RL and online-RL phases, we implement a model merge~\cite{li2023deep,team2025tongyi,team2025every1} strategy to consolidate the specialized expertise of our task-specific models into a single, unified GUI Agent. 
This approach leverages the principle that models fine-tuned from a common foundational ancestor occupy a shared parameter space, allowing for their weights to be integrated through strategic interpolation. Specifically, we explored two distinct merging paradigms as follows: Linear Merge~\cite{li2023deep} and TIES-Merge~\cite{yadav2023ties}. 

\textbf{Linear Merge:} We take the checkpoints optimized for grounding, web, and mobile navigation, and synthesize them into a global parameter set $\theta_{linear}$ using a weighted combination:
\begin{equation}
\theta_{linear} = \sum_{i=1}^{3} w_i \cdot \theta_{i}, \quad \text{subject to} \quad \sum_{i=1}^{3} w_i = 1,
\end{equation}
where $\theta_{i}$ denotes the parameters of the $i$-th specialized model and $w_i$ represents its relative importance in the fusion process.


\textbf{TIES-Merge:} Compared to standard linear merging, TIES-Merge reduces parameter interference through two key steps. First, it calculates task vectors (differences between fine-tuned models and the base model) and prunes low-magnitude updates to retain only the most significant changes. Second, before merging, it elects a dominant sign direction for each parameter and aggregates only updates aligned with that direction. By pruning noisy updates and resolving sign conflicts, TIES-Merge achieves markedly lower performance regression than simple interpolation. 


\textbf{Performance comparison:} According to our experiments in Section~\ref{sec:ablation}, \emph{TIES-Merge always performs better than Linear Merge}. Take our UI-Venus-1.5-30B-A3B for example, it achieves 71.0\% and 75.5\% accuracy on ScreenSpot-Pro and AndroidWorld before Model Merging, respectively. By adopting Linear Merge, the performances drop 2.9\%$\downarrow$ and 2.3\%$\downarrow$. Refer to the experiment results of TIES-Merging in Table~\ref{tab:ablation}, \emph{i.e.} 69.6\%(1.4\%$\downarrow$) and 77.6\%(2.1\%$\uparrow$), it significantly outperforms Linear Merge in the context of cross-task fusion. Note that although model merging may lead to performance drop of some tasks compared to domain-specific models, the resulting UI-Venus-1.5 achieves a harmonious balance across all three domains, delivering robust performance without the computational overhead of training a multi-task model from scratch.

\section{Experiments}

\begin{table*}[ht]
	\centering
	\footnotesize
	\setlength{\tabcolsep}{0pt}
	\begin{tabular*}{\textwidth}{@{\extracolsep{\fill}}l *{7}{c}}
		\toprule
		\multirow{2}{*}{\textbf{Models}} &
		\multicolumn{7}{c}{\textbf{Grounding Benchmarks}} \\
		\cmidrule(lr){2-8}
		& \textbf{VenusBench-GD} & \textbf{ScreenSpot-Pro} & \textbf{ScreenSpot-V2} & \textbf{MMbench} & \textbf{OSworld-G-R} & \textbf{OSworld-G} & \textbf{UI-Vision} \\
		\midrule
		\rowcolor{gray!15}
		\multicolumn{8}{l}{\textit{General VLMs}} \\
		Seed1.8~\citep{seed1.8}                         & - & 64.3 & - & - & - & - & - \\
		Qwen3-VL-2B*~\citep{Qwen3-VL}              & 45.2 & 40.6 & 85.6 & 69.5 & 60.6 & 47.7 & 13.1 \\
		Qwen3-VL-8B*~\citep{Qwen3-VL}             & 55.1 & 52.7 & 92.1 & 81.4 & 67.0 & 57.5 & 21.9 \\
		Qwen3-VL-30B-A3B*~\citep{Qwen3-VL}          & 52.4 & 53.7 & 91.7 & 83.7 & 69.3 & 61.2 & 25.6 \\
		\midrule
		\rowcolor{gray!15}
		\multicolumn{8}{l}{\textit{GUI-specific Models}} \\
		OpenCUA-7B~\citep{wang2025opencuaopenfoundationscomputeruse}                     & 48.2 & 50.0 & 92.3 & - & - & 55.3 & 29.7 \\
		OpenCUA-32B~\citep{wang2025opencuaopenfoundationscomputeruse}                     & 50.1 & 55.3 & 93.4 & - & 70.2 & 59.6 & 33.3 \\
		OpenCUA-72B~\citep{wang2025opencuaopenfoundationscomputeruse}                     & - & 60.8 & 92.9 & - & - & - & 37.3 \\
		GTA1-7B~\citep{yang2025gta1guitesttimescaling}                          & 46.4 & 50.1 & 92.4 & - & 67.7 & 60.1 & - \\
		GTA1-32B~\citep{yang2025gta1guitesttimescaling}                         & 58.8 & 63.6 & 95.2 & - & 72.2 & 65.2 & - \\
		GUI-Owl-7B~\citep{ye2025mobile}                       & - & 54.9 & 92.8 & 80.5 & - & 55.9 & - \\
		GUI-Owl-32B~\citep{ye2025mobile}                      & - & 58.0 & 93.1 & 83.0 & - & 58.0 & - \\
		UI-TARS-1.5-7B~\citep{ui-tars-15-seed}                   & 40.7 & 35.7 & 91.6 & 64.3 & 64.2 & 52.8 & 22.3 \\
		UI-Venus-1.0-7B~\citep{gu2025ui}                  & 49.0 & 50.8 & 94.1 & 79.9 & 61.7 & 54.6 & 26.5 \\
		UI-Venus-1.0-72B~\citep{gu2025ui}                 & {70.2} & {61.9} & {95.3} & {86.3} & 69.5 & {62.2} & {36.8} \\
		Holo2-8B~\citep{hai2025holo2modelfamily}                         & 56.4* & 58.9 & 93.2 & 84.5* & 70.1 & 63.5* & 35.1* \\
		Holo2-30B-A3B~\citep{hai2025holo2modelfamily}                    & 59.5* & 66.1 & 94.9 & 86.8* & 76.1 & 65.2* & 40.9* \\
		Step-GUI-4B~\citep{yan2025step}                      & 54.6* & 60.0 & 93.6 & 84.0 & 66.9 & 60.5* & 30.0* \\
		Step-GUI-8B~\citep{yan2025step}                      & - & 62.6 & 95.1 & 85.6 & 70.0 & - & - \\
		MAI-UI-2B~\citep{zhou2025mai}                        & 55.4* & 57.4 & 92.5 & 82.6 & 63.5 & 52.0 & 30.3 \\
		MAI-UI-8B~\citep{zhou2025mai}                        & 65.2* & 65.8 & 95.2 & \underline{88.8} & 68.6 & 60.1 & 40.7 \\
		MAI-UI-32B~\citep{zhou2025mai}                       & - & {67.9} & \textbf{96.5} & \textbf{91.3} & \underline{73.9} & {67.6} & \underline{47.1} \\
		\midrule
		\rowcolor{gray!15}
		\multicolumn{8}{l}{\textit{Ours}} \\
		\textbf{UI-Venus-1.5-2B}                       & 67.3 & 57.7 & 92.8 & 80.3 & 65.6 & 59.4 & 44.8 \\
		\textbf{UI-Venus-1.5-8B}                       & 72.3 & 68.4 & 95.9 & 88.1 & 74.1 & 69.7 & 46.5  \\
		\textbf{UI-Venus-1.5-30B-A3B}                      & \textbf{75.0} & \textbf{69.6} & \underline{96.2} & {88.6} & \textbf{76.4} & \textbf{70.6} & \textbf{54.7} \\ \bottomrule
	\end{tabular*}
	\vspace{0.4em}
	\caption{Performance comparison on various \textbf{Grounding Benchmarks}. For each benchmark, the best and second-best performing models are indicated in \textbf{bold} and \underline{underlined}, respectively. ``*'' indicates results that may require verification with original sources.}
	\label{tab:grounding_benchmarks}
\end{table*}

\subsection{Experimental Setup}

\subsubsection{Implementation details}

UI-Venus-1.5 incorporates the Qwen3-VL~\cite{Qwen3-VL} architecture as its core backbone, leveraging its advanced multimodal processing capabilities to interpret complex visual interfaces. Note that we map all spatial targets into a normalized $[0,1000]$ coordinate space following Qwen3-VL and unifies diverse events into a shared set of action space as shown in Table~\ref{tab:action_space}.

Note that methods marked with * indicate our own reproductions with official prompts, some of which achieve superior performance compared to their original reported results. Moreover, all methods evaluated in this section \textbf{follow the end-to-end design} without any \emph{zoom-in or agent-framework} strategies except specially mentioned.

\subsubsection{Benchmarks}

\textbf{Grounding:} We evaluate the model's grounding capabilities with seven complementary benchmarks: VenusBench-GD~\cite{zhou2025venusbench} for its assessment of high-level reasoning and refusal capabilities, ScreenSpot-Pro~\cite{wu2024osatlasfoundationactionmodel} focuses on high-resolution, fine-grained professional layouts, UI-Vision~\cite{nayak2025ui} assesses reasoning abilities (\emph{e.g.}, spatial and functional) in diverse applications, MMBench-GUI L2~\cite{wang2025mmbench} evaluates hierarchical-instruction following and compositional reasoning, OSWorldG and OSWorld-G Refine~\cite{xie2025scalingcomputerusegroundinguser} jointly assess comprehensive skills such as layout understanding, widget matching, and fine-grained manipulation, and ScreenSpot-V2~\cite{wu2024osatlasfoundationactionmodel} serves to broaden coverage across different operating systems.

\textbf{Navigation:} We first evaluated UI-Venus-1.5 on two widely-adopted online mobile benchmarks: AndroidWorld~\cite{androidworld} and Androidlab~\cite{xu2025androidlab}, based on a live Android emulator with various applications and tasks. In addition, we also performed experiments on VenusBench-Mobile, a more challenging benchmark whose tasks are more related to real-world human applications.
To further validate the cross-domain versatility of UI-Venus-1.5, we extended our evaluation to real-world web environments. We evaluate our model on a representative subset of WebVoyager~\cite{he2024webvoyager}, as the full benchmark is time-consuming to evaluate and some time-sensitive tasks are no longer valid.


\subsection{Main Results}
\subsubsection{Grounding Benchmarks}

In the experiments, we compare UI-Venus-1.5 models against various state-of-the-art baselines across two different model categories:
\textbf{(1) General VLMs}: Seed1.8~\cite{seed1.8} and widely used Qwen3-VL~\cite{Qwen3-VL}.
\textbf{(2) GUI-specific Models}: OpenCUA~\cite{wang2025opencuaopenfoundationscomputeruse}, UI-TARS-1.5~\cite{ui-tars-15-seed}, GTA1~\cite{yang2025gta1guitesttimescaling}, Gui-Owl~\cite{ye2025mobile}, Holo2~\cite{hai2025holo2modelfamily}, Step-GUI~\cite{yan2025step} and MAI-UI~\cite{zhou2025mai}.

\noindent \textbf{VenusBench-GD.} VenusBench-GD is a comprehensive grounding benchmark spanning web, desktop, and mobile UIs, covering both basic localization and advanced, reasoning-heavy cases, and further includes \emph{refusal grounding} to test whether agents can correctly reject infeasible instructions. As shown in Table~\ref{tab:grounding_benchmarks}, UI-Venus-1.5 achieves strong and scalable performance, with our 30B-A3B model reaching 75.0\% and outperforming competitive GUI-specialized baselines.

\noindent \textbf{ScreenSpot-Pro.} ScreenSpot-Pro focuses on high-resolution professional software interfaces (\emph{e.g.}, CAD, development tools, creative suites, and office applications), where dense layouts and small icons make fine-grained grounding particularly challenging. In Table~\ref{tab:grounding_benchmarks}, UI-Venus-1.5-30B-A3B achieves the best overall accuracy at 69.6\%, exceeding the strongest reported baseline MAI-UI-32B (67.9\%) and showing clear gains over smaller UI-Venus-1.5 variants (57.7\% for 2B, 68.4\% for 8B).

\noindent \textbf{ScreenSpot-V2.} ScreenSpot-V2 is a broad cross-platform grounding benchmark covering mobile, web, and desktop UIs with both text and icon/widget targets, reflecting everyday GUI interaction scenarios. As shown in Table~\ref{tab:grounding_benchmarks}, UI-Venus-1.5 achieves strong performance, with 30B-A3B reaching 96.2\% (second-best, 0.3\% behind MAI-UI-32B) and 8B reaching 95.9\%, demonstrating robust generalization even in a near-saturated benchmark.

\noindent \textbf{MMBench-GUI L2.} MMBench-GUI L2 evaluates instruction-following grounding with both \emph{Basic} (low-level attributes) and \emph{Advanced} (goal-oriented, compositional) instructions, testing a model's ability to align implicit user intent with the correct UI element. In Table~\ref{tab:grounding_benchmarks}, UI-Venus-1.5 remains competitive, reaching 88.6\% with 30B-A3B.

\noindent \textbf{OSWorld-G.} OSWorld-G(and its refined split OSWorld-G-R) contains fine-grained desktop grounding tasks that require diverse skills such as text matching, widget recognition, layout understanding, and precise manipulation. Table~\ref{tab:grounding_benchmarks} shows UI-Venus-1.5-30B-A3B achieves state-of-the-art performance on both settings, reaching 76.4\% on OSWorld-G-R and 70.6\% on OSWorld-G, surpassing strong baselines such as MAI-UI-32B (73.9\% / 67.6\%) and GTA1-32B (72.2\% / 65.2\%).

\noindent \textbf{UI-Vision.} UI-Vision is a license-permissive benchmark for desktop GUI grounding that emphasizes real-world applications and fine-grained reasoning (\emph{e.g.}, spatial and functional understanding). As shown in Table~\ref{tab:grounding_benchmarks}, UI-Venus-1.5-30B-A3B achieves the strongest result at 54.7\%, substantially outperforming prior GUI-specific baselines (\emph{e.g.}, MAI-UI-32B at 47.1\%).


In summary, our evaluation reveals several key insights: 
\begin{itemize}
    \item \textbf{Strong Overall Grounding:} UI-Venus-1.5-30B-A3B achieves state-of-the-art results on most benchmarks, leading VenusBench-GD (75.0\%), ScreenSpot-Pro (69.6\%), OSWorld-G-R (76.4\%), OSWorld-G (70.6\%), and UI-Vision (54.7\%), while remaining highly competitive on ScreenSpot-V2 (96.2\%, second-best, 0.3\% behind MAI-UI-32B).
    \item \textbf{Consistent Scaling Gains:} Increasing model scale yields steady improvements across all benchmarks (\emph{e.g.}, ScreenSpot-Pro: 57.7\%$\rightarrow$68.4\%$\rightarrow$69.6\% for 2B/8B/30B-A3B).
    \item \textbf{Broad Generalization Across Tasks:} UI-Venus shows robust performance on diverse grounding settings, from refusal-aware evaluation in VenusBench-GD to fine-grained professional UI layouts in ScreenSpot-Pro, and remains competitive on instruction-intensive MMBench (88.6\%).
\end{itemize}

\begin{table}[ht]
    \centering
    \footnotesize
    \setlength{\tabcolsep}{1.5pt} 
    
    \begin{minipage}[t]{0.49\textwidth}
        \centering
        \begin{tabular}{lcc}
        \toprule
        \textbf{Models} &\textbf{Params.}   & \textbf{Success Rate}                  \\ 
        \midrule
        \rowcolor{gray!15}
        \multicolumn{3}{l}{\textit{General VLMs}} \\
        Qwen3-VL-2B~\citep{Qwen3-VL}                       &  2B    & 36.4                                   \\ 
        Qwen3-VL-8B ~\citep{Qwen3-VL}                       &  8B    & 47.6                                  \\ 
        Qwen3-VL-30B-A3B ~\citep{Qwen3-VL}                       &  30B    &  54.3                        \\ 
        GLM-4.6V~\citep{vteam2025glm45vglm41vthinkingversatilemultimodal}       &   106B        & 57.0  \\
        Gemini-2.5-Pro~\citep{comanici2025gemini}  & -   &                     69.7       \\ 
        Seed1.8~\citep{seed1.8}  & -   &                     70.7       \\ 
        \midrule
        \rowcolor{gray!15}
        \multicolumn{3}{l}{\textit{GUI-specific Models}} \\
        UI-TARS-1.5-7B~\citep{ui-tars-15-seed}                       &  7B      & 30.0                       \\ 
        GUI-Owl-7B ~\citep{ye2025mobile}                       &  7B      &  66.4                      \\ 
        UI-TARS-72B~\citep{qin2025uitarspioneeringautomatedgui}                &   72B     & 46.6  \\
        UI-Venus-1.0-7B~\citep{gu2025ui}                  &  7B & 49.1               \\ 
        UI-Venus-1.0-72B~\citep{gu2025ui}                 &  72B      & 65.9                \\
        Step-GUI-4B~\citep{yan2025step}  &  4B      & 63.9                \\ 
        Step-GUI-8B~\citep{yan2025step}                       &  8B      &                 67.7       \\ 
        Holo2-8B~\citep{hai2025holo2modelfamily}                       &  8B      & 60.4                       \\ 
        Holo2-30B-A3B~\citep{hai2025holo2modelfamily}                       &  30B      &  71.6                       \\ 
        MAI-UI-2B~\citep{zhou2025mai}                       &  2B      &    49.1                   \\ 
        MAI-UI-8B~\citep{zhou2025mai}                       &  8B      &  70.7                   \\ 
        MAI-UI-32B~\citep{zhou2025mai}                       &  32B      &  73.3                   \\ 
        \midrule
        \rowcolor{gray!15}
        \multicolumn{3}{l}{\textit{Ours}} \\
        \textbf{UI-Venus-1.5-2B}               &  2B &      55.6        \\ 
        \textbf{UI-Venus-1.5-8B}               &  8B &      73.7        \\ 
        \textbf{UI-Venus-1.5-30B-A3B}               &  30B      &     \textbf{77.6} \\
        \bottomrule
        \end{tabular}
        \vspace{0.4em}
        \caption{Performance comparison on \textbf{AndroidWorld} for end-to-end models. 
        }
        \label{tab:Androidworld}
    \end{minipage}
    \hfill 
    \begin{minipage}[t]{0.49\textwidth}
        \centering
        \begin{tabular}{lcc}
        \toprule
        \textbf{Models} &\textbf{Params.}   & \textbf{Success Rate}                  \\ 
        \midrule
        \rowcolor{gray!15}
        \multicolumn{3}{l}{\textit{General VLMs}} \\
        Gemini-1.5-Pro~\citep{team2024gemini}                       &  -    & 16.7                                   \\ 
        GLM4-9B-ft~\citep{glm2024chatglm}                       &  9B    &  21.0                         \\ 
        LLaMA3.1-ft~\citep{grattafiori2024llama}                       &  8B    & 23.9                                  \\ 
        GPT-4o~\citep{hurst2024gpt}       &   -        & 31.2  \\
        Qwen3-VL-2B\textsuperscript{*}~\citep{Qwen3-VL}    &  2B    & 33.3                                   \\ 
        Qwen3-VL-8B\textsuperscript{*}~\citep{Qwen3-VL}   &  8B    &       43.5                  \\ 
        Qwen3-VL-30B-A3B\textsuperscript{*}~\citep{Qwen3-VL}   &  30B &       42.0                  \\ 
        \midrule
        \rowcolor{gray!15}
        \multicolumn{3}{l}{\textit{GUI-specific Models\&Frameworks}} \\
        V-Droid~\citep{dai2025advancing}   & 8B   & 38.3            \\
        UI-Genie~\citep{xiao2025ui} & 72B & 41.2  \\
        MobileUse~\citep{li2025mobileuse}  & 72B   & 44.2     \\
        UI-Venus-1.0-7B~\citep{gu2025ui} & 7B   &   41.3  \\
        UI-Venus-1.0-72B~\citep{gu2025ui} & 72B   &  49.3    \\
        AutoGLM-Mobile~\citep{liu2024autoglm}  & 9B   & 46.8     \\
        AutoGLM-Multilingual~\citep{liu2024autoglm}  & 9B   & 47.7     \\
        Step-GUI-4B\textsuperscript{*}~\citep{yan2025step}   &  4B      &  47.8               \\ 
        \midrule
        \rowcolor{gray!15}
        \multicolumn{3}{l}{\textit{Ours}} \\
        \textbf{UI-Venus-1.5-2B}               &  2B &  36.2/44.2$^{\dagger}$            \\ 
        \textbf{UI-Venus-1.5-8B}               &  8B &  \textbf{55.1}/68.1$^{\dagger}$            \\ 
        \textbf{UI-Venus-1.5-30B-A3B}          &  30B & 52.9/68.1$^{\dagger}$             \\ 
        \bottomrule
        \end{tabular}
        \vspace{0.4em}
        \caption{Performance comparison on \textbf{AndroidLab}. Note that \textsuperscript{*} indicates that the results are evaluated by us; \textsuperscript{$\dagger$} denotes results that have been manually verified by humans.
        }
        \label{tab:AndroidLab}
    \end{minipage}
\end{table}

\subsubsection{Navigation Benchmarks}
In addition to evaluating GUI grounding capabilities, we further assess the UI-Venus-1.5 series on four navigation benchmarks spanning both mobile and web environments, including Android World~\cite{androidworld}, Android Lab~\cite{xu2025androidlab}, VenusBench-Mobile and WebVoyager~\cite{he2024webvoyager}. These benchmarks are challenging, fully dynamic online suites that require GUI agents to perform multi-turn adaptive perception, reasoning, and action in evolving environments, thus providing a more reliable assessment of the model's navigation capabilities.



\noindent \textbf{Android World.} AndroidWorld is a comprehensive online evaluation environment for GUI agents. It includes 116 programmatic tasks across 20 real-world Android applications and is widely used as a benchmark for assessing GUI agent performance. As shown in Table~\ref{tab:Androidworld}, the Venus-1.5 model family achieves state-of-the-art (SOTA) performance among models of comparable scale. Specifically, our 2B~/~8B~/~30B-A3B variants reach accuracies of 55.6\%~/~73.7\%~/~77.6\%, outperforming existing domain-specific and general-purpose baselines. Relative to the strongest existing contender, MAI-UI-32B, UI-Venus-1.5-30B-A3B secures an absolute margin of 4.3\%.

\noindent \textbf{Android Lab.} Android Lab is another dynamic, online evaluation benchmark comprising 138 tasks across 9 Android applications. Our model uses only raw screen screenshots as input, yet it outperforms both a range of GUI-specialized models and general-purpose models, some of which take both screenshots and XML information as inputs. As shown in Table~\ref{tab:AndroidLab}, our UI-Venus-1.5 series models (2B, 8B, and 30A3B) achieve 36.2\%, 55.1\%, and 52.9\% on AndroidLab, respectively. It is worth noting that, due to bugs in the official AndroidLab evaluation code, we additionally report human-verified results (marked with a $\dagger$) in the Table~\ref{tab:AndroidLab}. The corrected results show that our UI-Venus-1.5-30A3B model does not exhibit any performance degradation on AndroidLab compared with the  UI-Venus-1.5-8B model (68.1\% vs.\ 68.1\%). Compared with our UI-Venus-1.0-72B model, the best model in the 1.5 series yields up to 5.8\% improvement. Notably, even UI-Venus-1.5-8B significantly outperforms other state-of-the-art models.

\begin{table}[ht]
    \centering
    \footnotesize
    \setlength{\tabcolsep}{1.5pt} 
    
    \begin{minipage}[t]{0.49\textwidth}
        \centering
        \footnotesize
        \begin{tabular}{lcc}
        \toprule
        \textbf{Models} &\textbf{Params.}   & \textbf{Success Rate}                  \\ 
        \midrule
        \rowcolor{gray!15}
        \multicolumn{3}{l}{\textit{General VLMs}} \\
        Qwen3-VL-8B ~\citep{Qwen3-VL}                      &  8B    &    6.7                      \\ 
        Qwen3-VL-30B-A3B ~\citep{Qwen3-VL}                      &  30B    &  8.7                        \\ 
        \midrule
        \rowcolor{gray!15}
        \multicolumn{3}{l}{\textit{GUI-specific Models}} \\ 
        GUI-Owl-7B ~\citep{ye2025mobile}                      &  7B     &    6.7                  \\ 
        UI-Venus-1.0-7B~\citep{gu2025ui}                &  7B &   8.1          \\ 
        UI-Venus-1.0-72B~\citep{gu2025ui}              &  72B &     15.4        \\ 
        Step-GUI-4B~\citep{yan2025step}  &  4B     &        8.0      \\ 
        MAI-UI-2B~\citep{zhou2025mai}                      &  2B     &     6.7                 \\ 
        MAI-UI-8B~\citep{zhou2025mai}                      &  8B     &     12.7                 \\ 
        \midrule
        \rowcolor{gray!15}
        \multicolumn{3}{l}{\textit{Ours}} \\
        \textbf{UI-Venus-1.5-2B}              &  2B &      8.7       \\ 
        \textbf{UI-Venus-1.5-8B}              &  8B &     16.1        \\ 
        \textbf{UI-Venus-1.5-30B-A3B}              &  30B     &     \textbf{21.5}           \\ 
        \bottomrule
        \end{tabular}
        \vspace{0.4em}
        \caption{Performance comparison on \textbf{VenusBench-Mobile} for end-to-end models. Our UI-Venus-1.5 achieves state-of-the-art performance on this challenging benchmark.}
        \label{tab:VenusBench-Mobile}
    \end{minipage}
    \hfill 
    \begin{minipage}[t]{0.49\textwidth}
        \centering
        \footnotesize
        \begin{tabular}{lcc}
        \toprule
        \textbf{Models} &\textbf{Params.}   & \textbf{Success Rate}                  \\ 
        \midrule
        \rowcolor{gray!15}
        \multicolumn{3}{l}{\textit{General VLMs}} \\
        GPT-4o~\citep{hurst2024gpt}   & -   & 55.5                     \\ 
        Claude-3.7~\citep{anthropic2025claude37}  & -   &      84.1                \\ 
        Qwen3-VL-2B\textsuperscript{*}~\citep{Qwen3-VL}                      &  2B    &     35.2                       \\ 
        Qwen3-VL-8B\textsuperscript{*}~\citep{Qwen3-VL}                      &  8B    &     45.2                     \\ 
        Qwen3-VL-30B-A3B\textsuperscript{*}~\citep{Qwen3-VL}                      &  30B    &   47.5                       \\
        \midrule
        \rowcolor{gray!15}
        \multicolumn{3}{l}{\textit{GUI-specific Models}} \\
        WebVoyager~\citep{he2024webvoyager}               &   -    &  59.1 \\
        OpenAI-CUA~\citep{openai2024computeruse}                      &  -     &  \textbf{87.0}          \\
        UI-TARS-1.5~\citep{qin2025uitarspioneeringautomatedgui}               &   -    & 84.8 \\
        Holo2-4B~\citep{hai2025holo2modelfamily}                      &  4B     &      80.2                \\ 
        Holo2-8B~\citep{hai2025holo2modelfamily}                     &  8B     &      80.2                \\ 
        Holo2-30B-A3B~\citep{hai2025holo2modelfamily}                   &  30B     &    83.0                  \\ 
        \midrule
        \rowcolor{gray!15}
        \multicolumn{3}{l}{\textit{Ours}} \\
        
        \textbf{UI-Venus-1.5-2B}              &  2B &      56.4       \\ 
        \textbf{UI-Venus-1.5-8B}              &  8B &     70.8        \\ 
        \textbf{UI-Venus-1.5-30B-A3B}              &  30B     &   76.0             \\ 
        
        \bottomrule
        \end{tabular}
        \vspace{0.4em}
        \caption{Performance comparison on \textbf{Webvoyager} existing GUI Agents. Note that \textsuperscript{*} indicates that the results are evaluated by us.}
        \label{tab:webvoyager}
    \end{minipage}
\end{table}

\noindent \textbf{VenusBench-Mobile.}
VenusBench-Mobile is a challenging benchmark designed to evaluate the end-to-end performance of GUI agents in complex mobile environments. As illustrated in Table~\ref{tab:VenusBench-Mobile}, the UI-Venus-1.5 model family achieves state-of-the-art (SOTA) performance across all scales. Specifically, our 2B, 8B, and 30B-A3B variants reach success rates of 8.7\%, 16.1\%, and 21.5\%, respectively, consistently outperforming both general-purpose VLMs and specialized GUI models. Notably, our 8B model (16.1\%) already surpasses the much larger UI-Venus-1.0-72B (15.4\%), while our 30B-A3B variant sets a new record with a substantial 6.1\% absolute margin over the previous best-performing model.

\noindent \textbf{WebVoyager.}
WebVoyager is a comprehensive end-to-end benchmark for evaluating web agents on 15 real-world websites including e-commerce, travel, and social platforms. 
It employs an automatic evaluation protocol leveraging MLLM to assess task completion rates, measuring agents' abilities to autonomously navigate and interact with dynamic web environments through visual screenshots and textual elements. 
As shown in Table~\ref{tab:webvoyager}, the UI-Venus-1.5 model family achieves comparable performance among models of comparable scale. Specifically, our 2B/8B/30B-A3B variants reach success rates of 56.4\%/70.8\%/76.0\%, outperforming WebVoyager~\cite{he2024webvoyager} and general VLMs (\emph{e.g.}, GPT-4o, Qwen3-VL). 

In summary, our evaluation reveals several key insights:
\begin{itemize}
    \item \textbf{Superior End-to-End Performance:} UI-Venus-1.5-30B-A3B achieves state-of-the-art or comparable results across a diverse range of GUI agent benchmarks, including Android World (77.6\%), Android Lab (55.1\%/68.1\%$\dagger$), VenusBench-Mobile (21.5\%), and WebVoyager (76.0\%). It consistently outperforms both specialized GUI models (\emph{e.g.}, MAI-UI-32B) and leading general-purpose VLMs (\emph{e.g.}, GPT-4o, Qwen3-VL), establishing a new performance ceiling for autonomous agents.
    \item \textbf{Significant Architectural Efficiency and Scaling:} Increasing the model scale leads to consistent performance gains across all benchmarks. Notably, the UI-Venus-1.5 family exhibits remarkable efficiency; our 8B model already surpasses the previous generation's 72B variant on both Android Lab (up to 5.8\% improvement) and VenusBench-Mobile (16.1\% vs. 15.4\%), demonstrating the effectiveness of our updated training methodology.
    \item \textbf{Robust Cross-Platform Generalization:} UI-Venus demonstrates exceptional adaptability across different operating systems and input modalities. It excels not only in programmatic Android environments but also in dynamic web navigation (WebVoyager). Furthermore, the models show strong visual-only reasoning capabilities, outperforming XML-augmented baselines in Android Lab even when relying solely on raw screenshots.

\end{itemize}
\begin{figure}[htbp]
    \centering
    \includegraphics[width=0.7\textwidth]{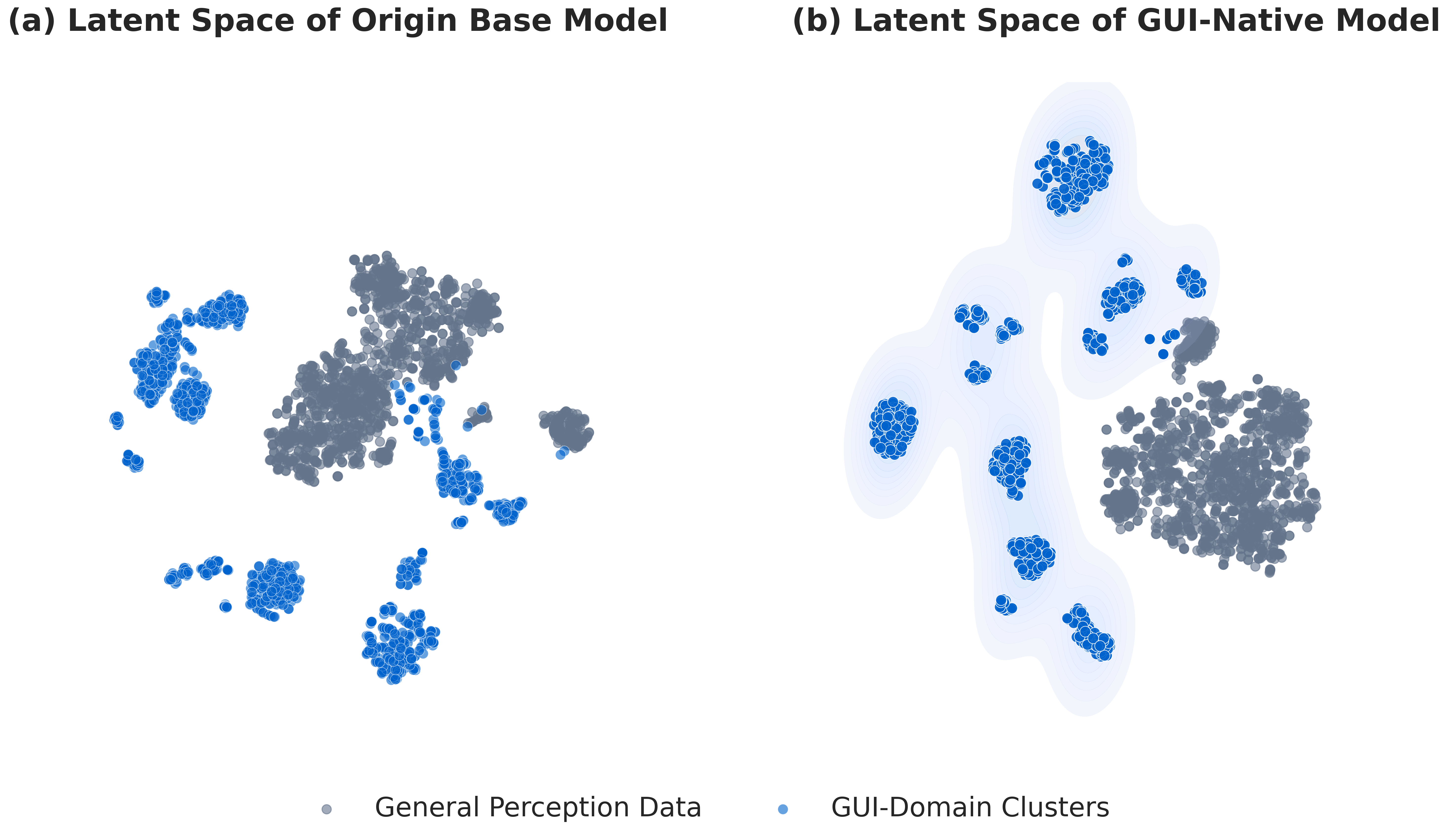} 
    \caption{\textbf{Latent space visualization} of (a) base model and (b) model with GUI knowledge. The emergence of distinct clusters indicates that our Mid-Training has successfully enriched the model with GUI domain knowledge. This increased discriminative power between GUI and general data provides a robust structural basis for the reinforcement learning stage.}
    \label{fig:tsne_visualization}
\end{figure}

\subsection{The Influence of Mid-Training}\label{sec:mid-train-exp}
To verify the effectiveness of our Mid-Training strategy, we conduct the qualitative latent space analysis as shown in Figure~\ref{fig:tsne_visualization}.
Specifically, we analyzed the latent representations using t-SNE visualization to quantify the impact of Mid-Training. By comparing the base model with our model after Mid-Training, we observe several key observations:

\begin{itemize}
    \item \textbf{Cluster Separability:} After Mid-Training, GUI-specific features exhibit a transition to high-density clusters. The Silhouette Score reached $0.315$, a relative increase of 34.0\% over the base model ($0.235 \rightarrow 0.315$).
    \item \textbf{Feature Sensitivity:} The 11.6\% decrease in Intra-class Consistency ($0.448 \rightarrow 0.396$) indicates enhanced discriminative power. It demonstrates that the model is now capable of capturing fine-grained functional and structural variances, allowing for a more granular characterization of GUI elements that were previously treated as uniform.
    \item \textbf{Global Space Stability:} The Inter-class Similarity remains stable (only 1.4\% increase), confirming that GUI-specific knowledge does not cause representation collapse.
\end{itemize}

\begin{table}[htbp]
\centering
\begin{tabular}{lccc}
\toprule
\textbf{Metric} & \textbf{Qwen3VL} & \textbf{Model After Mid-Training} & \textbf{Change} \\ \midrule
Silhouette Score & 0.235 & 0.315 & +34.0\% ($\uparrow$) \\
Intra-class Consistency & 0.448 & 0.396 & -11.6\% ($\downarrow$) \\
Inter-class Similarity & 0.220 & 0.223 & +1.4\% ($\approx$ stable) \\ \bottomrule
\end{tabular}
\caption{\textbf{Latent space metrics} of the model before and after Mid-Training.}
\label{tab:latent_metrics}
\end{table}

\subsection{Ablation Studies}\label{sec:ablation}

\begin{table}[ht]
    \centering
    \resizebox{1\linewidth}{!}{
    \begin{tabular}{l *{9}{c}}
    \toprule
    \multirow{2}{*}{\textbf{Models}} & \multicolumn{2}{c}{\textbf{Mid-Training}}  & \multicolumn{2}{c}{\textbf{Offline-RL}} & \multicolumn{2}{c}{\textbf{Online-RL}} & \multicolumn{2}{c}{\textbf{Model Merge}} \\ 
    \cmidrule(lr){2-3} \cmidrule(lr){4-5} \cmidrule(lr){6-7}\cmidrule(lr){8-9} 
    & \textbf{SS-Pro} & \textbf{AW} & \textbf{SS-Pro} & \textbf{AW} & \textbf{SS-Pro} & \textbf{AW}     & \textbf{SS-Pro} & \textbf{AW}     \\ 
    \midrule
    \rowcolor{gray!17}
    \multicolumn{9}{l}{\textit{UI-Venus-1.5}} \\
    \textbf{2B}    & 52.3 &39.0 & \textbf{59.0}(\textcolor{red}{+6.7$\uparrow$}) &45.3(\textcolor{red}{+6.3$\uparrow$})&- & \textbf{59.8}(\textcolor{red}{+14.5$\uparrow$}) & 57.7(\textcolor{green}{-1.3$\downarrow$})&    55.6 (\textcolor{green}{-4.2$\downarrow$})                       \\ 
    \textbf{8B}    & 63.1  &57.0 & \textbf{70.0}(\textcolor{red}{+6.9$\uparrow$}) & 63.5(\textcolor{red}{+6.5$\uparrow$})&- & 72.7(\textcolor{red}{+9.2$\uparrow$}) & 68.4(\textcolor{green}{-1.6$\downarrow$})&                           \textbf{73.7}(\textcolor{red}{+1.0$\uparrow$}) \\ 
    \textbf{30B-A3B}    & 65.2 &67.1 & \textbf{71.0}(\textcolor{red}{+5.8$\uparrow$})&68.0(\textcolor{red}{+0.9$\uparrow$}) &- & 75.5(\textcolor{red}{+7.5$\uparrow$})  &69.6(\textcolor{green}{-1.4$\downarrow$})&                           \textbf{77.6}(\textcolor{red}{+2.1$\uparrow$}) \\ 
    \bottomrule
    \end{tabular}
    }
    \caption{\textbf{Ablation studies of UI-Venus-1.5} on ScreenSpot-Pro (denote as ``SS-Pro'') and AndroidWorld (denote as ``AW'').}
    \label{tab:ablation}
\end{table}

In this section, we will show the performance gains of every step in the UI-Venus-1.5 pipeline, including Mid-Training, Offline-RL, Online-RL and Model Merge. As shown in Table~\ref{tab:ablation}, we can conclude following insights:

\begin{itemize}
\item \textbf{Offline-RL: Building Foundation for Grounding and Navigation.} The transition from Mid-Training to Offline-RL yields consistent improvements across all scales and tasks. Specifically, ScreenSpot-Pro scores increase by approximately 6--7\%, while AndroidWorld (AW) performance also sees a significant boost (up to +6.5\% for the 8B model). This confirms that GRPO on diverse, task-specific offline data effectively aligns the model's visual perception with GUI-specific action spaces.
\item \textbf{Online-RL: The Catalyst for Complex Navigation.} Online-RL serves as the most critical stage for enhancing autonomous navigation capabilities. We observe a substantial leap in AndroidWorld success rates, with the 2B model showing a remarkable +14.5\% absolute gain. By interacting with dynamic environments and learning from exploration, the models overcome the limitations of static datasets, significantly improving their ability to handle long-horizon tasks and error recovery in real-world scenarios.

\item \textbf{Model Merge: Balancing Specialization and Generalization.} The final model merge stage aims to unify specialized capabilities. While it leads to a minor, acceptable trade-off in fine-grained grounding (a drop of $\sim$1.4\% on ScreenSpot-Pro), it further stabilizes and enhances navigation performance for larger models. Notably, UI-Venus-1.5-30B-A3B gains an additional 2.1\% on AndroidWorld after the merge, suggesting that the unification process helps the model leverage cross-task knowledge to solve complex GUI sequences more effectively.

\end{itemize}



\section{Related Works}

GUI agents can automatically execute a series of operations on GUI screens based on given instructions. In early works, the system typically relied on predefined rules, which exhibited limited scalability. With the emergence and advancement of LLMs, it is possible to adopt a single model to handle diverse tasks as an intelligent GUI agent.

\noindent \textbf{GUI Grounding.} 
Some researches focus on GUI grounding that aims to map natural language descriptions or instructions to precise positions of GUI elements on screens, enabling autonomous agents to identify widgets with diverse functionalities and select the appropriate one to interact by an equipped planner. 
Early methods~\cite{cheng2024seeclickharnessingguigrounding,lin2024showuivisionlanguageactionmodelgui,xu2024aguvis,wu2024osatlasfoundationactionmodel,gu2023mobile,gou2024navigatingdigitalworldhumans,wang2024eantlargescaledatasetefficient,guiactor,xie2025scalingcomputerusegroundinguser,tang2025thinktwiceclickonce} usually adopt supervised fine-tuning (SFT) to train grounding models, which leverages labeled data rapidly and produces various grounding models that are able to recognize and locate diverse elements on common GUI scenarios.
As models evolve rapidly, the accuracies on some grounding benchmarks~\cite{cheng2024seeclickharnessingguigrounding,wu2024osatlasfoundationactionmodel} has hit a ceiling. To further evaluate the grounding capabilities of models in complex scenarios, more benchmarks like Screenspot-Pro~\cite{li2024screenspot-pro} and VenusBench-GD~\cite{zhou2025venusbench} have been introduced to explore the limit of performance, which additionally requires the understanding of professional software and comprehensive visual reasoning. In parallel, the training paradigms are shifting. Inspired by DeepSeek-R1~\cite{deepseekai2025deepseekr1incentivizingreasoningcapability}, recent works have incorporated reinforcement learning (RL) into training process, aiming to enhance model generalization across unseen scenarios with limited labeled data~\cite{lu2025uir1enhancingefficientaction,luo2025guir1generalistr1style,liu2025infiguir1advancingmultimodalgui,zhou2025guig1understandingr1zeroliketraining,yuan2025enhancingvisualgroundinggui,tang2025lpoaccurateguiagent,guig2,zhang2026omegausebuildinggeneralpurposegui,zhou2025mai,qin2025uitarspioneeringautomatedgui,ui-tars-15-seed,yan2025step}. In addition, some works~\cite{zhang2025mvp,jiang2025zoom} focus on post-processing to improve the test-time performance of the model. The mutual evolution between benchmarks and models drives the continuous advancement of the grounding research.

\noindent \textbf{End-to-End GUI Agent.}  
Also, some researchers attempt to tackle navigation tasks end-to-end with a unified model.
At early stage, the agents were trained on foundation models directly, which served as preliminary attempts at autonomous GUI agents and produced promising results~\cite{hong2024cogagentvisuallanguagemodel,lin2024showuivisionlanguageactionmodelgui,cheng2024seeclickharnessingguigrounding,deng2023mind2web,wu2024osatlasfoundationactionmodel}, but there remained a gap between the actual performance and the requirement of practical deployment. Driven by further practicality, with the incorporation of RL techniques like Direct Preference Optimization (DPO)~\cite{rafailov2023direct} and Group Relative Policy Optimization (GRPO)~\cite{shao2024deepseekmathpushinglimitsmathematical}, improved pipeline of data generation and increased computational resources, many research groups have developed more powerful agents~\cite{qin2025uitarspioneeringautomatedgui,zeng2025uitron,sun2025guixploreempoweringgeneralizablegui,yang2024ariauivisualgroundinggui,sun2025osgenesisautomatingguiagent,qiu2026unified}. More recently, as the alignment between real-world and training environments receives widespread focus, more end-to-end GUI agents exhibit robust practical deployment capabilities~\cite{wang2025ui,liu2024autoglm,yan2025step}, which advances the feasibility of GUI agents in real-life scenarios.




\noindent \textbf{GUI Agent Framework.} 
As a collaboration paradigm that distributes the extensive context across sub-agents with various functionalities, the GUI agent framework fully leverages the base model's understanding and reasoning capabilities to analyze the task progress and GUI screens, and subsequently takes a correct action. 
Agent S~\cite{agashe2024agent} introduces experience-augmented hierarchical planning strategy to improve task execution with several role-specific agents, and Agent S2~\cite{agashe2025agent} upgrades the strategy as proactive hierarchical planning to dynamically refine actions based on real-time observations. 
Besides, began with Mobile-Agent~\cite{wang2024mobileagentautonomousmultimodalmobile}, Mobile-Agent-v2~\cite{wang2024mobileagentv2mobiledeviceoperation} incorporates a multi-agent collaboration architecture for long-step navigation, which includes planning, decision, reflection agents and a memory unit to retain focus content. Moreover, Mobile-Agent-E~\cite{wang2025mobileagenteselfevolvingmobileassistant} implements a self-evolving hierarchical framework to store long-term memory and learn from the past, and Mobile-Agent-v3~\cite{ye2025mobile} further advances the framework by improved base model and training strategies based on its predecessors. More studies have also explored GUI agent frameworks~\cite{droidrun,xie2025gui}.
These agent frameworks probably possess a higher capacity ceiling, enabling them to perform navigation tasks that require intricate reasoning and analysis. Nevertheless, the data flow in the framework usually necessitates multiple rounds of LLM input and output, which leads to substantial computational costs and obvious operation latency.




\section{Conclusion}

In this work, we presented UI-Venus-1.5, a comprehensive advancement in the development of practical and reliable GUI agents. To address the limitations of previous iterations and current baselines, we introduced a three-tiered training paradigm: a large-scale Mid-Training stage for robust GUI knowledge injection, a task-specific Offline-RL phase unified by an efficient model merge strategy, and a scaled Online-RL framework to master complex navigation. Furthermore, the unified capability of UI-Venus-1.5 is achieved through strategic model merging, which effectively consolidates specialized domain expertise—including grounding, web, and mobile navigation—into a single end-to-end agent while preserving robust performance across tasks.

Experimental results demonstrate that UI-Venus-1.5 establishes a new state of the art across a wide spectrum of benchmarks, including GUI grounding and navigation. 
Beyond academic metrics, we have optimized the model for real-world utility within the 40+ Chinese third-party app ecosystem, enabling seamless automation for tasks such as ticket booking, and shopping. 
Collectively, these contributions on GUI Agents mark a significant step towards a truly autonomous and user-centric digital assistant.


\section{Contributions}
All contributors of UI-Venus-1.5 are listed in alphabetical order by their last names. 

\subsection{Core Contributors}
Changlong~Gao, Zhangxuan~Gu, Yulin~Liu, Xinyu~Qiu, Shuheng~Shen$^{\dag}$, Yue~Wen, Tianyu~Xia, Zhenyu~Xu, Zhengwen~Zeng, Beitong~Zhou, Xingran~Zhou

\subsection{Contributors}
Weizhi~Chen, Sunhao~Dai, Jingya~Dou, Yichen~Gong, Yuan~Guo, Zhenlin~Guo, Feng~Li, Qian~Li, Jinzhen~Lin, Yuqi~Zhou, Linchao~Zhu

\subsection{Supervisors}
Liang~Chen, Zhenyu~Guo, Changhua~Meng$^{\dag}$, Weiqiang~Wang

\footnote{$^{\dag}$Corresponding Authors: Shuheng~Shen(shuheng.ssh@antgroup.com), Changhua~Meng(changhua.mch@antgroup.com).}

\newpage
\bibliographystyle{assets/plainnat}
\bibliography{main}

@misc{androidworld,
  title         = {AndroidWorld: A Dynamic Benchmarking Environment for Autonomous Agents},
  author        = {Christopher Rawles and Sarah Clinckemaillie and Yifan Chang and Jonathan Waltz and Gabrielle Lau and Marybeth Fair and Alice Li and William Bishop and Wei Li and Folawiyo Campbell-Ajala and Daniel Toyama and Robert Berry and Divya Tyamagundlu and Timothy Lillicrap and Oriana Riva},
  year          = {2025},
  eprint        = {2405.14573},
  archiveprefix = {arXiv},
  primaryclass  = {cs.AI},
  url           = {https://arxiv.org/abs/2405.14573}
}

@misc{guiodyssey,
  title         = {GUIOdyssey: A Comprehensive Dataset for Cross-App GUI Navigation on Mobile Devices},
  author        = {Quanfeng Lu and Wenqi Shao and Zitao Liu and Lingxiao Du and Fanqing Meng and Boxuan Li and Botong Chen and Siyuan Huang and Kaipeng Zhang and Ping Luo},
  year          = {2025},
  eprint        = {2406.08451},
  archiveprefix = {arXiv},
  primaryclass  = {cs.CV},
  url           = {https://arxiv.org/abs/2406.08451}
}

@misc{androidcontrol,
  title         = {On the Effects of Data Scale on UI Control Agents},
  author        = {Wei Li and William Bishop and Alice Li and Chris Rawles and Folawiyo Campbell-Ajala and Divya Tyamagundlu and Oriana Riva},
  year          = {2024},
  eprint        = {2406.03679},
  archiveprefix = {arXiv},
  primaryclass  = {cs.AI},
  url           = {https://arxiv.org/abs/2406.03679}
}

@misc{tang2025lpoaccurateguiagent,
  title         = {LPO: Towards Accurate GUI Agent Interaction via Location Preference Optimization},
  author        = {Jiaqi Tang and Yu Xia and Yi-Feng Wu and Yuwei Hu and Yuhui Chen and Qing-Guo Chen and Xiaogang Xu and Xiangyu Wu and Hao Lu and Yanqing Ma and Shiyin Lu and Qifeng Chen},
  year          = {2025},
  eprint        = {2506.09373},
  archiveprefix = {arXiv},
  primaryclass  = {cs.LG},
  url           = {https://arxiv.org/abs/2506.09373}
}

@misc{bai2025qwen25vltechnicalreport,
  title         = {Qwen2.5-VL Technical Report},
  author        = {Shuai Bai and Keqin Chen and Xuejing Liu and Jialin Wang and Wenbin Ge and Sibo Song and Kai Dang and Peng Wang and Shijie Wang and Jun Tang and Humen Zhong and Yuanzhi Zhu and Mingkun Yang and Zhaohai Li and Jianqiang Wan and Pengfei Wang and Wei Ding and Zheren Fu and Yiheng Xu and Jiabo Ye and Xi Zhang and Tianbao Xie and Zesen Cheng and Hang Zhang and Zhibo Yang and Haiyang Xu and Junyang Lin},
  year          = {2025},
  eprint        = {2502.13923},
  archiveprefix = {arXiv},
  primaryclass  = {cs.CV},
  url           = {https://arxiv.org/abs/2502.13923}
}

@misc{zhu2025internvl3exploringadvancedtraining,
  title         = {InternVL3: Exploring Advanced Training and Test-Time Recipes for Open-Source Multimodal Models},
  author        = {Jinguo Zhu and Weiyun Wang and Zhe Chen and Zhaoyang Liu and Shenglong Ye and Lixin Gu and Hao Tian and Yuchen Duan and Weijie Su and Jie Shao and Zhangwei Gao and Erfei Cui and Xuehui Wang and Yue Cao and Yangzhou Liu and Xingguang Wei and Hongjie Zhang and Haomin Wang and Weiye Xu and Hao Li and Jiahao Wang and Nianchen Deng and Songze Li and Yinan He and Tan Jiang and Jiapeng Luo and Yi Wang and Conghui He and Botian Shi and Xingcheng Zhang and Wenqi Shao and Junjun He and Yingtong Xiong and Wenwen Qu and Peng Sun and Penglong Jiao and Han Lv and Lijun Wu and Kaipeng Zhang and Huipeng Deng and Jiaye Ge and Kai Chen and Limin Wang and Min Dou and Lewei Lu and Xizhou Zhu and Tong Lu and Dahua Lin and Yu Qiao and Jifeng Dai and Wenhai Wang},
  year          = {2025},
  eprint        = {2504.10479},
  archiveprefix = {arXiv},
  primaryclass  = {cs.CV},
  url           = {https://arxiv.org/abs/2504.10479}
}

@misc{shao2024deepseekmathpushinglimitsmathematical,
  title         = {DeepSeekMath: Pushing the Limits of Mathematical Reasoning in Open Language Models},
  author        = {Zhihong Shao and Peiyi Wang and Qihao Zhu and Runxin Xu and Junxiao Song and Xiao Bi and Haowei Zhang and Mingchuan Zhang and Y. K. Li and Y. Wu and Daya Guo},
  year          = {2024},
  eprint        = {2402.03300},
  archiveprefix = {arXiv},
  primaryclass  = {cs.CL},
  url           = {https://arxiv.org/abs/2402.03300}
}

@misc{guig2,
  title         = {GUI-G$^2$: Gaussian Reward Modeling for GUI Grounding},
  author        = {Fei Tang and Zhangxuan Gu and Zhengxi Lu and Xuyang Liu and Shuheng Shen and Changhua Meng and Wen Wang and Wenqi Zhang and Yongliang Shen and Weiming Lu and Jun Xiao and Yueting Zhuang},
  year          = {2025},
  eprint        = {2507.15846},
  archiveprefix = {arXiv},
  primaryclass  = {cs.LG},
  url           = {https://arxiv.org/abs/2507.15846}
}

@misc{anthropic2024cuda,
  title        = {Claude Computer Use},
  author       = {Anthropic},
  howpublished = {Available at: https://www.anthropic.com/news/developing-computer-use},
  year         = {2024}
}

@misc{yang2025gta1guitesttimescaling,
  title         = {GTA1: GUI Test-time Scaling Agent},
  author        = {Yan Yang and Dongxu Li and Yutong Dai and Yuhao Yang and Ziyang Luo and Zirui Zhao and Zhiyuan Hu and Junzhe Huang and Amrita Saha and Zeyuan Chen and Ran Xu and Liyuan Pan and Caiming Xiong and Junnan Li},
  year          = {2025},
  eprint        = {2507.05791},
  archiveprefix = {arXiv},
  primaryclass  = {cs.AI},
  url           = {https://arxiv.org/abs/2507.05791}
}

@article{zhou2025guig1understandingr1zeroliketraining,
  title         = {GUI-G1: Understanding R1-Zero-Like Training for Visual Grounding in GUI Agents},
  author        = {Yuqi Zhou and Sunhao Dai and Shuai Wang and Kaiwen Zhou and Qinglin Jia and Jun Xu},
  year          = {2025},
  eprint        = {2505.15810},
  archiveprefix = {arXiv},
  primaryclass  = {cs.CL},
  url           = {https://arxiv.org/abs/2505.15810}
}

@article{yuan2025enhancingvisualgroundinggui,
  title         = {Enhancing Visual Grounding for GUI Agents via Self-Evolutionary Reinforcement Learning},
  author        = {Xinbin Yuan and Jian Zhang and Kaixin Li and Zhuoxuan Cai and Lujian Yao and Jie Chen and Enguang Wang and Qibin Hou and Jinwei Chen and Peng-Tao Jiang and Bo Li},
  year          = {2025},
  eprint        = {2505.12370},
  archiveprefix = {arXiv},
  primaryclass  = {cs.AI},
  url           = {https://arxiv.org/abs/2505.12370}
}

@article{liu2025infiguir1advancingmultimodalgui,
  title         = {InfiGUI-R1: Advancing Multimodal GUI Agents from Reactive Actors to Deliberative Reasoners},
  author        = {Yuhang Liu and Pengxiang Li and Congkai Xie and Xavier Hu and Xiaotian Han and Shengyu Zhang and Hongxia Yang and Fei Wu},
  year          = {2025},
  eprint        = {2504.14239},
  archiveprefix = {arXiv},
  primaryclass  = {cs.AI},
  url           = {https://arxiv.org/abs/2504.14239}
}

@article{luo2025guir1generalistr1style,
  title         = {GUI-R1 : A Generalist R1-Style Vision-Language Action Model For GUI Agents},
  author        = {Run Luo and Lu Wang and Wanwei He and Xiaobo Xia},
  year          = {2025},
  eprint        = {2504.10458},
  archiveprefix = {arXiv},
  primaryclass  = {cs.CV},
  url           = {https://arxiv.org/abs/2504.10458}
}

@article{lu2025uir1enhancingefficientaction,
  title         = {UI-R1: Enhancing Efficient Action Prediction of GUI Agents by Reinforcement Learning},
  author        = {Zhengxi Lu and Yuxiang Chai and Yaxuan Guo and Xi Yin and Liang Liu and Hao Wang and Han Xiao and Shuai Ren and Guanjing Xiong and Hongsheng Li},
  year          = {2025},
  eprint        = {2503.21620},
  archiveprefix = {arXiv},
  primaryclass  = {cs.AI},
  url           = {https://arxiv.org/abs/2503.21620}
}

@article{tang2025thinktwiceclickonce,
  title         = {Think Twice, Click Once: Enhancing GUI Grounding via Fast and Slow Systems},
  author        = {Fei Tang and Yongliang Shen and Hang Zhang and Siqi Chen and Guiyang Hou and Wenqi Zhang and Wenqiao Zhang and Kaitao Song and Weiming Lu and Yueting Zhuang},
  year          = {2025},
  eprint        = {2503.06470},
  archiveprefix = {arXiv},
  primaryclass  = {cs.AI},
  url           = {https://arxiv.org/abs/2503.06470}
}

@article{guiactor,
  title   = {GUI-Actor: Coordinate-Free Visual Grounding for GUI Agents},
  author  = {Wu, Qianhui and Cheng, Kanzhi and Yang, Rui and Zhang, Chaoyun and Yang, Jianwei and Jiang, Huiqiang and Mu, Jian and Peng, Baolin and Qiao, Bo and Tan, Reuben and others},
  journal = {arXiv preprint arXiv:2506.03143},
  year    = {2025}
}

@artice{deepseekai2025deepseekr1incentivizingreasoningcapability,
  title         = {DeepSeek-R1: Incentivizing Reasoning Capability in LLMs via Reinforcement Learning},
  author        = {DeepSeek-AI},
  year          = {2025},
  eprint        = {2501.12948},
  archiveprefix = {arXiv},
  primaryclass  = {cs.CL},
  url           = {https://arxiv.org/abs/2501.12948}
}

@artice{qin2025uitarspioneeringautomatedgui,
  title         = {UI-TARS: Pioneering Automated GUI Interaction with Native Agents},
  author        = {Yujia Qin and Yining Ye and Junjie Fang and Haoming Wang and Shihao Liang and Shizuo Tian and Junda Zhang and Jiahao Li and Yunxin Li and Shijue Huang and Wanjun Zhong and Kuanye Li and Jiale Yang and Yu Miao and Woyu Lin and Longxiang Liu and Xu Jiang and Qianli Ma and Jingyu Li and Xiaojun Xiao and Kai Cai and Chuang Li and Yaowei Zheng and Chaolin Jin and Chen Li and Xiao Zhou and Minchao Wang and Haoli Chen and Zhaojian Li and Haihua Yang and Haifeng Liu and Feng Lin and Tao Peng and Xin Liu and Guang Shi},
  year          = {2025},
  eprint        = {2501.12326},
  archiveprefix = {arXiv},
  primaryclass  = {cs.AI},
  url           = {https://arxiv.org/abs/2501.12326}
}

@artice{hong2024cogagentvisuallanguagemodel,
  title         = {CogAgent: A Visual Language Model for GUI Agents},
  author        = {Wenyi Hong and Weihan Wang and Qingsong Lv and Jiazheng Xu and Wenmeng Yu and Junhui Ji and Yan Wang and Zihan Wang and Yuxuan Zhang and Juanzi Li and Bin Xu and Yuxiao Dong and Ming Ding and Jie Tang},
  year          = {2024},
  eprint        = {2312.08914},
  archiveprefix = {arXiv},
  primaryclass  = {cs.CV},
  url           = {https://arxiv.org/abs/2312.08914}
}

@artice{wang2024mobileagentautonomousmultimodalmobile,
  title         = {Mobile-Agent: Autonomous Multi-Modal Mobile Device Agent with Visual Perception},
  author        = {Junyang Wang and Haiyang Xu and Jiabo Ye and Ming Yan and Weizhou Shen and Ji Zhang and Fei Huang and Jitao Sang},
  year          = {2024},
  eprint        = {2401.16158},
  archiveprefix = {arXiv},
  primaryclass  = {cs.CL},
  url           = {https://arxiv.org/abs/2401.16158}
}

@artice{wang2024mobileagentv2mobiledeviceoperation,
  title         = {Mobile-Agent-v2: Mobile Device Operation Assistant with Effective Navigation via Multi-Agent Collaboration},
  author        = {Junyang Wang and Haiyang Xu and Haitao Jia and Xi Zhang and Ming Yan and Weizhou Shen and Ji Zhang and Fei Huang and Jitao Sang},
  year          = {2024},
  eprint        = {2406.01014},
  archiveprefix = {arXiv},
  primaryclass  = {cs.CL},
  url           = {https://arxiv.org/abs/2406.01014}
}

@artice{wang2025mobileagenteselfevolvingmobileassistant,
  title         = {Mobile-Agent-E: Self-Evolving Mobile Assistant for Complex Tasks},
  author        = {Zhenhailong Wang and Haiyang Xu and Junyang Wang and Xi Zhang and Ming Yan and Ji Zhang and Fei Huang and Heng Ji},
  year          = {2025},
  eprint        = {2501.11733},
  archiveprefix = {arXiv},
  primaryclass  = {cs.CL},
  url           = {https://arxiv.org/abs/2501.11733}
}

@artice{cheng2024seeclickharnessingguigrounding,
  title         = {SeeClick: Harnessing GUI Grounding for Advanced Visual GUI Agents},
  author        = {Kanzhi Cheng and Qiushi Sun and Yougang Chu and Fangzhi Xu and Yantao Li and Jianbing Zhang and Zhiyong Wu},
  year          = {2024},
  eprint        = {2401.10935},
  archiveprefix = {arXiv},
  primaryclass  = {cs.HC},
  url           = {https://arxiv.org/abs/2401.10935}
}

@artice{lin2024showuivisionlanguageactionmodelgui,
  title         = {ShowUI: One Vision-Language-Action Model for GUI Visual Agent},
  author        = {Kevin Qinghong Lin and Linjie Li and Difei Gao and Zhengyuan Yang and Shiwei Wu and Zechen Bai and Weixian Lei and Lijuan Wang and Mike Zheng Shou},
  year          = {2024},
  eprint        = {2411.17465},
  archiveprefix = {arXiv},
  primaryclass  = {cs.CV},
  url           = {https://arxiv.org/abs/2411.17465}
}

@artice{gou2024navigatingdigitalworldhumans,
  title         = {Navigating the Digital World as Humans Do: Universal Visual Grounding for GUI Agents},
  author        = {Boyu Gou and Ruohan Wang and Boyuan Zheng and Yanan Xie and Cheng Chang and Yiheng Shu and Huan Sun and Yu Su},
  year          = {2024},
  eprint        = {2410.05243},
  archiveprefix = {arXiv},
  primaryclass  = {cs.AI},
  url           = {https://arxiv.org/abs/2410.05243}
}

@artice{li2024screenspot-pro,
  title  = {ScreenSpot-Pro: GUI Grounding for Professional High-Resolution Computer Use},
  author = {Kaixin Li and Ziyang Meng and Hongzhan Lin and Ziyang Luo and Yuchen Tian and Jing Ma and Zhiyong Huang and Tat-Seng Chua},
  year   = {2025}
}

@article{xu2024aguvis,
  title  = {Aguvis: Unified Pure Vision Agents for Autonomous GUI Interaction},
  author = {Yiheng Xu and Zekun Wang and Junli Wang and Dunjie Lu and Tianbao Xie and Amrita Saha and Doyen Sahoo and Tao Yu and Caiming Xiong},
  year   = {2024},
  url    = {https://arxiv.org/abs/2412.04454}
}

@artice{yang2024ariauivisualgroundinggui,
  title         = {Aria-UI: Visual Grounding for GUI Instructions},
  author        = {Yuhao Yang and Yue Wang and Dongxu Li and Ziyang Luo and Bei Chen and Chao Huang and Junnan Li},
  year          = {2024},
  eprint        = {2412.16256},
  archiveprefix = {arXiv},
  primaryclass  = {cs.HC},
  url           = {https://arxiv.org/abs/2412.16256}
}

@article{rawles2023androidinthewild,
  title   = {Androidinthewild: A large-scale dataset for android device control},
  author  = {Rawles, Christopher and Li, Alice and Rodriguez, Daniel and Riva, Oriana and Lillicrap, Timothy},
  journal = {Advances in Neural Information Processing Systems},
  volume  = {36},
  pages   = {59708--59728},
  year    = {2023}
}

@article{nayak2025ui,
  title   = {Ui-vision: A desktop-centric gui benchmark for visual perception and interaction},
  author  = {Nayak, Shravan and Jian, Xiangru and Lin, Kevin Qinghong and Rodriguez, Juan A and Kalsi, Montek and Awal, Rabiul and Chapados, Nicolas and {\"O}zsu, M Tamer and Agrawal, Aishwarya and Vazquez, David and others},
  journal = {arXiv preprint arXiv:2503.15661},
  year    = {2025}
}

@misc{ui-tars-15-seed,
  title        = {UI-TARS-1.5},
  author       = {ByteDance Seed},
  howpublished = {\url{https://seed-tars.com/1.5}},
  year         = {2025}
}

@artice{wu2024osatlasfoundationactionmodel,
  title         = {OS-ATLAS: A Foundation Action Model for Generalist GUI Agents},
  author        = {Zhiyong Wu and Zhenyu Wu and Fangzhi Xu and Yian Wang and Qiushi Sun and Chengyou Jia and Kanzhi Cheng and Zichen Ding and Liheng Chen and Paul Pu Liang and Yu Qiao},
  year          = {2024},
  eprint        = {2410.23218},
  archiveprefix = {arXiv},
  primaryclass  = {cs.CL},
  url           = {https://arxiv.org/abs/2410.23218}
}

@article{agashe2024agent,
  title   = {Agent s: An open agentic framework that uses computers like a human},
  author  = {Agashe, Saaket and Han, Jiuzhou and Gan, Shuyu and Yang, Jiachen and Li, Ang and Wang, Xin Eric},
  journal = {arXiv preprint arXiv:2410.08164},
  year    = {2024}
}

@article{agashe2025agent,
  title   = {Agent s2: A compositional generalist-specialist framework for computer use agents},
  author  = {Agashe, Saaket and Wong, Kyle and Tu, Vincent and Yang, Jiachen and Li, Ang and Wang, Xin Eric},
  journal = {arXiv preprint arXiv:2504.00906},
  year    = {2025}
}

@misc{droidrun,
  title        = {Droidrun},
  howpublished = {\url{https://github.com/droidrun/droidrun}},
  year         = {2025}
}

@misc{xie2025scalingcomputerusegroundinguser,
  title         = {Scaling Computer-Use Grounding via User Interface Decomposition and Synthesis},
  author        = {Tianbao Xie and Jiaqi Deng and Xiaochuan Li and Junlin Yang and Haoyuan Wu and Jixuan Chen and Wenjing Hu and Xinyuan Wang and Yuhui Xu and Zekun Wang and Yiheng Xu and Junli Wang and Doyen Sahoo and Tao Yu and Caiming Xiong},
  year          = {2025},
  eprint        = {2505.13227},
  archiveprefix = {arXiv},
  primaryclass  = {cs.AI},
  url           = {https://arxiv.org/abs/2505.13227}
}

@inproceedings{gu2023mobile,
  title     = {Mobile user interface element detection via adaptively prompt tuning},
  author    = {Gu, Zhangxuan and Xu, Zhuoer and Chen, Haoxing and Lan, Jun and Meng, Changhua and Wang, Weiqiang},
  booktitle = {Proceedings of the IEEE/CVF Conference on Computer Vision and Pattern Recognition},
  pages     = {11155--11164},
  year      = {2023}
}

@misc{glm-4.5v,
  title        = {GLM-4.5V},
  author       = {Zhipu-AI},
  howpublished = {Available at: https://docs.z.ai/guides/vlm/glm-4.5v},
  year         = {2025}
}

@misc{wang2025opencuaopenfoundationscomputeruse,
  title         = {OpenCUA: Open Foundations for Computer-Use Agents},
  author        = {Xinyuan Wang and Bowen Wang and Dunjie Lu and Junlin Yang and Tianbao Xie and Junli Wang and Jiaqi Deng and Xiaole Guo and Yiheng Xu and Chen Henry Wu and Zhennan Shen and Zhuokai Li and Ryan Li and Xiaochuan Li and Junda Chen and Boyuan Zheng and Peihang Li and Fangyu Lei and Ruisheng Cao and Yeqiao Fu and Dongchan Shin and Martin Shin and Jiarui Hu and Yuyan Wang and Jixuan Chen and Yuxiao Ye and Danyang Zhang and Dikang Du and Hao Hu and Huarong Chen and Zaida Zhou and Yipu Wang and Heng Wang and Diyi Yang and Victor Zhong and Flood Sung and Y. Charles and Zhilin Yang and Tao Yu},
  year          = {2025},
  eprint        = {2508.09123},
  archiveprefix = {arXiv},
  primaryclass  = {cs.AI},
  url           = {https://arxiv.org/abs/2508.09123}
}

@misc{wang2024eantlargescaledatasetefficient,
  title         = {E-ANT: A Large-Scale Dataset for Efficient Automatic GUI NavigaTion},
  author        = {Ke Wang and Tianyu Xia and Zhangxuan Gu and Yi Zhao and Shuheng Shen and Changhua Meng and Weiqiang Wang and Ke Xu},
  year          = {2024},
  eprint        = {2406.14250},
  archiveprefix = {arXiv},
  primaryclass  = {cs.CV},
  url           = {https://arxiv.org/abs/2406.14250}
}

@misc{sun2025guixploreempoweringgeneralizablegui,
  title         = {GUI-Xplore: Empowering Generalizable GUI Agents with One Exploration},
  author        = {Yuchen Sun and Shanhui Zhao and Tao Yu and Hao Wen and Samith Va and Mengwei Xu and Yuanchun Li and Chongyang Zhang},
  year          = {2025},
  eprint        = {2503.17709},
  archiveprefix = {arXiv},
  primaryclass  = {cs.CV},
  url           = {https://arxiv.org/abs/2503.17709}
}

@misc{sun2025osgenesisautomatingguiagent,
  title         = {OS-Genesis: Automating GUI Agent Trajectory Construction via Reverse Task Synthesis},
  author        = {Qiushi Sun and Kanzhi Cheng and Zichen Ding and Chuanyang Jin and Yian Wang and Fangzhi Xu and Zhenyu Wu and Chengyou Jia and Liheng Chen and Zhoumianze Liu and Ben Kao and Guohao Li and Junxian He and Yu Qiao and Zhiyong Wu},
  year          = {2025},
  eprint        = {2412.19723},
  archiveprefix = {arXiv},
  primaryclass  = {cs.AI},
  url           = {https://arxiv.org/abs/2412.19723}
}

@article{Qwen3-VL,
  title   = {Qwen3-VL Technical Report},
  author  = {Shuai Bai and Yuxuan Cai and Ruizhe Chen and Keqin Chen and Xionghui Chen and Zesen Cheng and Lianghao Deng and Wei Ding and Chang Gao and Chunjiang Ge and Wenbin Ge and Zhifang Guo and Qidong Huang and Jie Huang and Fei Huang and Binyuan Hui and Shutong Jiang and Zhaohai Li and Mingsheng Li and Mei Li and Kaixin Li and Zicheng Lin and Junyang Lin and Xuejing Liu and Jiawei Liu and Chenglong Liu and Yang Liu and Dayiheng Liu and Shixuan Liu and Dunjie Lu and Ruilin Luo and Chenxu Lv and Rui Men and Lingchen Meng and Xuancheng Ren and Xingzhang Ren and Sibo Song and Yuchong Sun and Jun Tang and Jianhong Tu and Jianqiang Wan and Peng Wang and Pengfei Wang and Qiuyue Wang and Yuxuan Wang and Tianbao Xie and Yiheng Xu and Haiyang Xu and Jin Xu and Zhibo Yang and Mingkun Yang and Jianxin Yang and An Yang and Bowen Yu and Fei Zhang and Hang Zhang and Xi Zhang and Bo Zheng and Humen Zhong and Jingren Zhou and Fan Zhou and Jing Zhou and Yuanzhi Zhu and Ke Zhu},
  journal = {arXiv preprint arXiv:2511.21631},
  year    = {2025}
}

@article{zhou2025mai,
  title   = {MAI-UI Technical Report: Real-World Centric Foundation GUI Agents},
  author  = {Zhou, Hanzhang and Zhang, Xu and Tong, Panrong and Zhang, Jianan and Chen, Liangyu and Kong, Quyu and Cai, Chenglin and Liu, Chen and Wang, Yue and Zhou, Jingren and others},
  journal = {arXiv preprint arXiv:2512.22047},
  year    = {2025}
}

@misc{seed1.8,
  title  = {Seed1. 8 Model Card: Towards Generalized Real-World Agency},
  author = {Seed, Bytedance},
  url    = {https://lf3-static.bytednsdoc.com/obj/eden-cn/lapzild-tss/ljhwZthlaukjlkulzlp/research/Seed-1.8-Modelcard.pdf},
  year   = {2025}
}

@article{zhou2025venusbench,
  title   = {VenusBench-GD: A Comprehensive Multi-Platform GUI Benchmark for Diverse Grounding Tasks},
  author  = {Zhou, Beitong and Huang, Zhexiao and Guo, Yuan and Gu, Zhangxuan and Xia, Tianyu and Luo, Zichen and Tang, Fei and Kong, Dehan and Shang, Yanyi and Ou, Suling and others},
  journal = {arXiv preprint arXiv:2512.16501},
  year    = {2025}
}

@article{wang2025mmbench,
  title   = {Mmbench-gui: Hierarchical multi-platform evaluation framework for gui agents},
  author  = {Wang, Xuehui and Wu, Zhenyu and Xie, JingJing and Ding, Zichen and Yang, Bowen and Li, Zehao and Liu, Zhaoyang and Li, Qingyun and Dong, Xuan and Chen, Zhe and others},
  journal = {arXiv preprint arXiv:2507.19478},
  year    = {2025}
}

@article{gu2025ui,
  title   = {Ui-venus technical report: Building high-performance ui agents with rft},
  author  = {Gu, Zhangxuan and Zeng, Zhengwen and Xu, Zhenyu and Zhou, Xingran and Shen, Shuheng and Liu, Yunfei and Zhou, Beitong and Meng, Changhua and Xia, Tianyu and Chen, Weizhi and others},
  journal = {arXiv preprint arXiv:2508.10833},
  year    = {2025}
}

@article{liu2024autoglm,
  title   = {Autoglm: Autonomous foundation agents for guis},
  author  = {Liu, Xiao and Qin, Bo and Liang, Dongzhu and Dong, Guang and Lai, Hanyu and Zhang, Hanchen and Zhao, Hanlin and Iong, Iat Long and Sun, Jiadai and Wang, Jiaqi and others},
  journal = {arXiv preprint arXiv:2411.00820},
  year    = {2024}
}

@article{he2024webvoyager,
  title   = {Webvoyager: Building an end-to-end web agent with large multimodal models},
  author  = {He, Hongliang and Yao, Wenlin and Ma, Kaixin and Yu, Wenhao and Dai, Yong and Zhang, Hongming and Lan, Zhenzhong and Yu, Dong},
  journal = {arXiv preprint arXiv:2401.13919},
  year    = {2024}
}

@article{chen2025tgrpo,
  title   = {Tgrpo: Fine-tuning vision-language-action model via trajectory-wise group relative policy optimization},
  author  = {Chen, Zengjue and Niu, Runliang and Kong, He and Wang, Qi and Xing, Qianli and Fan, Zipei},
  journal = {arXiv preprint arXiv:2506.08440},
  year    = {2025}
}

@article{team2025tongyi,
  title   = {Tongyi deepresearch technical report},
  author  = {Tongyi-Team DeepResearch and Li, Baixuan and Zhang, Bo and Zhang, Dingchu and Huang, Fei and Li, Guangyu and Chen, Guoxin and Yin, Huifeng and Wu, Jialong and Zhou, Jingren and others},
  journal = {arXiv preprint arXiv:2510.24701},
  year    = {2025}
}

@article{team2025every1,
  title   = {Every Activation Boosted: Scaling General Reasoner to 1 Trillion Open Language Foundation},
  author  = {Ling-Team and Li, Ang and Liu, Ben and Hu, Binbin and Li, Bing and Zeng, Bingwei and Ye, Borui and Tang, Caizhi and Tian, Changxin and Huang, Chao and others},
  journal = {arXiv preprint arXiv:2510.22115},
  year    = {2025}
}

@article{ai2025ming,
  title   = {Ming-Flash-Omni: A Sparse, Unified Architecture for Multimodal Perception and Generation},
  author  = {Inclusion-AI and Ma, Bowen and Zou, Cheng and Yan, Canxiang and Jin, Chunxiang and Shen, Chunjie and Lian, Chenyu and Zheng, Dandan and Wang, Fudong and Xu, Furong and others},
  journal = {arXiv preprint arXiv:2510.24821},
  year    = {2025}
}

@article{team2025every2,
  title   = {Every step evolves: Scaling reinforcement learning for trillion-scale thinking model},
  author  = {Ling-Team and Shen, Anqi and Li, Baihui and Hu, Bin and Jing, Bin and Chen, Cai and Huang, Chao and Zhang, Chao and Yang, Chaokun and Lin, Cheng and others},
  journal = {arXiv preprint arXiv:2510.18855},
  year    = {2025}
}

@article{yadav2023ties,
  title   = {Ties-merging: Resolving interference when merging models},
  author  = {Yadav, Prateek and Tam, Derek and Choshen, Leshem and Raffel, Colin A and Bansal, Mohit},
  journal = {Advances in Neural Information Processing Systems},
  volume  = {36},
  pages   = {7093--7115},
  year    = {2023}
}

@article{li2023deep,
  title   = {Deep model fusion: A survey},
  author  = {Li, Weishi and Peng, Yong and Zhang, Miao and Ding, Liang and Hu, Han and Shen, Li},
  journal = {arXiv preprint arXiv:2309.15698},
  year    = {2023}
}

@article{zeng2025uitron,
  title   = {Uitron: Foundational gui agent with advanced perception and planning},
  author  = {Zeng, Zhixiong and Huang, Jing and Zheng, Liming and Han, Wenkang and Zhong, Yufeng and Chen, Lei and Yang, Longrong and Chu, Yingjie and He, Yuzhi and Ma, Lin},
  journal = {arXiv preprint arXiv:2508.21767},
  year    = {2025}
}

@article{xie2025gui,
  title   = {Gui-explorer: Autonomous exploration and mining of transition-aware knowledge for gui agent},
  author  = {Xie, Bin and Shao, Rui and Chen, Gongwei and Zhou, Kaiwen and Li, Yinchuan and Liu, Jie and Zhang, Min and Nie, Liqiang},
  journal = {arXiv preprint arXiv:2505.16827},
  year    = {2025}
}

@article{wang2025ui,
  title   = {Ui-tars-2 technical report: Advancing gui agent with multi-turn reinforcement learning},
  author  = {Wang, Haoming and Zou, Haoyang and Song, Huatong and Feng, Jiazhan and Fang, Junjie and Lu, Junting and Liu, Longxiang and Luo, Qinyu and Liang, Shihao and Huang, Shijue and others},
  journal = {arXiv preprint arXiv:2509.02544},
  year    = {2025}
}

@article{rafailov2023direct,
  title   = {Direct preference optimization: Your language model is secretly a reward model},
  author  = {Rafailov, Rafael and Sharma, Archit and Mitchell, Eric and Manning, Christopher D and Ermon, Stefano and Finn, Chelsea},
  journal = {Advances in neural information processing systems},
  volume  = {36},
  pages   = {53728--53741},
  year    = {2023}
}

@inproceedings{xu2025androidlab,
  title     = {Androidlab: Training and systematic benchmarking of android autonomous agents},
  author    = {Xu, Yifan and Liu, Xiao and Sun, Xueqiao and Cheng, Siyi and Yu, Hao and Lai, Hanyu and Zhang, Shudan and Zhang, Dan and Tang, Jie and Dong, Yuxiao},
  booktitle = {Proceedings of the 63rd Annual Meeting of the Association for Computational Linguistics (Volume 1: Long Papers)},
  pages     = {2144--2166},
  year      = {2025}
}

@article{deng2023mind2web,
  title   = {Mind2web: Towards a generalist agent for the web},
  author  = {Deng, Xiang and Gu, Yu and Zheng, Boyuan and Chen, Shijie and Stevens, Sam and Wang, Boshi and Sun, Huan and Su, Yu},
  journal = {Advances in Neural Information Processing Systems},
  volume  = {36},
  pages   = {28091--28114},
  year    = {2023}
}

@misc{vteam2025glm45vglm41vthinkingversatilemultimodal,
  title         = {GLM-4.5V and GLM-4.1V-Thinking: Towards Versatile Multimodal Reasoning with Scalable Reinforcement Learning},
  author        = {V-Team and Wenyi Hong and Wenmeng Yu and Xiaotao Gu and Guo Wang and Guobing Gan and Haomiao Tang and Jiale Cheng and Ji Qi and Junhui Ji and Lihang Pan and Shuaiqi Duan and Weihan Wang and Yan Wang and Yean Cheng and Zehai He and Zhe Su and Zhen Yang and Ziyang Pan and Aohan Zeng and Baoxu Wang and Bin Chen and Boyan Shi and Changyu Pang and Chenhui Zhang and Da Yin and Fan Yang and Guoqing Chen and Jiazheng Xu and Jiale Zhu and Jiali Chen and Jing Chen and Jinhao Chen and Jinghao Lin and Jinjiang Wang and Junjie Chen and Leqi Lei and Letian Gong and Leyi Pan and Mingdao Liu and Mingde Xu and Mingzhi Zhang and Qinkai Zheng and Sheng Yang and Shi Zhong and Shiyu Huang and Shuyuan Zhao and Siyan Xue and Shangqin Tu and Shengbiao Meng and Tianshu Zhang and Tianwei Luo and Tianxiang Hao and Tianyu Tong and Wenkai Li and Wei Jia and Xiao Liu and Xiaohan Zhang and Xin Lyu and Xinyue Fan and Xuancheng Huang and Yanling Wang and Yadong Xue and Yanfeng Wang and Yanzi Wang and Yifan An and Yifan Du and Yiming Shi and Yiheng Huang and Yilin Niu and Yuan Wang and Yuanchang Yue and Yuchen Li and Yutao Zhang and Yuting Wang and Yu Wang and Yuxuan Zhang and Zhao Xue and Zhenyu Hou and Zhengxiao Du and Zihan Wang and Peng Zhang and Debing Liu and Bin Xu and Juanzi Li and Minlie Huang and Yuxiao Dong and Jie Tang},
  year          = {2025},
  eprint        = {2507.01006},
  archiveprefix = {arXiv},
  primaryclass  = {cs.CV},
  url           = {https://arxiv.org/abs/2507.01006}
}

@article{comanici2025gemini,
  title   = {Gemini 2.5: Pushing the frontier with advanced reasoning, multimodality, long context, and next generation agentic capabilities},
  author  = {Comanici, Gheorghe and Bieber, Eric and Schaekermann, Mike and Pasupat, Ice and Sachdeva, Noveen and Dhillon, Inderjit and Blistein, Marcel and Ram, Ori and Zhang, Dan and Rosen, Evan and others},
  journal = {arXiv preprint arXiv:2507.06261},
  year    = {2025}
}

@article{yan2025step,
  title   = {Step-gui technical report},
  author  = {Yan, Haolong and Wang, Jia and Huang, Xin and Shen, Yeqing and Meng, Ziyang and Fan, Zhimin and Tan, Kaijun and Gao, Jin and Shi, Lieyu and Yang, Mi and others},
  journal = {arXiv preprint arXiv:2512.15431},
  year    = {2025}
}

@article{ye2025mobile,
  title   = {Mobile-agent-v3: Fundamental agents for gui automation},
  author  = {Ye, Jiabo and Zhang, Xi and Xu, Haiyang and Liu, Haowei and Wang, Junyang and Zhu, Zhaoqing and Zheng, Ziwei and Gao, Feiyu and Cao, Junjie and Lu, Zhengxi and others},
  journal = {arXiv preprint arXiv:2508.15144},
  year    = {2025}
}

@misc{hai2025holo2modelfamily,
  title  = {Holo2 - Open Foundation Models for Navigation and Computer Use Agents},
  author = {H-Company},
  year   = {2025},
  url    = https://huggingface.co/collections/Hcompany/holo2
}

@article{grattafiori2024llama,
  title   = {The llama 3 herd of models},
  author  = {Grattafiori, Aaron and Dubey, Abhimanyu and Jauhri, Abhinav and Pandey, Abhinav and Kadian, Abhishek and Al-Dahle, Ahmad and Letman, Aiesha and Mathur, Akhil and Schelten, Alan and Vaughan, Alex and others},
  journal = {arXiv preprint arXiv:2407.21783},
  year    = {2024}
}

@article{hurst2024gpt,
  title   = {Gpt-4o system card},
  author  = {Hurst, Aaron and Lerer, Adam and Goucher, Adam P and Perelman, Adam and Ramesh, Aditya and Clark, Aidan and Ostrow, AJ and Welihinda, Akila and Hayes, Alan and Radford, Alec and others},
  journal = {arXiv preprint arXiv:2410.21276},
  year    = {2024}
}

@article{glm2024chatglm,
  title   = {Chatglm: A family of large language models from glm-130b to glm-4 all tools},
  author  = {GLM, Team and Zeng, Aohan and Xu, Bin and Wang, Bowen and Zhang, Chenhui and Yin, Da and Zhang, Dan and Rojas, Diego and Feng, Guanyu and Zhao, Hanlin and others},
  journal = {arXiv preprint arXiv:2406.12793},
  year    = {2024}
}

@article{dai2025advancing,
  title   = {Advancing mobile gui agents: A verifier-driven approach to practical deployment},
  author  = {Dai, Gaole and Jiang, Shiqi and Cao, Ting and Li, Yuanchun and Yang, Yuqing and Tan, Rui and Li, Mo and Qiu, Lili},
  journal = {arXiv preprint arXiv:2503.15937},
  year    = {2025}
}

@article{xiao2025ui,
  title   = {UI-Genie: A Self-Improving Approach for Iteratively Boosting MLLM-based Mobile GUI Agents},
  author  = {Xiao, Han and Wang, Guozhi and Chai, Yuxiang and Lu, Zimu and Lin, Weifeng and He, Hao and Fan, Lue and Bian, Liuyang and Hu, Rui and Liu, Liang and others},
  journal = {arXiv preprint arXiv:2505.21496},
  year    = {2025}
}

@article{li2025mobileuse,
  title   = {MobileUse: A GUI Agent with Hierarchical Reflection for Autonomous Mobile Operation},
  author  = {Li, Ning and Qu, Xiangmou and Zhou, Jiamu and Wang, Jun and Wen, Muning and Du, Kounianhua and Lou, Xingyu and Peng, Qiuying and Zhang, Weinan},
  journal = {arXiv preprint arXiv:2507.16853},
  year    = {2025}
}

@article{team2024gemini,
  title   = {Gemini 1.5: Unlocking multimodal understanding across millions of tokens of context},
  author  = {Gemini-Team and Georgiev, Petko and Lei, Ving Ian and Burnell, Ryan and Bai, Libin and Gulati, Anmol and Tanzer, Garrett and Vincent, Damien and Pan, Zhufeng and Wang, Shibo and others},
  journal = {arXiv preprint arXiv:2403.05530},
  year    = {2024}
}

@misc{zhang2026omegausebuildinggeneralpurposegui,
  title         = {OmegaUse: Building a General-Purpose GUI Agent for Autonomous Task Execution},
  author        = {Le Zhang and Yixiong Xiao and Xinjiang Lu and Jingjia Cao and Yusai Zhao and Jingbo Zhou and Lang An and Zikan Feng and Wanxiang Sha and Yu Shi and Congxi Xiao and Jian Xiong and Yankai Zhang and Hua Wu and Haifeng Wang},
  year          = {2026},
  eprint        = {2601.20380},
  archiveprefix = {arXiv},
  primaryclass  = {cs.AI},
  url           = {https://arxiv.org/abs/2601.20380}
}

@misc{openai2024computeruse,
  title        = {Computer Using Agent},
  author       = {{OpenAI}},
  year         = {2025a},
  howpublished = {\url{https://platform.openai.com/docs/guides/tools-computer-use}}
}

@misc{anthropic2025claude37,
  title        = {Claude-3-7-Sonnet},
  author       = {{Anthropic}},
  year         = {2025a},
  howpublished = {\url{https://www.anthropic.com/news/claude-3-7-sonnet}}
}

@article{zhang2025mvp,
  title={MVP: Multiple View Prediction Improves GUI Grounding},
  author={Zhang, Yunzhu and Pan, Zeyu and Zeng, Zhengwen and Shen, Shuheng and Meng, Changhua and Zhu, Linchao},
  journal={arXiv preprint arXiv:2512.08529},
  year={2025}
}

@article{jiang2025zoom,
  title={Zoom in, Click out: Unlocking and Evaluating the Potential of Zooming for GUI Grounding},
  author={Jiang, Zhiyuan and Xie, Shenghao and Li, Wenyi and Zu, Wenqiang and Li, Peihang and Qiu, Jiahao and Pei, Siqi and Ma, Lei and Huang, Tiejun and Wang, Mengdi and others},
  journal={arXiv preprint arXiv:2512.05941},
  year={2025}
}

@article{qiu2026unified,
  title={Unified Generation and Self-Verification for Vision-Language Models via Advantage Decoupled Preference Optimization},
  author={Qiu, Xinyu and Jia, Heng and Zeng, Zhengwen and Shen, Shuheng and Meng, Changhua and Yang, Yi and Zhu, Linchao},
  journal={arXiv preprint arXiv:2601.01483},
  year={2026}
}

\newpage

\begin{appendix}

	\section{Action Space and Prompt Templates}

\subsection{Action Space}
\begin{table*}[ht]
\centering
\renewcommand{\arraystretch}{1.1} 
\scalebox{0.8}{
\begin{tabular}{ll}
\toprule
\textbf{Action} &\textbf{Definition}\\
\midrule
    Click(box=(x, y)) & Click at coordinates (x, y).\\
    Drag(start=(x1, y1), end=(x2, y2)) & Drag from (x1, y1) to (x2, y2).\\
    Scroll(start=(x1, y1), end=(x2, y2), direction=`') & Scroll from (x1, y1) to (x2, y2) with specified direction.\\
   Type(content=`')   & Type the specified content.\\
   Launch(app=`')   & Launch the specified app.\\
   Wait()   & Wait for loading.\\
   Finished(content=`')   & Finish the task, with optional information.\\
   CallUser(content=`')   & Conclude the answer for information-retrieval.\\
   LongPress(box=(x, y))    & Long press at coordinates (x, y).\\
   PressBack()    & Press the `back' button.\\
   PressHome()   & Press the `home' button.\\
   PressEnter() & Press the `enter' button.\\
   PressRecent() & Press the `recent' button.\\
   Hover(box=(x,y))& Move the mouse cursor to coordinates (x, y) without clicking. \\
   DoubleClick(box=(x,y))& Perform a double-click at coordinates (x, y).\\
   Hotkey(keys=[`ctrl', `c'])& Press the specified key combination (e.g., Ctrl+C for copy).\\
\bottomrule
\end{tabular}
}
\caption{All actions and their definitions used in \textbf{UI-Venus-1.5}. We unify the action space and map all the actions in the existing open-source dataset to this space. }
\label{tab:action_space}
\end{table*}

\subsection{Grounding}

\begin{tcolorbox}[
		title=Grounding Prompt]\label{grounding_prompt}
	Output the center point of the position corresponding to the following instruction: \texttt{\color{red} \{problem\}}. The output should just be the coordinates of a point, in the format [x,y]. Additionally, if the task is infeasible (e.g., the task is not related to the image), the output should be [-1,-1].
\end{tcolorbox}

\subsection{Mobile}

\begin{tcolorbox}[
		title=Mobile Prompt,breakable]\label{mobile_prompt}
	**You are a GUI Agent**.

	Your task is to analyze a given user task, review current screenshot and previous actions, and determine the next action to complete the task.

	\medskip

	\#\#\# Available Actions

	You may execute one of the following functions:

	- Click(box=(x1,y1))

	- Drag(start=(x1,y1), end=(x2,y2))

	- Scroll(start=(x1,y1), end=(x2,y2))

	- Type(content=`')

	- Launch(app=`')

	- Wait()

	- Finished(content=`')

	- CallUser(content=`')

	- LongPress(box=(x1,y1))

	- PressBack()

	- PressHome()

	- PressEnter()

	- PressRecent()

	\medskip

	\#\#\# User Task

	\texttt{\color{red} \{problem\}}

	\medskip

	\#\#\# Previous Actions

	\texttt{\color{red} \{previous\_actions\}}

	\medskip

	\#\#\# Output Format

	<think> your thinking process </think>

	<action> the next action </action>

	<conclusion> the conclusion about the next action </conclusion>

	\medskip

	\#\#\# Instruction

	- Make sure you understand the task goal to avoid wrong actions.

	- Make sure you carefully examine the the current screenshot. Sometimes the summarized history might not be reliable, over-claiming some effects.

	- For requests that are questions (or chat messages), remember to use the `CallUser' action to reply to user explicitly before finishing! Then, after you have replied, use the Finished action if the goal is achieved.

	- Consider exploring the screen by using the `scroll' action with different directions to reveal additional content.

	- To copy some text: first select the exact text you want to copy, which usually also brings up the text selection bar, then click the `copy' button in bar.

	- To paste text into a text box, first long press the text box, then usually the text selection bar will appear with a `paste' button in it.
\end{tcolorbox}

\subsection{Web}

\begin{tcolorbox}[
		title=Web Prompt,breakable]\label{web_prompt}
	**You are a GUI Agent**.

	Your task is to analyze a given user task, review current screenshot and previous actions, and determine the next action to complete the task.

	\medskip

	\#\#\# Available Actions

	You may execute one of the following functions:

	- Click(box=(x1,y1))

	- Drag(start=(x1,y1), end=(x2,y2))

	- Scroll(direction='down or up')

	- Type(content=`')

	- Launch(url=`')

	- Wait()

	- Finished(content=`')

	- CallUser(content=`')

	- LongPress(box=(x1,y1))

	- PressBack()

	- PressHome()

	- PressEnter()

	- PressRecent()

	- Hover(box=(x1,y1))

	- DoubleClick(box=(x1,y1))

	- Hotkey(keys=[`ctrl', `c']) \# Split keys with comma and wrap each key in single quotes. Do not use more than 3 keys in one Hotkey action.

	\medskip

	\#\#\# User Task

	\texttt{\color{red} \{problem\}}

	\medskip

	\#\#\# Previous Actions

	\texttt{\color{red} \{previous\_actions\}}

	\medskip

	\#\#\# Output Format

	<think> your thinking process </think>

	<action> the next action </action>

	<conclusion> the conclusion about the next action </conclusion>

	\medskip

	\#\#\# Instruction

	- Make sure you understand the task goal to avoid wrong actions.

	- Make sure you carefully examine the the current screenshot. Sometimes the summarized history might not be reliable, over-claiming some effects.

	- For requests that are questions (or chat messages), remember to use the `CallUser' action to reply to user explicitly before finishing! Then, after you have replied, use the Finished action if the goal is achieved.

	- Consider exploring the screen by using the `scroll' action with different directions to reveal additional content.
\end{tcolorbox}

\subsection{Chinese APPs Prompt}

\begin{CJK}{UTF8}{gbsn}
	\begin{tcolorbox}[
			title=Chinese APP Prompt,breakable]\label{zh_prompt}
		**你是一个手机图形界面智能体代理**

		你的任务是根据历史操作和当前设备状态去执行一系列操作来完成用户的任务。

		\medskip

		\#\#\# 你可以用的操作以及对应功能如下:
		- Click(box=(x1,y1))

		>>点击操作，点击屏幕上的指定位置。坐标区间从左上角(0,0)到右下角(999,999)。

		- Drag(start=(x1,y1), end=(x2,y2))

		>>拖动操作，从起始坐标长按数秒之后拖动到结束坐标。用于调整app布局，滑动滑块验证码等。

		- Scroll(start=(x1,y1), end=(x2,y2))

		>>滑动操作，从起始坐标拖动到结束坐标。用于滚动查找内容，切换选项卡，下拉通知栏等。坐标区间从左上角(0,0)到右下角(999,999)。

		- Type(content=`')

		>>输入操作，在当前激活的输入框输入指定内容。

		- Launch(app=`')

		>>启动目标app。当目标app在当前界面不可见时，可以使用该动作打开app。

		- Wait()

		>>等待页面加载。

		- Finished(content=`')

		>>任务结束，退出设备接管。

		- CallUser(content=`')

		>>回答用户的问题或者当前界面有多个符合要求的选项时需要用户接管。

		- LongPress(box=(x1,y1))

		>>长按操作，在指定位置长按一定的时间。该操作可以触发更多功能选项，例如复制、转发消息，删除等。坐标区间从左上角(0,0)到右下角(999,999)。

		- PressBack()

		>>返回上一个界面，一般用于错误回退或继续执行剩余任务。

		- PressHome()

		>>返回系统桌面，一般用于跨app任务中快速打开下一个app或遇到严重错误时回退到系统桌面。

		- PressEnter()

		>>回车操作，用于换行或者在搜索框中输入内容之后执行搜索操作。

		- PressRecent()

		>>打开系统后台界面。
		\medskip

		\#\#\# 用户任务

		\texttt{\color{red} \{problem\}}

		\medskip

		\#\#\# 先前的动作和推理过程

		\texttt{\color{red} \{previous\_actions\}}

		\medskip

		\#\#\# 输出格式

		<think>你的思考过程</think>

		<action>执行的操作</action>

		<conclusion>总结当前操作</conclusion>

		\medskip

		\#\#\# Instruction

		-输入内容之前，确保输入框已经被激活（出现键盘或者 `ADB Keyboard {ON}'字样代表输入框已经激活）。

		-在app内找不到任务要求的入口时，尝试使用搜索功能，或者如果当前页面上方存在选多个项卡，尝试使用Scroll操作查看。

		-如果在执行任务的过程中进入到和任务无关的界面，使用PressBack进行回退。

		-任务结束之前，确保已经完整准确地完成用户的任务，如果存在漏做、错做的内容，需要返回重新执行。
	\end{tcolorbox}

\end{CJK}

\section{Experiment Details of All Grounding Benchmarks}
Note that in OSWorld-G, we re-calculate the performance of UI-Venus-1.0 with the refusal task, and thus, its results are different from our previous report.

\begin{table}[ht]
	\centering
	\footnotesize
	\setlength{\tabcolsep}{0pt}
	\begin{tabular*}{\columnwidth}{@{\extracolsep{\fill}}l *{9}{c}}
		\toprule
		\multirow{2}{*}{\textbf{Models}} &
		\multicolumn{4}{c}{\textbf{Basic Tasks}} &
		\multicolumn{4}{c}{\textbf{Advanced Tasks}} &
		\multirow{2}{*}{\textbf{Overall}} \\
		\cmidrule(lr){2-5} \cmidrule(lr){6-9}
		& \textbf{Element} & \textbf{Visual} & \textbf{Spatial} & \textbf{Avg} &
		\textbf{Reasoning} & \textbf{Functional} & \textbf{Refusal} & \textbf{Avg} & \\
		\midrule
		\rowcolor{gray!15}
		\multicolumn{10}{l}{\textit{General VLMs}} \\
		Qwen3-VL-2B*~\citep{Qwen3-VL}              & 66.6 & 79.7 & 61.3 & 68.2 & 12.6 & 41.6 & 0.0 & 15.9 & 45.2 \\
		Qwen3-VL-8B*~\citep{Qwen3-VL}             & 73.9 & 83.8 & 75.5 & 76.8 & 22.6 & 61.3 & 6.8 & 27.3 & 55.1 \\
		Qwen3-VL-30B-A3B*~\citep{Qwen3-VL}          & 68.0 & 84.3 & 69.2 & 72.3 & 19.9 & 58.6 & 11.3 & 27.0 & 52.4 \\
		\midrule
		\rowcolor{gray!15}
		\multicolumn{10}{l}{\textit{GUI-specific Models}} \\
		OpenCUA-7B~\citep{wang2025opencuaopenfoundationscomputeruse}                     & {62.23} & {84.39} & 67.44 & {69.15} & 21.32 & 49.14 & 0.00 & 21.43 & 48.20 \\
		OpenCUA-32B~\citep{wang2025opencuaopenfoundationscomputeruse}                     & {65.49} & 78.55 & {68.80} & {69.64} & 29.09 & {51.00} & 0.00 & 25.08 & {50.08} \\
		GTA1-7B~\citep{yang2025gta1guitesttimescaling}                          & 63.73 & 76.64 & 57.05 & 64.87 & 23.31 & 51.14 & 0.00 & 22.75 & 46.38 \\
		GTA1-32B~\citep{yang2025gta1guitesttimescaling}                         & 75.36 & 88.08 & 76.77 & 78.87 & 38.84 & 67.14 & 0.00 & 33.25 & 58.84 \\
		UI-Venus-1.0-7B~\citep{gu2025ui}                  & 64.30 & 78.78 & 67.15 & 68.66 & 24.39 & 53.85 & 0.00 & 23.90 & 49.01 \\
		UI-Venus-1.0-72B~\citep{gu2025ui}                 & {81.58} & {91.30} & {78.81} & {83.12} & \underline{46.16} & {68.86} & {51.33} & {53.75} & {70.23} \\
		Holo2-8B*~\citep{hai2025holo2modelfamily}                         & 71.4 & 85.8 & 77.9 & 76.8 & 34.0 & 63.1 & 0.0 & 30.2 & 56.4 \\
		Holo2-30B-A3B*~\citep{hai2025holo2modelfamily}                    & 78.1 & 89.7 & 81.0 & 81.8 & 32.2 & 68.7 & 0.0 & 31.0 & 59.5 \\
		Step-GUI-4B*~\citep{yan2025step}                      & 73.9 & 81.8 & 77.9 & 77.0 & 26.1 & 59.0 & 0.0 & 25.9 & 54.6 \\
		MAI-UI-2B*~\citep{zhou2025mai}                        & 72.8 & 87.1 & 76.6 & 77.4 & 27.3 & 62.3 & 0.0 & 27.3 & 55.4 \\
		MAI-UI-8B*~\citep{zhou2025mai}                        & 81.3 & 90.8 & 84.5 & 84.5 & \textbf{55.4} & \underline{69.1} & 0.0 & 40.5 & 65.2 \\
		\midrule
		\rowcolor{gray!15}
		\multicolumn{10}{l}{\textit{Ours}} \\
		UI-Venus-1.5-2B                       & 79.4 & 85.8 & 80.9 & 81.4 & 22.0 & 57.3 & \textbf{76.3} & 49.2 & 67.3 \\
		UI-Venus-1.5-8B                       & \underline{84.2} & \underline{93.1} & \underline{84.9} & \underline{86.6} & 38.1 & \textbf{70.1} & 61.6 & \underline{54.2} & \underline{72.3} \\
		UI-Venus-1.5-30B-A3B                      & \textbf{85.1} & \textbf{93.2} & \textbf{86.4} & \textbf{87.5} & 41.8 & 68.1 & \underline{73.1} & \textbf{59.0} & \textbf{75.0} \\
		\bottomrule
	\end{tabular*}
	\vspace{-0.4em}
	\caption{Performance comparison on \textbf{VenusBench-GD}. For each benchmark, the best and second-best performing models are indicated in \textbf{bold} and \underline{underlined}, respectively. Asterisk (*) indicates results that may require verification with original sources.}
	\label{tab:venusbench-gd}
\end{table}

\begin{table*}[t]
	\centering
	\footnotesize
	\setlength{\tabcolsep}{0pt}
	\begin{tabular*}{\textwidth}{@{\extracolsep{\fill}}l *{15}{c}}
		\toprule
		\multirow{3}{*}{\textbf{Model}} & \multicolumn{2}{c}{\textbf{CAD}} & \multicolumn{2}{c}{\textbf{Dev}} & \multicolumn{2}{c}{\textbf{Creative}} & \multicolumn{2}{c}{\textbf{Scientific}} & \multicolumn{2}{c}{\textbf{Office}} & \multicolumn{2}{c}{\textbf{OS}} & \multicolumn{3}{c}{\textbf{Avg.}} \\
		\cmidrule(lr){2-3} \cmidrule(lr){4-5} \cmidrule(lr){6-7} \cmidrule(lr){8-9} \cmidrule(lr){10-11} \cmidrule(lr){12-13} \cmidrule(lr){14-16}
		& Text & Icon & Text & Icon & Text & Icon & Text & Icon & Text & Icon & Text & Icon & Text & Icon & \textbf{Avg.} \\
		\midrule
		\rowcolor{gray!15}
		\multicolumn{16}{l}{\textit{General VLMs}} \\
		Seed1.8~\citep{seed1.8}                         & - & - & - & - & - & - & - & - & - & - & - & - & - & - & 64.3 \\
		Qwen3-VL-2B*~\citep{Qwen3-VL}              & 27.9 & 10.9 & 57.1 & 9.7 & 58.1 & 16.8 & 62.5 & 22.7 & 73.4 & 34.0 & 55.1 & 19.1 & 55.0 & 17.4 & {40.6} \\
		Qwen3-VL-8B*~\citep{Qwen3-VL}             & 56.9 & 10.9 & 75.3 & 22.8 & 68.2 & 16.1 & 78.5 & 32.7 & 80.8 & 39.6 & 71.0 & 20.2 & 71.1 & 22.8 & 52.7 \\
		Qwen3-VL-30B-A3B*~\citep{Qwen3-VL}          & 51.8 & 15.6 & 76.0 & 24.8 & 69.2 & 20.3 & 76.4 & 27.3 & 80.8 & 37.7 & 75.7 & 38.2 & 70.6 & 26.3 & 53.7 \\
		\midrule
		\rowcolor{gray!15}
		\multicolumn{16}{l}{\textit{GUI-specific Models}} \\
		OpenCUA-7B~\citep{wang2025opencuaopenfoundationscomputeruse}                     & - & - & - & - & - & - & - & - & - & - & - & - & - & - & 50.0 \\
		OpenCUA-32B~\citep{wang2025opencuaopenfoundationscomputeruse}                     & - & - & - & - & - & - & - & - & - & - & - & - & - & - & 55.3 \\
		OpenCUA-72B~\citep{wang2025opencuaopenfoundationscomputeruse}                     & - & - & - & - & - & - & - & - & - & - & - & - & - & - & 60.8 \\
		GTA1-7B~\citep{yang2025gta1guitesttimescaling}                          & 66.9   & 20.7   & 62.6   & 18.2   & 53.3   & 17.2   & 76.4 & 31.8   & 82.5   & 50.9   & 48.6  & 25.9   & 65.5   & 25.2   & 50.1 \\
		GTA1-32B~\citep{yang2025gta1guitesttimescaling}                         & \textbf{83.1} & {37.9} & {72.2} & {25.9} & {70.1} & 31.3 & {84.7} & {39.1} & {89.3} & {64.2} & {76.6} & {51.7} & {78.9} & {38.9} & {63.6} \\
		GUI-Owl-7B~\citep{ye2025mobile}                       & 64.5 & 21.9 & 76.6 & 31.0 & 59.6 & 27.3 & 79.1 & 37.3 & 77.4 & 39.6 & 59.8 & 33.7 & - & - & 54.9 \\
		GUI-Owl-32B~\citep{ye2025mobile}                      & 62.4 & 28.1 & {84.4} & 39.3 & 65.2 & 18.2 & 82.6 & 39.1 & 81.4 & 39.6 & 70.1 & 36.0 & - & - & 58.0\\
		UI-Venus-1.0-7B~\citep{gu2025ui}                  & {60.4} & 21.9  & {74.7} & 24.1  & 63.1  & {14.7} & {76.4} & 31.8  & 75.7  & 41.5  & {49.5} & 22.5  & {67.1} & 24.3  & 50.8 \\
		UI-Venus-1.0-72B~\citep{gu2025ui}                 & {66.5}  & {29.7}  & {84.4}  & {33.1} & {73.2} & {30.8}  & {84.7}  & {42.7} & {83.1} & {60.4}  & {75.7}  & {36.0} & {77.4}  & {36.8} & {61.9} \\
		Holo2-8B~\citep{hai2025holo2modelfamily}                         & - & - & - & - & - & - & - & - & - & - & - & - & - & - & 58.9 \\
		Holo2-30B-A3B~\citep{hai2025holo2modelfamily}                    & - & - & - & - & - & - & - & - & - & - & - & - & - & - & 66.1 \\
		Step-GUI-4B~\citep{yan2025step}                      & - & - & - & - & - & - & - & - & - & - & - & - & - & - & 60.0 \\
		Step-GUI-8B~\citep{yan2025step}                      & - & - & - & - & - & - & - & - & - & - & - & - & - & - & 62.6 \\
		MAI-UI-2B~\citep{zhou2025mai}                        &61.4 &23.4 &76.6 &32.4 &69.2 &21.7 &81.2 &34.5 &85.9 &39.6 &68.2 &41.6 & - & -&57.4 \\
		MAI-UI-8B~\citep{zhou2025mai}                        &72.6 &35.9 &83.8 &52.4 &\underline{76.3} &33.6 &79.9 &37.3 & 88.7 &60.4 &76.6 &49.4 & - & -& 65.8\\
		MAI-UI-32B~\citep{zhou2025mai}                       &70.1 & \textbf{45.3} & \underline{86.4} &40.7 & \textbf{82.8} &\underline{37.8} & \textbf{91.7} &\underline{46.4} & \underline{90.4} &\textbf{71.7} & 78.5 &34.8 & - & -& 67.9\\
		\midrule
		\rowcolor{gray!15}
		\multicolumn{16}{l}{\textit{Ours}} \\
		UI-Venus-1.5-2B                       & 54.3 & 32.8 & 70.1 & 43.4 & 63.6 & 28.7 & 76.4 & 38.2 & 81.9 & 47.2 & 73.8 & 51.7 & 69.1 & 39.4 & 57.7 \\
		UI-Venus-1.5-8B                       & \underline{75.1} & 31.2 & 85.7 & \underline{54.5} & 75.3 & 32.9 & \underline{86.1} & 44.5 & \textbf{92.7} & 66.0 & \underline{82.2} & \underline{52.8} & \textbf{82.4} & \underline{45.9} & \underline{68.4} \\
		UI-Venus-1.5-30B-A3B                      & 70.6 & \underline{40.6} & \textbf{87.7} & \textbf{57.9} & 75.8 & \textbf{41.3} & 84.0 & \textbf{47.3} & 89.8 & \underline{69.8} & \textbf{83.2} & \textbf{56.2} & \underline{81.2} & \textbf{51.0} & \textbf{69.6} \\
		\bottomrule
	\end{tabular*}
	\caption{Performance comparison on \textbf{ScreenSpot-Pro}. For each benchmark, the best and second-best performing models are indicated in \textbf{bold} and \underline{underlined}, respectively. Asterisk (*) indicates results that may require verification with original sources.}
	\label{tab:screenspot_pro}
\end{table*}

\begin{table}[ht]
	\centering
	\footnotesize
	\setlength{\tabcolsep}{0pt}
	\begin{tabular*}{\columnwidth}{@{\extracolsep{\fill}}l *{7}{c}}
		\toprule
		\multirow{2}{*}{\textbf{Models}} & \multicolumn{2}{c}{\textbf{Mobile}}  & \multicolumn{2}{c}{\textbf{Desktop}} & \multicolumn{2}{c}{\textbf{Web}}     & \multirow{2}{*}{\textbf{Avg}} \\
		\cmidrule(lr){2-3} \cmidrule(lr){4-5} \cmidrule(lr){6-7}
		& \textbf{Text} & \textbf{Icon/Widget} & \textbf{Text} & \textbf{Icon/Widget} & \textbf{Text} & \textbf{Icon/Widget} &                               \\
		\midrule
		\rowcolor{gray!15}
		\multicolumn{8}{l}{\textit{General VLMs}} \\
		Qwen3-VL-2B*~\citep{Qwen3-VL}              & 94.1 & 80.6 & 94.8 & 74.3 & 89.7 & 72.9 & 85.6 \\
		Qwen3-VL-8B*~\citep{Qwen3-VL}             & \textbf{99.7} & 87.7 & 94.8 & 83.6 & 95.3 & 85.7 & 92.1 \\
		Qwen3-VL-30B-A3B*~\citep{Qwen3-VL}          & 99.0 & 87.7 & 95.4 & 82.9 & 95.3 & 83.7 & 91.7 \\
		\midrule
		\rowcolor{gray!15}
		\multicolumn{8}{l}{\textit{GUI-specific Models}} \\
		OpenCUA-7B~\citep{wang2025opencuaopenfoundationscomputeruse}                     & - & -  & - & - & - & - & 92.3 \\
		OpenCUA-32B~\citep{wang2025opencuaopenfoundationscomputeruse}                     & - & -  & - & - & - & - & 93.4 \\
		OpenCUA-72B~\citep{wang2025opencuaopenfoundationscomputeruse}                     & - & -  & - & - & - & - & 92.9 \\
		GTA1-7B~\citep{yang2025gta1guitesttimescaling}                          & 99.0 & 88.6 & 94.9 & 89.3 & 92.3 & 86.7 & 92.4  \\
		GTA1-32B~\citep{yang2025gta1guitesttimescaling}                         & \textbf{99.7} & 90.5 & \underline{99.0} & \textbf{94.3} & 95.7 & 90.1 & 95.2 \\
		GUI-Owl-7B~\citep{ye2025mobile}                       & 99.0 & 92.4 & 96.9 & 85.0 & 93.6 & 85.2 & 92.8 \\
		GUI-Owl-32B~\citep{ye2025mobile}                      & 98.6 & 90.0 & 97.9 & 87.8 & 94.4 & 86.7 & 93.2 \\
		UI-Venus-1.0-7B~\citep{gu2025ui}                  & 99.0  & 90.0         & 97.0  & 90.7 & 96.2 & 88.7       &      94.1               \\
		UI-Venus-1.0-72B~\citep{gu2025ui}                 & \textbf{99.7}  & \underline{93.8}        & 95.9  & 90.0 & 96.2 & 92.6         & 95.3      \\
		Holo2-8B~\citep{hai2025holo2modelfamily}                         & - & -  & - & - & - & - & 93.2 \\
		Holo2-30B-A3B~\citep{hai2025holo2modelfamily}                    & - & -  & - & - & - & - & 94.9 \\
		Step-GUI-4B~\citep{yan2025step}                      & - & -  & - & - & - & - & 93.6 \\
		Step-GUI-8B~\citep{yan2025step}                      & - & -  & - & - & - & - & 95.1 \\
		MAI-UI-2B~\citep{zhou2025mai}                        & \underline{99.3} &87.2 &97.4 &88.6 &94.0 &84.7 &92.5\\
		MAI-UI-8B~\citep{zhou2025mai}                        &\underline{99.3} &89.1 &\underline{99.0} &92.1 &\underline{97.9} &91.1 &95.2\\
		MAI-UI-32B~\citep{zhou2025mai}                       &99.0 &92.9 &\textbf{99.5} &\underline{93.6} &97.4 &\textbf{94.6} &\textbf{96.5}\\
		\midrule
		\rowcolor{gray!15}
		\multicolumn{8}{l}{\textit{Ours}} \\
		UI-Venus-1.5-2B                       & 98.6 & 91.0 & 93.3 & 92.9 & 93.2 & 85.7 & 92.8 \\
		UI-Venus-1.5-8B                       & \underline{99.3} & 92.9 & 96.4 & 92.9 & \textbf{98.3} & \underline{93.1} & 95.9 \\
		UI-Venus-1.5-30B-A3B                      & \underline{99.3} & \textbf{94.8} & 95.9 & \textbf{94.3} & \underline{97.9} & \underline{93.1} & \underline{96.2} \\
		\bottomrule
	\end{tabular*}
	\vspace{0.4em}
	\caption{Performance comparison on \textbf{ScreenSpot-V2}. For each benchmark, the best and second-best performing models are indicated in \textbf{bold} and \underline{underlined}, respectively. Asterisk (*) indicates results that may require verification with original sources.}
	\label{tab:screenspot-v2}
\end{table}

\begin{table*}[t]
	\centering
	\footnotesize
	\setlength{\tabcolsep}{0pt}
	\begin{tabular*}{\textwidth}{@{\extracolsep{\fill}}l *{13}{c}}
		\toprule
		\multirow{2}{*}{\textbf{Model}} &
		\multicolumn{2}{c}{\textbf{Windows}} &
		\multicolumn{2}{c}{\textbf{MacOS}} &
		\multicolumn{2}{c}{\textbf{Linux}} &
		\multicolumn{2}{c}{\textbf{iOS}} &
		\multicolumn{2}{c}{\textbf{Android}} &
		\multicolumn{2}{c}{\textbf{Web}} &
		\multirow{2}{*}{\textbf{Avg.}} \\
		\cmidrule(lr){2-3}
		\cmidrule(lr){4-5}
		\cmidrule(lr){6-7}
		\cmidrule(lr){8-9}
		\cmidrule(lr){10-11}
		\cmidrule(lr){12-13}
		& Bas. & Adv. & Bas. & Adv. & Bas. & Adv. & Bas. & Adv. & Bas. & Adv. & Bas. & Adv. & \\
		\midrule
		\rowcolor{gray!15}
		\multicolumn{14}{l}{\textit{General VLMs}} \\
		Qwen3-VL-2B*~\citep{Qwen3-VL}              & 82.3 & 41.2 & 79.1 & 45.4 & 67.5 & 44.4 & 92.0 & 68.2 & 91.6 & 69.6 & 85.8 & 52.9 & 69.5 \\
		Qwen3-VL-8B*~\citep{Qwen3-VL}             & 88.6 & 62.5 & 86.1 & 66.8 & 72.8 & 57.1 & 95.9 & 83.9 & 95.8 & 84.8 & 94.8 & 72.7 & 81.4 \\
		Qwen3-VL-30B-A3B*~\citep{Qwen3-VL}          & 87.8 & 69.9 & 85.2 & 68.8 & 78.0 & 60.7 & \underline{96.5} & 84.5 & 96.3 & 88.5 & \underline{96.5} & 78.6 & 83.7 \\
		\midrule
		\rowcolor{gray!15}
		\multicolumn{14}{l}{\textit{GUI-specific Models}} \\
		GUI-Owl-7B~\citep{ye2025mobile}                       & 86.4 & 61.8 & 81.7 & 64.5 & 74.4 & 61.7& 94.9 & 83.0 & 95.8 & 83.7 & 93.2 & 72.7& 80.5 \\
		GUI-Owl-32B~\citep{ye2025mobile}                      & 85.6 & 65.1 & 84.9& 67.1 & 77.0 & 63.3& 95.2 & 85.5 & 96.1 & 87.0 & 95.5 & 80.8& 83.0 \\
		Holo2-8B~\citep{hai2025holo2modelfamily}                         & 90.8 & 70.2 & 87.5 & 71.4 & 78.5 & 60.2 & 96.2 & 88.2 & 96.3 & 87.9 & 95.5 & 77.9 & 84.5 \\
		Holo2-30B-A3B~\citep{hai2025holo2modelfamily}                    & 91.9 & 72.8 & 88.1 & 74.9 & 84.3 & 67.3 & \underline{96.5} & 89.7 & 96.3 & 90.1 & \underline{96.5} & 82.5 & 86.8 \\
		Step-GUI-4B~\citep{yan2025step}                      & - & - & - & - & - & - & - & - & - & - & - & - & 84.0 \\
		Step-GUI-8B~\citep{yan2025step}                      & - & - & - & - & - & - & - & - & - & - & - & - & 85.6 \\
		MAI-UI-2B~\citep{zhou2025mai}                        &84.9 &64.0 &89.3 &72.5 &75.4 &60.2 &95.2 &85.2 &96.3 &84.2 &92.9 &76.0 & 82.6 \\
		MAI-UI-8B~\citep{zhou2025mai}                        & 92.3 & 74.3 & \underline{90.7} & \underline{86.4} & 81.2 & 67.3 & \textbf{97.1} & 90.0 & \underline{97.5} & 92.7 & 95.8 & 86.0 & \underline{88.8}\\
		MAI-UI-32B~\citep{zhou2025mai}                       & \textbf{93.0} & \textbf{78.7} & \textbf{92.8} & \textbf{87.6} & \textbf{86.9} & \textbf{77.6} & \textbf{97.1} & \underline{92.4} & \textbf{98.0} & \underline{93.2} & 96.1 & \textbf{92.5} & \textbf{91.3}\\
		\midrule
		\rowcolor{gray!15}
		\multicolumn{14}{l}{\textit{Ours}} \\
		UI-Venus-1.5-2B                       & 88.2 & 61.8 & 82.3 & 65.6 & 77.0 & 60.7 & 92.7 & 80.3 & 93.3 & 83.9 & 94.8 & 72.1 & 80.3 \\
		UI-Venus-1.5-8B                       & \underline{92.6} & 74.6 & 86.1 & 82.1 & 84.3 & 67.3 & \textbf{97.1} & 89.4 & 96.9 & 92.4 & \textbf{96.8} & 86.0 & 88.1 \\
		UI-Venus-1.5-30B-A3B                      & 91.5 & \underline{76.5} & 88.1 & 76.6 & \underline{85.9} & \underline{69.9} & \underline{96.5} & \textbf{93.0} & 97.2 & \textbf{93.8} & \underline{96.5} & \underline{87.7} & 88.6 \\
		\bottomrule
	\end{tabular*}
	\caption{Performance comparison on \textbf{MMbench-GUI-L2}. For each benchmark, the best and second-best performing models are indicated in \textbf{bold} and \underline{underlined}, respectively. Asterisk (*) indicates results that may require verification with original sources.}
	\label{tab:mmbench_gui_l2}
\end{table*}

\begin{table*}[t]
	\centering
	\footnotesize
	\setlength{\tabcolsep}{0pt}
	\begin{tabular*}{\textwidth}{@{\extracolsep{\fill}}l *{6}{c}}
		\toprule
		\textbf{Models} &
		\textbf{\begin{tabular}[c]{@{}c@{}}Text \\ Matching\end{tabular}} &
		\textbf{\begin{tabular}[c]{@{}c@{}}Element \\ Recognition\end{tabular}} &
		\textbf{\begin{tabular}[c]{@{}c@{}}Layout \\ Understanding\end{tabular}}&
		\textbf{\begin{tabular}[c]{@{}c@{}}Fine-grained \\ Manipulation\end{tabular}}&
		\textbf{Refusal} &
		\textbf{Avg} \\
		\midrule
		\rowcolor{gray!15}
		\multicolumn{7}{l}{\textit{General VLMs}} \\
		Qwen3-VL-2B*~\citep{Qwen3-VL}              & 60.9 & 49.7 & 57.3 & 38.9 & 0.0 & 47.7 \\
		Qwen3-VL-8B*~\citep{Qwen3-VL}             & 71.6 & 59.4 & 61.3 & 49.7 & 1.9 & 57.4 \\
		Qwen3-VL-30B-A3B*~\citep{Qwen3-VL}          & 73.9 & 65.2 & 67.2 & 51.0 & 5.6 & 61.2 \\
		\midrule
		\rowcolor{gray!15}
		\multicolumn{7}{l}{\textit{GUI-specific Models}} \\
		OpenCUA-7B~\citep{wang2025opencuaopenfoundationscomputeruse}                     & - & - & - & - & - & 55.3 \\
		OpenCUA-32B~\citep{wang2025opencuaopenfoundationscomputeruse}                     & - & - & - & - & - & 59.6 \\
		GTA1-7B~\citep{yang2025gta1guitesttimescaling}                          & 42.1 & 65.7 & 62.7 & 56.1 & 0.0 & 55.1\\
		GTA1-32B~\citep{yang2025gta1guitesttimescaling}                         & 63.2 & \textbf{78.4} & 73.3 & \textbf{65.2} & 0.0 & 65.2\\
		GUI-Owl-7B~\citep{ye2025mobile}                       & 64.8 & 63.6 & 61.3 & 41.0 & - & 55.9 \\
		GUI-Owl-32B~\citep{ye2025mobile}                      & 67.0 & 64.5 & 67.2 & 45.6 & - & 58.0 \\
		UI-TARS-1.5-7B~\citep{ui-tars-15-seed}                   & - & - & - & - & - & 52.8 \\
		UI-Venus-1.0-7B~\citep{gu2025ui}                 & 74.6 & 60.5 & 61.5 & 45.5 & - & 54.6 \\
		UI-Venus-1.0-72B~\citep{gu2025ui}                 & \textbf{82.1} & 71.2 & 70.7 & \underline{64.4} & - & 62.2 \\
		Holo2-8B*~\citep{hai2025holo2modelfamily}                         & 74.3 & 68.2 & 67.6 & 59.1 & 0.0 & 63.5 \\
		Holo2-30B-A3B*~\citep{hai2025holo2modelfamily}                    & 77.0 & 68.8 & 70.0 & 59.7 & 0.0 & 65.2 \\
		Step-GUI-4B*~\citep{yan2025step}                      & 70.1 & 65.8 & 67.6 & 53.0 & 0.0 & 60.5 \\
		MAI-UI-2B~\citep{zhou2025mai}                        & 62.8 &56.7 &59.3 &40.3 &- & 52.0\\
		MAI-UI-8B~\citep{zhou2025mai}                        & 72.0 &63.3 &66.0 &51.0 &- & 60.1\\
		MAI-UI-32B~\citep{zhou2025mai}                       & 73.6 & 72.4 & \underline{73.9} & 57.7 & - & 67.6\\
		\midrule
		\rowcolor{gray!15}
		\multicolumn{7}{l}{\textit{Ours}} \\
		UI-Venus-1.5-2B                       & 67.4 & 66.1 & 66.4 & 44.3 & 7.4 & 59.4 \\
		UI-Venus-1.5-8B                       & 79.7 & \underline{76.1} & 72.3 & 60.4 & \textbf{22.2} & \underline{69.7} \\
		UI-Venus-1.5-30B-A3B                      & \underline{80.1} & \underline{76.1} & \textbf{75.1} & 61.1 & \underline{9.3} & \textbf{70.6} \\
		\bottomrule
	\end{tabular*}
	\vspace{0.4em}
	\caption{Performance comparison on \textbf{OS-World-G}. For each benchmark, the best and second-best performing models are indicated in \textbf{bold} and \underline{underlined}, respectively. Asterisk (*) indicates results that may require verification with original sources.}
	\label{tab:osworld_g}
\end{table*}

\begin{table*}[t]
	\centering
	\footnotesize
	\setlength{\tabcolsep}{0pt}
	\begin{tabular*}{\textwidth}{@{\extracolsep{\fill}}l *{6}{c}}
		\toprule
		\textbf{Models} &
		\textbf{\begin{tabular}[c]{@{}c@{}}Text \\ Matching\end{tabular}} &
		\textbf{\begin{tabular}[c]{@{}c@{}}Element \\ Recognition\end{tabular}} &
		\textbf{\begin{tabular}[c]{@{}c@{}}Layout \\ Understanding\end{tabular}}&
		\textbf{\begin{tabular}[c]{@{}c@{}}Fine-grained \\ Manipulation\end{tabular}}&
		\textbf{Refusal} &
		\textbf{Avg} \\
		\midrule
		\rowcolor{gray!15}
		\multicolumn{7}{l}{\textit{General VLMs}} \\
		Qwen3-VL-2B*~\citep{Qwen3-VL}              & 73.2 & 64.2 & 70.8 & 47.0 & 0.0 & 60.6 \\
		Qwen3-VL-8B*~\citep{Qwen3-VL}             & 78.2 & 71.5 & 72.3 & 55.7 & 1.9 & 67.0 \\
		Qwen3-VL-30B-A3B*~\citep{Qwen3-VL}          & 77.8 & 75.8 & 74.7 & 54.4 & 5.6 & 69.3 \\
		\midrule
		\rowcolor{gray!15}
		\multicolumn{7}{l}{\textit{GUI-specific Models}} \\
		OpenCUA-32B~\citep{wang2025opencuaopenfoundationscomputeruse}                     & 63.2 & 79.9 & \textbf{84.9} & 62.1 & 7.4 & 70.2 \\
		GTA1-7B~\citep{yang2025gta1guitesttimescaling}                          & 63.2 & \underline{82.1} & 74.2 & \textbf{70.5} & 0.0 & 67.7 \\
		GTA1-32B~\citep{yang2025gta1guitesttimescaling}                         & 63.2 & \textbf{83.6} & \underline{84.4} & \textbf{70.5} & 0.0 & 72.2 \\
		UI-Venus-1.0-7B~\citep{gu2025ui}                 & 74.6 & 60.5 & 61.5 & 45.5 & - & 58.8 \\
		UI-Venus-1.0-72B~\citep{gu2025ui}                 & 82.1 & 71.2 & 70.7 & 64.4 & - & 70.4 \\
		Holo2-8B*~\citep{hai2025holo2modelfamily}                         & - & - & - & - & - & 70.1 \\
		Holo2-30B-A3B*~\citep{hai2025holo2modelfamily}                    & - & - & - & - & - & \underline{76.1} \\
		Step-GUI-4B*~\citep{yan2025step}                      & - & - & - & - & - & 66.9 \\
		MAI-UI-2B~\citep{zhou2025mai}                        & 70.9 &69.1 &72.7 &47.7 &- & 63.5\\
		MAI-UI-8B~\citep{zhou2025mai}                        & 77.4 &73.0 &78.3 &55.7 &- & 68.6\\
		MAI-UI-32B~\citep{zhou2025mai}                       & 79.7 & 79.4 & 81.0 & 61.7 & - & 73.9\\
		\midrule
		\rowcolor{gray!15}
		\multicolumn{7}{l}{\textit{Ours}} \\
		UI-Venus-1.5-2B                       & 75.1 & 70.0 & 73.5 & 52.3 & 7.4 & 65.6 \\
		UI-Venus-1.5-8B                       & \underline{82.4} & 81.5 & 80.2 & 59.7 & \textbf{22.2} & 74.1 \\
		UI-Venus-1.5-30B-A3B                      & \textbf{83.1} & \underline{82.1} & 83.4 & \underline{65.8} & \underline{9.3} & \textbf{76.4} \\
		\bottomrule
	\end{tabular*}
	\vspace{0.4em}
	\caption{Performance comparison on \textbf{OS-World-G-Refine}.
		For each benchmark, the best and second-best performing models are indicated in \textbf{bold} and \underline{underlined}, respectively. Asterisk (*) indicates results that may require verification with original sources.
	}
	\label{tab:osworld_g}
\end{table*}

\begin{table}[ht]
	\centering
	\footnotesize
	\begin{tabular}{lcccc}
		\toprule
		\textbf{Models}                                               & \textbf{Basic}   & \textbf{Functional} & \textbf{Spatial} & \textbf{Avg}     \\
		\midrule
		\rowcolor{gray!15}
		\multicolumn{5}{l}{\textit{General VLMs}}                                                                                                    \\
		Qwen3-VL-2B*~\citep{Qwen3-VL}                                 & 16.4             & 19.1                & 4.6              & 13.1             \\
		Qwen3-VL-8B*~\citep{Qwen3-VL}                                 & 27.8             & 29.6                & 9.4              & 21.9             \\
		Qwen3-VL-30B-A3B*~\citep{Qwen3-VL}                            & 31.2             & 31.9                & 14.6             & 25.6             \\
		\midrule
		\rowcolor{gray!15}
		\multicolumn{5}{l}{\textit{GUI-specific Models}}                                                                                             \\
		OpenCUA-7B~\citep{wang2025opencuaopenfoundationscomputeruse}  & -                & -                   & -                & 29.7             \\
		OpenCUA-32B~\citep{wang2025opencuaopenfoundationscomputeruse} & -                & -                   & -                & 33.3             \\
		OpenCUA-72B~\citep{wang2025opencuaopenfoundationscomputeruse} & -                & -                   & -                & 37.3             \\
		UI-Venus-1.0-7B~\citep{gu2025ui}                              & 36.1             & 32.8                & 11.9             & 26.5             \\
		UI-Venus-1.0-72B~\citep{gu2025ui}                             & 45.6             & 42.3                & 23.7             & 36.8             \\
		Holo2-8B*~\citep{hai2025holo2modelfamily}                     & 43.6             & 43.5                & 19.7             & 35.1             \\
		Holo2-30B-A3B*~\citep{hai2025holo2modelfamily}                & 51.0             & 50.1                & 23.2             & 40.9             \\
		Step-GUI-4B*~\citep{yan2025step}                              & 39.2             & 36.5                & 15.7             & 30.0             \\
		MAI-UI-2B~\citep{zhou2025mai}                                 & 41.0             & 41.2                & 10.4             & 30.3             \\
		MAI-UI-8B~\citep{zhou2025mai}                                 & 51.7             & 49.6                & 22.5             & 40.7             \\
		MAI-UI-32B~\citep{zhou2025mai}                                & 59.1             & \underline{57.1}    & 26.9             & \underline{47.1} \\
		\midrule
		\rowcolor{gray!15}
		\multicolumn{5}{l}{\textit{Ours}}                                                                                                            \\
		UI-Venus-1.5-2B                                               & \underline{63.5} & 51.5                & 21.6             & 44.8             \\
		UI-Venus-1.5-8B                                               & 56.3             & 52.4                & \underline{32.0} & 46.5             \\
		UI-Venus-1.5-30B-A3B                                          & \textbf{69.0}    & \textbf{59.3}       & \textbf{37.4}    & \textbf{54.7}    \\
		\bottomrule

	\end{tabular}
	\vspace{0.4em}
	\caption{Performance comparison on \textbf{UI-Vision}. For each benchmark, the best and second-best performing models are indicated in \textbf{bold} and \underline{underlined}, respectively.Asterisk (*) indicates results that may require verification with original sources.}
	\label{tab:ui-vison}
\end{table}

\section{Experiment Details of All Navigation Benchmarks}

\begin{table*}[t]
	\footnotesize
  \centering
  \caption{
Performance comparison on \textbf{VenusBench-Mobile}.
 The best performing model is indicated in \textbf{bold}.
  }
  \label{tab:mainresults}
  \resizebox{\textwidth}{!}{
  \begin{tabular}{lccccccccc|c}
    \toprule
    \textbf{Agent} & \textbf{FA} & \textbf{CF} & \textbf{VA} & \textbf{MR} & \textbf{GSA} & \textbf{GUIM} & \textbf{HGB} & \textbf{NR} & \textbf{BC} & \textbf{Total} \\
    \midrule
    \rowcolor{gray!15}
    \multicolumn{11}{l}{\textit{General VLMs}} \\
    Qwen3-VL-8B~\cite{Qwen3-VL}         & 18.2 & 4.6 & 18.8 & 0.0 & 0.0 & 0.0 & 0.0 & 6.3 & {10.0} & 6.7 \\
    Qwen3-VL-30B-A3B~\cite{Qwen3-VL}    & {22.7} & 4.6 & 18.8 & 0.0 & 0.0 & 0.0 & 5.9 & 6.3 & {10.0} & 8.7 \\
    \midrule
    \rowcolor{gray!15}
    \multicolumn{11}{l}{\textit{GUI-specific Models}} \\
    UI-Venus-7B~\cite{gu2025ui}         & 13.6 & 4.6 & {25.0} & 0.0 & {10.0} & 0.0 & 0.0 & 18.8 & 0.0 & 8.1 \\
    UI-Venus-72B~\cite{gu2025ui}        & {22.7} & 4.6 & 12.5 & 0.0 & {10.0} & 0.0 & \textbf{17.7} & \textbf{50.0} & 0.0 & {15.4} \\
    GUI-Owl-7B~\cite{ye2025mobile}          & 13.6 & 0.0 & 18.8 & 0.0 & 0.0 & {11.1} & {2.9} & 12.5 & 0.0 & 6.7 \\
    MA3{\small(GUI-Owl-7B)}~\cite{ye2025mobile} & 18.2 & {9.1} & 6.3 & 0.0 & 0.0 & 0.0 & {11.8} & {31.3} & \textbf{20.0} & {12.1} \\
    MAI-UI-2B~\cite{zhou2025mai}           & 9.1 & 0.0 & 18.8 & 0.0 & 0.0 & 0.0 & 0.0 & 25.0 & {10.0} & 6.7 \\
    MAI-UI-8B~\cite{zhou2025mai}           & 9.1 & \textbf{13.6} & {25.0} & 0.0 & {10.0} & {11.1} & 5.9 & {31.3} & {10.0} & {12.8} \\
    \midrule
    \rowcolor{gray!15}\multicolumn{11}{l}{\textit{Ours}} \\
    UI-Venus-1.5-2B     & {22.7} & 0.0 & 12.5 & 0.0 & {10.0} & {11.1} & {2.9} & 18.8 & 0.0 & 8.7 \\
    UI-Venus-1.5-8B     & {22.7} & 0.0 & {25.0} & 0.0 & 0.0 & \textbf{22.2} & 8.8 & \textbf{50.0} & \textbf{20.0} & {16.1} \\
    UI-Venus-1.5-30B-A3B& \textbf{40.9} & 0.0 & \textbf{37.5} & \textbf{10.0} & \textbf{20.0} & \textbf{22.2} & {14.7} & {43.8} & 0.0 & \textbf{21.5} \\
    \bottomrule
  \end{tabular}
  }
\end{table*}

\end{appendix}

\end{document}